%% file: main_a1.tex
\documentclass{article}

 \usepackage[preprint]{neurips_2026}

\title{Uniform Stability and Generalization Error of \\GD and SGD on Fixed-Point Parameters}

\author{
  Jonghyun Shin \\
  Department of Artificial Intelligence\\
  Korea University\\
  Seoul, Korea \\
  \texttt{uenjgieonj5448@korea.ac.kr} \\
  \And
  Sejun Park\thanks{Corresponding author}\\
  Department of Artificial Intelligence\\
  Korea University\\
  Seoul, Korea \\
  \texttt{sejun.park000@gmail.com} 
}

\input{packages}
\input{fonts}

\input{macros}

\AtBeginDocument{%
    \crefname{section}{Section}{Sections}%
    \crefname{appendix}{Appendix}{Appendices}%
    \crefname{subsection}{Section}{Sections}%
    \crefname{figure}{Figure}{Figures}%
}

\begin{document}

\maketitle

\input{0-Abstract_a1}
\input{1-Introduction_a1}

\input{2-Preliminaries_a1}

\input{3-Deterministic_Rounding_a1}

\input{4-Stochastic_Rounding_a1}

\input{5-Discussions_a1}

\input{6-Conclusion_a1}

\clearpage

\clearpage

\bibliography{reference_a1}
\bibliographystyle{plainnat}
\clearpage
\appendix
\clearpage
\input{appendix_a1}

\end{document}

%% file: fonts.tex
\newcommand{\mathboldcommand}[1]{\mathbb{#1}}

\newcommand{\bbE}{\mathboldcommand{E}}

\newcommand{\bbN}{\mathboldcommand{N}}

\newcommand{\bbP}{\mathboldcommand{P}}

\newcommand{\bbR}{\mathboldcommand{R}}

\newcommand{\bbZ}{\mathboldcommand{Z}}

\usepackage{euscript}
\newcommand{\mathcalcommand}[1]{\mathcal{#1}}

\newcommand{\mcD}{\mathcalcommand{D}}

\newcommand{\mcZ}{\mathcalcommand{Z}}

\DeclareMathAlphabet{\mathpzc}{T1}{pzc}{m}{it}

%% file: macros.tex
\DeclareMathOperator*{\argmin}{arg\,min}

\newcommand*{\commentout}[1]{}

\newlength{\parskiptrue}
\makeatletter

\newcommand*\wthelper[2]{%
        \hbox{\dimen@\accentfontxheight#1%
                \accentfontxheight#11.1\dimen@
                $\m@th#1\widetilde{#2}$%
                \accentfontxheight#1\dimen@
        }%
}
\newcommand*\accentfontxheight[1]{%
        \fontdimen5\ifx#1\displaystyle
                \textfont
        \else\ifx#1\textstyle
                \textfont
        \else\ifx#1\scriptstyle
                \scriptfont
        \else
                \scriptscriptfont
        \fi\fi\fi
}

\newcommand*\whhelper[2]{%
        \hbox{\dimen@\accentfontxheight#1%
                \accentfontxheight#11.2\dimen@
                $\m@th#1\widehat{#2}$%
                \accentfontxheight#1\dimen@
        }%
}
\makeatother

\makeatletter
\newcommand{\oset}[3][0ex]{%
  \mathrel{\mathop{#3}\limits^{
    \vbox to#1{\kern-3\ex@
    \hbox{$\scriptstyle#2$}\vss}}}}
\makeatother

\newcommand*{\defeq}{\triangleq}

\newcommand*{\zr}{\boldsymbol{0}_d}
\newcommand*{\one}{\boldsymbol{1}_d}

\newcommand*{\sr}{\mathsf{SR}} %
\newcommand*{\dr}{\mathsf{DR}}

\makeatletter
\def\moverlay{\mathpalette\mov@rlay}
\def\mov@rlay#1#2{\leavevmode\vtop{%
   \baselineskip\z@skip \lineskiplimit-\maxdimen
   \ialign{\hfil$\m@th#1##$\hfil\cr#2\crcr}}}
\newcommand{\charfusion}[3][\mathord]{
    #1{\ifx#1\mathop\vphantom{#2}\fi
        \mathpalette\mov@rlay{#2\cr#3}
      }
    \ifx#1\mathop\expandafter\displaylimits\fi}
\makeatother

\newcommand*{\inn}[2]{\left\langle{#1},{#2}\right\rangle}

\newcommand*{\gd}{{\text{\rm{GD}}}}
\newcommand*{\sgd}{{\text{\rm{SGD}}}}
\newcommand*{\drgd}{{\dr\text{\rm{-GD}}}}
\newcommand*{\drsgd}{{\dr\text{\rm{-SGD}}}}
\newcommand*{\srgd}{{\sr\text{\rm{-GD}}}}
\newcommand*{\srsgd}{{\sr\text{\rm{-SGD}}}}
\newcommand*{\grid}{\bbZ_\Delta}
\newcommand*{\gridd}{\bbZ_\Delta^d}
\newcommand*{\alg}{{\text{\rm{alg}}}}

%% file: 0-Abstract_a1.tex
\begin{abstract}
We analyze generalization error, uniform stability, and uniform argument stability of gradient descent (GD) and stochastic gradient descent (SGD) over discrete parameter spaces, where each update involves deterministic or stochastic rounding. We show that deterministic rounding degrades the generalization error of GD on convex, Lipschitz, and smooth loss functions, increasing the rate from $O(T/n)$ to $O(T/\sqrt{n})$, and establish matching lower bounds. We further prove that uniform stability of GD becomes $\Omega(T)$, showing that stability-based generalization bounds are vacuous in this setting. In contrast, for the same losses, stochastic gradient descent with deterministic rounding admits nontrivial uniform stability guarantees, which differ qualitatively from the real-valued case and exhibit distinct dependencies on the number of iterations and the dimension: we prove tight bounds $O(T/n)$ for one dimension and $O(T^2/n)$ for higher dimensions. We also show that stochastic rounding can introduce generalization error that increases with the dimension; such a phenomenon is absent in standard real-valued optimization and in the deterministic rounding case. Finally, we provide upper bounds on uniform argument stability for stochastic rounding schemes and show that these bounds are tight when the loss can be represented as a sum of coordinate-wise functions.
\end{abstract}

%% file: 1-Introduction_a1.tex
\section{Introduction}

Gradient-based optimization algorithms are the backbone of modern machine learning and are widely used to train models ranging from linear predictors to deep neural networks. Starting from classical methods such as gradient descent (GD) and stochastic gradient descent (SGD), numerous variants have been developed to improve convergence speed and robustness in large-scale and non-convex settings \citep{duchi11adagrad,tieleman12RMSProp,kingma15adam,liu25muon}. These algorithms have been extensively studied, with a rich theory characterizing their convergence behavior under various assumptions \citep{bottou18,reddi18,zhou24convergence}, as well as their generalization performance in statistical learning frameworks \citep{hardt16,Bassily20,zhou20genadap,Nguyen2022}. However, most existing analyses rely on an idealized setting in which the optimization variables lie in a continuous space and all arithmetic operations are performed exactly.

In practice, optimization algorithms are implemented on digital hardware using discrete parameters, such as fixed-point or floating-point representations \citep{goldberg91,gupta15}. In this setting, the parameter space is inherently discrete, and every arithmetic operation incurs round-off errors. As a result, the optimization dynamics can differ fundamentally from their real-valued counterparts: updates are quantized and numerical errors accumulate over time. Recent works have begun to study gradient-based methods in discrete or quantized settings by establishing convergence guarantees \citep{Markov23,xin25,xia25}. Nevertheless, a theoretical understanding of how discrete parameters affect generalization remains limited. %

Beyond convergence, understanding the generalization properties of optimization algorithms is a central question in statistical learning theory. Classical approaches based on uniform convergence provide bounds on the gap between empirical and population risks uniformly over a hypothesis class. However, such bounds typically depend on the ambient dimension \citep{shalev14,feldman16} and are agnostic to the specific optimization algorithm used, often leading to overly pessimistic bounds in modern high-dimensional settings. This limitation has motivated \emph{algorithm-dependent} analyses, which take into account the dynamics of the optimization process.

A canonical framework in this direction is \emph{algorithmic stability} \citep{Bousquet02,Shalev10}. In this framework, a learning algorithm $A$ maps a dataset $S = (z_1,\dots,z_n)$ to a parameter vector $A(S)$, and the generalization error is bounded by the sensitivity of $A(S)$ to small perturbations of the dataset. In particular, \emph{uniform stability} measures how much the loss changes when a single training example is replaced, while \emph{uniform argument stability} measures how much the learned parameters themselves change under such perturbations. When the loss function is Lipschitz, these two notions are directly related, allowing one to bound the generalization error by tracking how perturbations propagate through the optimization dynamics. This framework has been particularly successful for gradient-based methods, yielding dimension-independent generalization guarantees for GD and SGD \citep{hardt16,chen2018,Bassily20}.
\vspace{-0.1in}
\subsection{Our contributions}
\vspace{-0.05in}
In this paper, we revisit this framework with fixed-point parameters. Specifically, we consider gradient-based optimization over a discrete parameter space $\{\Delta x:x\in\mathbb Z\}^d$, where each update involves deterministic or stochastic rounding.
Within this setting, we analyze how finite-precision effects influence generalization error, uniform stability, and uniform argument stability\footnote{$\text{generalization error}\le\text{uniform stability}\le \text{Lipschitz constant}\times\text{uniform argument stability}$ (see \cref{sec:stability}).} of gradient-based optimization algorithms when the loss functions are convex, Lipschitz, and smooth. We show that the discrete and inexact nature of machine arithmetic fundamentally alters the behavior of these stability measures, leading to qualitatively different generalization guarantees compared to the classical real-valued setting. In particular, our contributions can be summarized as below (see also \cref{table:summary}). Here, $\drgd$, $\srgd$, $\drsgd$, and $\srsgd$ denote (stochastic) gradient descent on $\{\Delta x:x\in\mathbb Z\}^d$ with deterministic/stochastic rounding, and $\gd$ and $\sgd$ denote standard (stochastic) gradient descent on $\bbR^d$ without rounding operations. In addition, we hide the Lipschitz constant and $\Delta$ for concise representation; we present the precise bounds including them in \cref{sec:dr_results,sec:sr_results}.\vspace{-0.05in}

\begin{itemize}[leftmargin = 0.2in]
\item We first prove that the generalization error of $T$-step $\drgd$ is upper bounded by $O(\eta T/\sqrt{n})$\footnote{$\eta$ denotes the learning rate and $n$ denotes the number of samples.} when the underlying samples are binary (\cref{thm:ub-ge-drgd}); we also show the tightness of this bound, even for an arbitrary sample space (\cref{thm:lb-ge-drgd}).
Since the generalization error of $\gd$ is $O(\eta T/n)$ for convex, Lipschitz, and smooth losses, these results imply that deterministic rounding to the fixed-point parameter space degrades the sample complexity of $\gd$. 

\item %
However, a stability-based approach cannot achieve this bound: the uniform stability of $\drgd$ can be lower bounded by $\Omega(T)$ (\cref{thm:lb-us-drgd}). Hence,
the uniform stability of $\drgd$ can only produce vacuous generalization bounds in this case.
\item On the other hand, this is not the case for $\drsgd$. We prove that the uniform stability of $\drsgd$ is bounded by $O(\eta T/n)$ for $d=1$ (\cref{thm:ub-us-drsgd-1dim}), and it is bounded by $O(\eta T^2/n)$ for $d>1$ (\cref{thm:ub-us-drsgd}). We further show the tightness of these bounds (\cref{thm:lb-us-drsgd-1dim,thm:lb-us-drsgd}), which implies that depending on the ambient dimension $d$, the uniform stability can have a different dependence on $T$, while such a dependency was not observed in $\sgd$.
\item We next consider stochastic rounding and show that the generalization errors of $\srgd$ and $\srsgd$ can increase with $d$. Specifically, \cref{thm:lb-ge-srgd} states that there exists a convex, Lipschitz, and smooth loss and a data distribution for which the generalization errors of $\srgd$ and $\srsgd$ are $\Omega(\sqrt{\eta d^{1/2}}/n)$.\footnote{\label{fn:1}We assume $\eta d^{1/2}=\Omega(1)$ here. See \cref{sec:sr_results} for bounds without this assumption.} We note that such a dimension dependency is from the stochasticity in rounding, and does not arise in $\gd,\sgd,\drgd$, and $\drsgd$.
\item We also prove upper bounds on the uniform argument stability: $O(\sqrt{\eta d^{1/2}T^3}/n)$ for $\srsgd$ and $O(\eta T/n+\min\{1,\eta d^{1/2}T/n\}\sqrt{\eta d^{1/2}T})$  for $\srgd$ (\cref{thm:ub-uas-srgd-general,thm:ub-uas-srsgd-general}).\cref{fn:1} We further reduce the $T^{3/2}$ dependency here to $T$ for losses that can be represented as a sum of coordinate-wise functions (\cref{thm:ub-uas-srgd-separable}), and show the tightness of these reduced bounds (\cref{thm:lb-uas-srgd}). 
\end{itemize}

%% file: 2-Preliminaries_a1.tex
\input{Summary}

\section{Preliminaries}
\subsection{Notations and problem setup}
For $n\in\bbN$, we use $[n]$ to denote $\{1, \cdots, n\}$. For $d\in\bbN$, we use $\zr$ to denote $(0, \cdots, 0)\in\bbR^d$ and $\one$ to denote $(1, \cdots, 1)\in\bbR^d$. For $n\in\bbN$ and $S, S'\in\bbR^n$, we say $S$ and $S'$ are neighboring, denoted by $S\simeq S'$, if they differ at a single entry. For $\Delta>0$, we use $\grid$ to denote the set $\{\Delta n: n\in\bbZ\}$. For any $x\in\bbR$, we use $\lfloor x\rfloor_\Delta$ to denote $\max\{y\in\grid:y\le x\}$. We say a function $f:\bbR^d\to\bbR$ is separable if there exist functions $f_1, \cdots, f_d:\bbR\to\bbR$ such that $f(x_1, \cdots, x_d) = \sum_{i=1}^df_i(x_i)$. For a logical proposition $P$, we define $\mathbbm{1}[P]=1$ if $P$ is true and $\mathbbm{1}[P]=0$ if $P$ is false.

\subsection{Supervised learning and generalization error}
We consider a supervised learning problem. Let $\mcD$ be an underlying distribution over a sample space $\mcZ$ and  
let $S = (z_1, \cdots, z_n)$ be a dataset of $n$ samples drawn i.i.d. from $\mcD$.
For a parameter space $\bbR^d$, a loss function $f:\bbR^d\times\mcZ\to\bbR$, and a learning algorithm $A_f:\mcZ^n\to\bbR^d$, we define the \emph{generalization error} of $A_f$ as follows:
\begin{align*}
\varepsilon_g(A_f;\mcD)\defeq|\bbE_{A_f, S=(z_1,\dots,z_n)}[f(A_f(S);S) - f(A_f(S); \mcD)]|    
\end{align*}
where $f(w; \mcD)\defeq\bbE_{z\sim\mcD}[f(w; z)]$ and $f(w;S)\defeq\frac{1}{n}\sum_{i=1}^n f(w; z_i)$ denote the population loss and the empirical loss, respectively. 
Throughout this paper, we primarily focus on loss functions such that for any $z\in\mcZ$, $f(\cdot,z)$ is differentiable, convex, Lipschitz, and smooth.
\begin{definition}\label{asm:convexity}
$f:\bbR^d\to\bbR$ is convex if for any $u,v\in\bbR^d$, 
$f(u)\ge f(v)+\langle\nabla f(v), u-v\rangle.$
\end{definition}

\begin{definition}\label{asm:lipshictz}
$f:\bbR^d\to\bbR$ is $L$-Lipschitz if for any $u,v\in\bbR^d$,  $|f(u) - f(v)|\le L\|u-v\|_2.$
\end{definition}

\begin{definition}\label{asm:smooth}
$f:\bbR^d\to\bbR$ is $M$-smooth if for any $u,v\in\bbR^d$, $\|\nabla f(u) - \nabla f(v)\|_2\le M\|u-v\|_2.$
\end{definition}

\subsection{Gradient descent and stochastic gradient descent with rounding operations}\label{sec:rounding}
Gradient descent (GD) and stochastic gradient descent (SGD) are iterative learning algorithms defined as follows: given a learning rate $\eta>0$, initial parameter $w_0\in\bbR^d$, total number of iterations $T$, 
\begin{align}
A^{\gd}_{f,\theta}(S)=G^\gd_{f,\eta,S,T}\circ\cdots\circ G^\gd_{f,\eta,S,1}(w_0)~~\text{and}~~ A^{\sgd}_{f,\theta}(S)=G^\sgd_{f,\eta,S,T}\circ\cdots\circ G^\sgd_{f,\eta,S,1}(w_0)\label{eq:gd}
\end{align}
where $\theta=(\eta,T,w_0)$,
\begin{align}
G^\gd_{f,\eta,S,t}(w)=w-\eta\cdot\nabla f(w;S),~~\text{and}~~ G^\sgd_{f,\eta,S,t}(w)=w-\eta\cdot\nabla f(w;z_{i_t})\label{eq:gd2}
\end{align}
for all $t\in[T]$.
Here, $i_t$ is a random variable following the uniform distribution over $[n]$.

Given a discrete grid $\bbZ_\Delta^d$, we define deterministic rounding $\dr$ and stochastic rounding $\sr$ that maps $\bbR^d$ to $\bbZ_\Delta^d$ as follows:
\begin{align}
\dr(x)_j \in
\argmin_{y\in\bbZ_\Delta}|y-x_j|, \quad \sr(x)_j = \lfloor x_j\rfloor_\Delta+\Delta\cdot\mathbbm{1}\left[u_j\le \frac{x_j - \lfloor x_j \rfloor_\Delta}{\Delta}\right]\label{eq:rounding}
\end{align}
where $\dr(x)_j$ is the even multiple of $\Delta$ if $\min_{y\in\bbZ_\Delta}|y-x_j|=0.5\Delta$ and $u_j$ is a random variable following the uniform distribution over $[0,1]$.
We note that choosing an arbitrary $\dr(x)_j$ in $\argmin_{y\in\bbZ_\Delta}|y-x_j|$ does not affect our main results as long as it is deterministic.

Given $\theta=(\eta,T,w_0\in\bbZ_\Delta^d)$, we define GD with deterministic rounding onto $\bbZ_\Delta^d$ ($\drgd$) as 
\begin{align*}
A_{f,\theta}^\drgd(S)=G^\drgd_{f,\eta,S,T}\circ\cdots\circ G^\drgd_{f,\eta,S,1}(w_0)~~\text{where}~~G^\drgd_{f,\eta,S,t}(w)=w-\dr(\eta\cdot\nabla f(w;S)).
\end{align*}
We similarly define GD with stochastic rounding ($\srgd$) and SGD with deterministic and stochastic rounding ($\drsgd$ and $\srsgd$, respectively). For $\srgd$ and $\srsgd$, $\sr$ at the $t$-th iteration uses fresh independent random variables $u_{t,1}, \cdots, u_{t,d}$ in \cref{eq:rounding}.

\subsection{Algorithmic stability}\label{sec:stability}

Algorithmic stability is a framework for bounding the generalization error of a learning algorithm based on its sensitivity to dataset perturbations.
Specifically, if a learning algorithm $A$ is \emph{symmetric} (i.e., the distribution of $A(S)$ does not depend on the order of the samples in $S$), then we can represent the generalization error as stated in the following lemma. 
\begin{lemma}[Lemma 7 in \citep{Bousquet02}]\label{lem:average-stability}
Let $\mcD$ be a distribution over $\mcZ$, $f:\bbR^d\times\mcZ\to\bbR$ be a loss function, $S = (z_1, \cdots, z_n)$ and $S' = (z_1', z_2, \cdots, z_n)$ be neighboring datasets where $z_1',z_1,\dots,z_n$ are drawn from $\mcD$ independently. If $A_f$ is symmetric, then
\begin{align*}
\varepsilon_g(A_f; \mcD) = \left| \mathbb{E}_{A_f,S', z_1}\left[ f(A_f(S); z_1) - f(A_f(S'); z_1) \right] \right|.
\end{align*}
\end{lemma}
From \cref{lem:average-stability}, one can observe that the \emph{uniform stability} of $A_f$ defined as
\begin{align*}
\varepsilon_{us}(A_f) = \sup_{S\simeq S',z}\bbE_{A_f}[f(A_f(S);z) - f(A_f(S');z)]
\end{align*}
upper bounds the generalization error. %
In particular, if the loss function is $L$-Lipschitz, a typical process to bound the uniform stability is to bound the \emph{uniform argument stability} defined as follows:
\begin{align*}
\varepsilon_{uas}(A_f) = \sup_{S\simeq S'}\bbE_{A_f}[\|A_f(S) - A_f(S')\|_2].
\end{align*}
For randomized algorithms, the expectation here is taken under \emph{coupled} randomness: for $\alg\in\{\sgd,\drsgd,\srsgd\}$, $A_{f,\theta}^\alg(S)$ and $A_{f,\theta}^\alg(S')$ share the same random indices $i_1, \cdots, i_T$ in \cref{eq:gd2}, and for $\alg\in\{\srgd,\srsgd\}$, stochastic rounding operations use the same random variable $u_{t,j}$ for each $t\in[T]$ and $j\in[d]$ in \cref{eq:rounding}. 
We note that this coupling is used only for our upper bounds for $\varepsilon_{uas}(A_f)$; our lower bounds hold for any joint distribution of $A_f(S)$ and $A_f(S')$.

Then, the following relation between the generalization error, uniform stability, and uniform argument stability holds:
\begin{align*}
\varepsilon_g(A_f;\mcD)\le\varepsilon_{us}(A_f)\le L\cdot\varepsilon_{uas}(A_f).
\end{align*}

We say $A^\alg_{f,\theta}$ is \emph{non-expansive} if for any $u,v\in\bbZ_\Delta^d$ and $t\in[T]$, there exists a joint distribution $p$ of $G_{f,\eta,S,t}^\alg(u)\defeq u'$ and $G_{f,\eta,S,t}^\alg(v)\defeq v'$ such that $\bbE_p[\|u'-v'\|_2]\le\|u-v\|_2$. %

%% file: Summary.tex
\begin{table*}[t]
\begin{center}
\caption{A summary of generalization and stability bounds for iterative algorithms.
All results are for convex, $L$-Lipschitz, and smooth losses except for \citep{Bassily20}, which considers convex, Lipschitz but non-smooth losses.
$\varepsilon_g,\varepsilon_{us},\varepsilon_{uas}$ denote the generalization error, uniform stability, and uniform argument stability, respectively. The lower bounds hold for worst-case examples, and the upper bounds hold for all examples.
For simplicity, we consider $L,\Delta=\Theta(1)$, $n>T$, $\eta=O(1)$, and $\eta d^{1/2}=\Omega(1)$ here; the precise bounds without these assumptions are in \cref{sec:dr_results,sec:sr_results}. %
}
\renewcommand{\arraystretch}{1.2}
\label{table:summary}
\begin{tabular}{| c | c | c |} 
 \hline
{\bf Algorithm}  & {\bf Reference} & {\bf Bound} \\ 
 \hline\hline
\multirow{3}{*}{GD}&\citet{nikolakakis25}&$\varepsilon_{g}\gtrsim \eta T/n$
\\ &\citet{chen2018} & $ \varepsilon_{us}\asymp \eta T/n$ \\ \cline{2-3}
&  \citet{Bassily20}   &$\varepsilon_{uas}\asymp\eta \sqrt{T}+\eta T/n$\textsuperscript{(1)}

\\
\hline
\multirow{3}{*}{SGD}&\citet{hardt16} &$\varepsilon_{us} \lesssim \eta T/n$  \\ %
&\citet{zhang22} & $\varepsilon_{us} \gtrsim \eta T/n$  \\\cline{2-3}
&\citet{Bassily20} & $\varepsilon_{uas}\asymp \eta T^{3/2}/n$\textsuperscript{(1)}

\\ \hline
 \hline

\multirow{3}{*}{$\drgd$} &\cref{thm:ub-ge-drgd} &  $\varepsilon_g\lesssim\eta T /\sqrt{n}$\textsuperscript{(2)}  \\
 &\cref{thm:lb-ge-drgd} &  $\varepsilon_g\gtrsim T /\sqrt{n}$ \\ \cline{2-3}
&\cref{thm:lb-us-drgd} & $\varepsilon_{us}\gtrsim T$
\\\hline

\multirow{4}{*}{$\drsgd$} & \cref{thm:ub-us-drsgd-1dim}&$\varepsilon_{us}\lesssim T/n$\textsuperscript{(3)}  \\ 
&\cref{thm:lb-us-drsgd-1dim}&$\varepsilon_{us}\gtrsim T/n$\\\cline{2-3}
& \cref{thm:lb-us-drsgd}& $\varepsilon_{us}\gtrsim T^2/n$\textsuperscript{(4)}\\
&\cref{thm:ub-us-drsgd}&$\varepsilon_{us}\lesssim \eta  T^2/n$
\\\hline
\hline

\multirow{4}{*}{\vspace{-0.2cm}$\srgd$} & \cref{thm:lb-ge-srgd} & $\varepsilon_g\gtrsim\sqrt{\eta  d^{1/2}}/n$\textsuperscript{(5)} \\ \cline{2-3}
&\cref{thm:ub-uas-srgd-general} & $\varepsilon_{uas}\lesssim\eta T/n+\min\{1,\eta d^{1/2}T/n\}\sqrt{\eta d^{1/2}T} $\\ \cline{2-3}

&\cref{thm:ub-uas-srgd-separable}&$\varepsilon_{uas}\lesssim \sqrt{\eta d^{1/2}T/n}\hspace{0.01in}\hspace{-0.01in}$\textsuperscript{(6)}
\\ 
& \cref{thm:lb-uas-srgd} &  $\varepsilon_{uas}\gtrsim\eta T/n+\sqrt{\min \{ 1, \eta d^{1/2} T/n \}\eta d^{1/2}T/n}\hspace{0.01in}\hspace{-0.01in}$\textsuperscript{(6)}\\
\hline

\multirow{4}{*}{\vspace{-0.1cm}$\srsgd$} &\cref{thm:lb-ge-srgd}&  $\varepsilon_g\gtrsim\sqrt{\eta d^{1/2}}/n$\textsuperscript{(5)}\\ \cline{2-3}
&\cref{thm:ub-uas-srsgd-general} &$ \varepsilon_{uas}\lesssim \sqrt{\eta d^{1/2}T^3}/n$ \\ \cline{2-3}
&\cref{thm:ub-uas-srgd-separable}&$\varepsilon_{uas}\lesssim T\sqrt{\eta d^{1/2}}/n$\textsuperscript{(6)}\\
&\cref{thm:lb-uas-srgd} & $\varepsilon_{uas}\gtrsim T\sqrt{\eta  d^{1/2}}/n$\textsuperscript{(6)}

\\ \hline
\end{tabular}
\end{center}
{\footnotesize
$n$, $\eta$, $T$, and $d$ denote the number of samples, learning rate, total number of iterations, and parameter dimension, respectively.\\ %
\textsuperscript{(1)} requires $d\ge \min\{T, 1/\eta^2\}$ for the lower bounds,
\textsuperscript{(2)} requires the sample space to be binary, 
\textsuperscript{(3)} requires $d=1$, 
\textsuperscript{(4)} requires $d\ge 2$, 
\textsuperscript{(5)} requires $T=1$, and  \textsuperscript{(6)} requires that the loss function is a sum of coordinate-wise functions.
}\vspace{-0.05in}
\end{table*}

%% file: 3-Deterministic_Rounding_a1.tex
\vspace{-0.05in}
\section{Generalization error and uniform stability of $\drgd$ and 
$\drsgd$}\label{sec:dr_results}
\vspace{-0.05in}
In this section, we formally present our bounds on generalization error, uniform stability, and uniform argument stability of gradient methods over $\gridd$ with deterministic rounding schemes. 
In our theoretical results, we use asymptotic notation to hide universal constants. Specifically, for nonnegative functions $f$ and $g$ of $n,T,d,\eta,\Delta,M,$ and $L$, we write $f\lesssim g$ if there exists a universal constant $C>0$ such that $f\le C\cdot g$. We write $f\gtrsim g$ if $g\lesssim f$, and $f\asymp g$ if both $f\lesssim g$ and $f\gtrsim g$ hold. We use $f=O(g)$, $f=\Omega(g)$, and $f=\Theta(g)$ to denote $f\lesssim g$, $f\gtrsim g$, and $f\asymp g$, respectively. %

We first provide lower and upper bounds on the generalization error of $\drgd$. 

\begin{theorem}\label{thm:ub-ge-drgd}
Let $n,d,T\in\bbN$, $\eta,\Delta, L>0$, $w_0\in\gridd$, $\mcD$ be a distribution over $\{0,1\}$, and $f:\bbR^d\times\{0,1\}\to\bbR$ be an $L$-Lipschitz loss function. Then, for $\theta=(\eta, T, w_0)$, %
\begin{align*}
\varepsilon_g(A_{f,\theta}^{\drgd}; \mcD)\le \frac{2\eta L^2 T}{\sqrt n}.
\end{align*}
\end{theorem}
\begin{theorem}\label{thm:lb-ge-drgd}
Let $n,T\in\bbN$ and $\eta,\Delta,L>0$ such that $\eta L>c\Delta$ for some universal constant $c>0.5$, and $n\ge \max\{8c/(2c-1),6\}$. Then, there exist a distribution $\mcD$ over $\{0,1\}$ and an $L$-Lipschitz, affine loss function $f:\bbR\times\{0,1\}\to\bbR$ such that for  $\theta = (\eta, T,w_0=0)$, %
\begin{align*}
\varepsilon_g(A^{\drgd}_{f,\theta};\mcD) \gtrsim \max\left\{\frac{\Delta LT}{\sqrt{n}}, LT\left(\frac{\eta L } {n} - \Delta\right)\right\}.
\end{align*}
\end{theorem}

\cref{thm:ub-ge-drgd} shows that the generalization error of $\drgd$ is $O(\eta L^2T/\sqrt{n})$ when the sample space is binary (i.e., $\mcZ=\{0,1\}$), and %
\cref{thm:lb-ge-drgd} shows the tightness of this bound:
there exists an affine function $f$ with $|f'(x;z)|\le L$ and a data distribution $\mcD$ such that the generalization error of $\dr$-GD is lower bounded by $\Omega(\Delta LT/\sqrt{n})$.
We note that our counterexample function $f$ in \cref{thm:lb-ge-drgd} can easily be extended to general $d>1$ and a non-binary sample space, e.g., consider $g(x_1,\dots,x_d)=f(x_1)$ and data distribution supported on two samples.

These results show that $\drgd$ and GD have fundamentally different generalization properties since a stability-based approach can achieve $\varepsilon_g(A_{f,\theta}^\gd;\mcD)=O(\eta L^2T/n)$ \citep{chen2018}, i.e., the deterministic rounding onto $\grid$ induces worse sample complexity compared to the exact real counterpart. We further note that, as $\Delta\to0$, our bound in \cref{thm:lb-ge-drgd} converges to the existing uniform stability bound $O(\eta L^2T/n)$ \citep{chen2018}. %

To prove \cref{thm:lb-ge-drgd}, we explicitly choose the underlying data space $\mcZ=\{0,1\}$ and a loss function $f(w,z)=-zLw$ for $z\in\mcZ$.
We then show that for some neighboring datasets $S\simeq S'$ where $S'$ contains more ones compared to $S$, the $\drgd$ iterate for $S'$ moves strictly faster than that for $S$. That is, we have $\dr(\eta\nabla f(w_t;S))>\dr(\eta\nabla f(w_t';S'))$, which implies $|w_T'-w_T|\ge\Delta T$, and hence, $|f(A^{\drgd}_{f,\theta}(S);1)-f(A^{\drgd}_{f,\theta}(S');1)|\ge L\Delta T$.
Here, we choose the underlying data distribution over $\mcZ$ so that such an event occurs with probability $\Omega(1/\sqrt{n})$, which results in the $\Omega(\Delta LT/\sqrt{n})$ lower bound.
We note that such an observation only holds for $\gd$ with deterministic rounding: for $\gd,\sgd,\drsgd,\srgd,$ and $\srsgd$, $|w_T'-w_T|\to0$ as $n\to\infty$ (see \cref{sec:sr_results} for the $\srgd$ and $\srsgd$ cases).
We present the full proofs of \cref{thm:lb-ge-drgd,thm:ub-ge-drgd} in  \cref{sec:pfthm:lb-ge-drgd,sec:pfthm:ub-ge-drgd}, respectively.

We next show that the bound in \cref{thm:ub-ge-drgd} cannot be achieved by using the uniform stability.
While the bound in \cref{thm:ub-ge-drgd} converges to zero as $n\to\infty$,
the following theorem shows that it is not the case for the uniform stability of $\drgd$. In particular, it shows that the uniform stability can be independent of $n$ but can increase  with $T$ at the same time. %
\begin{theorem}\label{thm:lb-us-drgd}
Let $n,T\in\bbN$, $\eta,L,\Delta>0$ with $\eta L>0.5\Delta$. Then, there exists an $L$-Lipschitz, affine loss function $f:\bbR\times\{0,1\}\to\bbR$ such that for  $\theta = (\eta, T,w_0=0)$, %
\begin{align*}
\varepsilon_{us}(A^\drgd_{f,\theta}) \gtrsim\max\left\{\Delta L T, LT\left(\frac{\eta L } {n} - \Delta\right)\right\}.
\end{align*}
\end{theorem}
\cref{thm:lb-us-drgd} shows the existence of an affine function $f$ with $|f'(x;z)|\le L$ such that the uniform stability-based approach can only produce a vacuous bound $\Omega(\Delta LT)$. As in \cref{thm:lb-ge-drgd}, the bound in \cref{thm:lb-us-drgd} naturally extends to a general input dimension $d>1$. 

The main proof idea behind \cref{thm:lb-us-drgd} is almost identical to that of \cref{thm:lb-ge-drgd}: we also use the counterexample loss function $f=-zLw$ and two datasets $S\simeq S'$ such that $\dr(\eta\nabla f(w_t;S))>\dr(\eta\nabla f(w_t';S'))$.
However, unlike the $\Omega(\Delta LT/\sqrt{n})$ bound in \cref{thm:lb-ge-drgd}, the bound $\Omega(\Delta LT)$ in \cref{thm:lb-us-drgd} does not contain the $1/\sqrt{n}$ factor. This is because the definition of the uniform stability $\varepsilon_{us}$ does not depend on the underlying data distribution, which incurs the $1/\sqrt{n}$ factor in \cref{thm:lb-ge-drgd} (see our discussions after \cref{thm:lb-ge-drgd} and \cref{sec:pfthm:lb-ge-drgd}). See \cref{sec:pfthm:lb-us-drgd} for the full proof.

\citet{Bassily20} also derived the lower bound $\Omega(\eta L T/n+\eta L\sqrt T)$ on the uniform argument stability of $\gd$ for high-dimensional non-smooth and convex losses, by exploiting the expansiveness of $\gd$ for such losses. Specifically, they designed loss functions so that at the $t$-th iteration, one $\gd$ iterate using $S$ moves $\Theta(\eta)$ along the $t$-th coordinate, while the other $\gd$ iterate using $S'\simeq S$ stays at the origin. Hence, the distance between two iterates scales with $\sqrt{T}$.
We note that $\drgd$ is non-expansive for convex and smooth losses if $d=1$ (see \cref{lem:dr-non-expansiveness} in \cref{sec:pfthm:ub-us-drsgd-1dim}), but the rounding operation incurs a similar effect so that the bound in \cref{thm:lb-us-drgd} follows.
In addition, while there is a clear gap between the generalization error and the uniform stability of $\drgd$ (\cref{thm:ub-ge-drgd,thm:lb-us-drgd}), to our knowledge, such an observation has not been made for $\gd$ with convex, Lipschitz, but non-smooth losses due to the lack of tight bounds on the generalization error. %

On the other hand, %
the uniform stability of $\drsgd$ %
converges to zero as $n\to\infty$. We present the proofs of \cref{thm:ub-us-drsgd-1dim,thm:lb-us-drsgd-1dim} in \cref{sec:pfthm:ub-us-drsgd-1dim,sec:pfthm:lb-us-drsgd-1dim}, respectively.

\begin{theorem}\label{thm:ub-us-drsgd-1dim}
Let $\mcZ$ be a sample space, $n, T\in\bbN$, $\Delta,L,\eta,M>0$ with $\eta <2/M$,  $w_0\in\grid$, %
and $f:\bbR\times\mcZ\to\bbR$ be a convex, $L$-Lipschitz, and $M$-smooth loss function. Then, for $\theta=(\eta, T, w_0)$, 
\begin{align*}
\varepsilon_{us}(A^{\drsgd}_{f,\theta}) \le \frac{(2\eta L+\Delta) L T}{n}.
\end{align*}
\end{theorem}
\cref{thm:ub-us-drsgd-1dim} states that when $d=1$, the uniform stability of $\dr$-SGD for convex, Lipschitz, and smooth losses is upper bounded by ${(2\eta L+\Delta) L T}/{n}$. %
As $\Delta\to0$, this bound converges to the existing bound $O(\eta L^2 T/n)$ for $\sgd$ \citep{hardt16}.

Unlike generalization bounds $\Theta(\eta L^2T/n)$ for GD and SGD with a convex, Lipschitz, and smooth loss \citep{hardt16,chen2018},
these results show that there is a fundamental difference between $\dr$-GD and $\dr$-SGD in terms of the sample complexity: for convex, Lipschitz, and smooth loss function, we have $\varepsilon_g(A^{\dr\text{\rm-GD}}_{f,\theta})\ge\Omega(\Delta LT/\sqrt{n})$ in general but $\varepsilon_g(A^{\dr\text{\rm-SGD}}_{f,\theta})\le O((\eta L+\Delta)LT/n)$ when $d=1$.
Such a difference is due to the stochasticity in $\drsgd$. Specifically, for coupled $\drsgd$ iterates for neighboring datasets $S\simeq S'$, at each iteration, they choose the same loss function with probability $(n-1)/n$. Hence, the coupled $\sgd$ iterates are non-expansive with probability $(n-1)/n$ before rounding, which also implies that they are non-expansive after rounding since $d=1$. Namely, the distance between them can increase with probability $1/n$.

We also show the tightness of the bound in \cref{thm:ub-us-drsgd-1dim}. The proof of \cref{thm:lb-us-drsgd-1dim} is in \cref{sec:pfthm:lb-us-drsgd-1dim}.
\begin{theorem}\label{thm:lb-us-drsgd-1dim}
Let $n, T\in\bbN$ and $\Delta,L,\eta>0$ such that $\eta L>0.5\Delta$. Then there exists an $L$-Lipschitz affine loss function $f:\bbR\times\{0,1\}\to\bbR$ such that for $\theta=( \eta, T, w_0=0)$, %
\begin{align*}
\varepsilon_{us}(A_{f,\theta}^\drsgd)\gtrsim\frac{(2\eta L-\Delta)LT}{n}.
\end{align*}
\end{theorem}

However, the bound linear to $T$ in \cref{thm:ub-us-drsgd-1dim} does not extend to general $d>1$. %
Our next result shows that the uniform stability of $\drsgd$ %
can
increase with $T^2$ when $T/n\le1$. 

\begin{theorem}\label{thm:lb-us-drsgd}
Let $n, d, T\in\bbN$ with $n,d,T\ge 2$, and $\Delta>0$. Then, there exist a convex, $\Delta$-Lipschitz and $\left(\frac{9}{10T^2}+\frac{8}{5}\right)$-smooth loss function $f:\bbR^d\times\{0,1\}\to\bbR$ such that for  $\theta=(\eta =1, T,w_0 =  \zr)$, %
\begin{align*}
\varepsilon_{us}(A_{f,\theta}^\drsgd) \gtrsim \min\left\{1, \frac{T}{n}\right\}\cdot\Delta^2 T.
\end{align*}
\end{theorem}

\cref{thm:lb-us-drsgd} also implies the difference between the generalization properties of $\sgd$ and $\drsgd$ when $d>1$ since $\varepsilon_{us}(A^\sgd_{f,\theta})=O(\eta L^2T/n)$ \citep{hardt16}. Here, the main difference between $\sgd$ and $\drsgd$ is that when $d>1$, $\drsgd$ can be expansive due to the rounding scheme, even if a loss function is convex and smooth. Using this property, we design a counterexample loss function $f:\bbR^d\times\{0,1\}\to\bbR$ that satisfies $\dr(\eta\nabla f(\zr;0))=\zr$ and $\dr(\eta\nabla f(\zr;1))=u\ne0$, i.e., $\drsgd$ iterate never moves with dataset $S=(0,\dots,0)$ but it can move with $S'=(1,0,\dots,0)\simeq S$, with probability $1/n$.
Here, we design $f$ so that once the $\drsgd$ iterate touches $u$, then for each iteration, the distance between the iterate and the origin increases by $\Theta(\Delta)$.
Here, the $\Theta(T/n)$ term of the bound in \cref{thm:lb-us-drsgd} can be considered as the probability that one is sampled from $S'$ in the first $T/2$ $\drsgd$ iterations. Once one is sampled, the distance between the $T$-th $\drsgd$ iterate for $S'$ and the origin will be $\Theta(\Delta T)$, which results in a $\Theta(\Delta^2T)$ difference in the loss function values $f(A^\drsgd_{f,\theta}(S);0)$ and $f(A^\drsgd_{f,\theta}(S');0)$. For the full proof, see \cref{sec:pfthm:lb-us-drsgd}.

We next show that the bound in \cref{thm:lb-us-drsgd} is tight in general, by providing a matching upper bound in the following theorem. The proof of \cref{thm:ub-us-drsgd} is presented in \cref{sec:pfthm:ub-us-drsgd}.
\begin{theorem}\label{thm:ub-us-drsgd}
Let $\mcZ$ be a sample space, $n, d,T\in\bbN$, $\Delta,L,\eta>0$, $w_0\in\gridd$, and  %
$f:\bbR^d\times\mcZ\to\bbR$ be an $L$-Lipschitz loss function. Then, for $\theta=( \eta, T, w_0)$, %
\begin{align*}
\varepsilon_{us}(A_{f,\theta}^\drsgd) \le \min\left\{1, \frac{T}{n}\right\}\cdot 8\eta L^2T.
\end{align*}
\end{theorem}

We remark that for $\sgd$ with non-smooth and convex losses, the uniform argument stability is
$\Theta(\eta LT/n+\min\{1,T/n\}\eta L\sqrt T)$ \citep{Bassily20}.
This bound is also derived using the expansiveness of counterexample losses. While the distance between the coupled $\drsgd$ iterates can increase by $\Delta$ per iteration, the squared distance between the coupled $\sgd$  iterates can increase by $\Theta(\eta^2)$, yielding the $T^{3/2}$ factor in the bound.

%% file: 4-Stochastic_Rounding_a1.tex
\section{Generalization error and uniform stability of $\srgd$ and $\srsgd$}\label{sec:sr_results}

We next present our bounds for $\srgd$ and $\srsgd$.
\begin{theorem}\label{thm:lb-ge-srgd}
Let $\eta,L,\Delta>0$ and $n,d\in\bbN$ with $n\ge 14$ and $\eta L<\Delta\sqrt{d}$. There exist a data distribution $\mcD$ over $\{0,1\}$ and a convex, $L$-Lipschitz, and $(4L/\Delta)$-smooth loss function $f:\bbR^d\times\{0,1\}\to\bbR$ such that for $\theta = (\eta,T=1,w_0=\zr)$, %
\begin{align*}
\varepsilon_g(A_{f,\theta}^\srgd,\mcD),\varepsilon_g(A_{f,\theta}^\srsgd,\mcD)\gtrsim \frac{\sqrt{\min\{\Delta, \eta L  d^{1/2}\}\eta L^3 d^{1/2}}}{n}.
\end{align*}
\end{theorem}
\cref{thm:lb-ge-srgd} shows that the generalization error of $\srgd$ and $\srsgd$ is inherently dependent on the ambient dimension:
there exist a data distribution $\mcD$ and a convex, $L$-Lipschitz, and $(4L/\Delta)$-smooth loss function $f$ such that the generalization error of $\srgd$ and $\srsgd$ with $T=1$ can scale with (at least) $d^{1/4}$. %
This result shows that the generalization properties of $\srgd$ and $\srsgd$ are fundamentally different from those of $\gd$ and $\sgd$, which are independent of $d$ \citep{hardt16}. %

Here, the dimension dependency in \cref{thm:lb-ge-srgd} is from the stochastic rounding. For $\gd$ and $\sgd$, the distance between coupled iterates for neighboring datasets increases at most $O(\eta L/n)$. Then, after $T$ iterations, the distance can be bounded by $O(\eta LT/n)$, and hence, the difference in the function value can be bounded by $O(\eta L^2T/n)$. However, if 
we apply the stochastic rounding to two points with distance $\ell$, the resulting distance can be $\Omega(\ell\sqrt{d})$ in expectation (e.g., consider one point is $\zr$ and the other point is $(\ell/\sqrt{d},\dots,\ell/\sqrt{d})$).
Motivated by this observation, we design the loss function and a data distribution so that the generalization error scales with $d$. See \cref{sec:pfthm:lb-ge-srgd} for the full proof.

We now shift our focus to the stability bounds of $\srgd$ and $\srsgd$ for general $T>1$.
To this end, we present an upper bound of the uniform argument stability of $\srsgd$ in the following theorem. %
\begin{theorem}\label{thm:ub-uas-srsgd-general}
Let $\mcZ$ be a sample space, $n,d,T\in\bbN$, $\Delta,\eta,L,M>0$ with $\eta< 2/M$, $w_0\in\gridd$, and $f:\bbR^d\times\mcZ\to\bbR$ be a convex, $L$-Lipschitz, and $M$-smooth loss function. Then, for $\theta = (\eta,T,w_0)$, %
\begin{align*}
\varepsilon_{uas}(A_{f,\theta}^\srsgd) \le\frac{8\eta LT}{n}+\min\left\{1, \frac{T}{n}\right\}\cdot\sqrt{2\Delta\eta L d^{1/2}T}.
\end{align*}
\end{theorem}

\cref{thm:ub-uas-srsgd-general} provides an upper bound on $\varepsilon_{uas}(A_{f,\theta}^\srsgd)$ that scales with $d^{1/4}$; such dimension dependence is unavoidable in general as shown by \cref{thm:lb-ge-srgd}.
Unlike bounds for $\sgd$, which scales with $T$, our bound scales with $T^{3/2}$
as in the uniform argument stability bound for $\sgd$ on a non-smooth convex loss: $\varepsilon_{uas}(A_{f,\theta}^\sgd) = O(\min\{1,T/n\}\eta L\sqrt T+\eta LT/n)$ \citep{Bassily20}.
This follows from the potential expansiveness of $\srsgd$ even for convex and smooth losses. Here, the $\min\{1,T/n\}$ term corresponds to the probability that different examples (i.e. $z_1$ and $z_1'$) are selected from neighboring datasets during the coupled SGD iterates, and the $\Theta(\sqrt{\Delta\eta Ld^{1/2}T})$ term is the accumulated distance between two coupled $\srsgd$ iterates after the differing samples are selected. 
We also note that this bound recovers the uniform argument stability bound for $\sgd$ as $\Delta\to0$.
For the full proof of \cref{thm:ub-uas-srsgd-general}, see \cref{sec:pfthm:ub-uas-srsgd-general}.

We also provide an upper bound on the uniform argument stability of $\srgd$ in the following theorem, whose proof is presented in \cref{sec:pfthm:ub-uas-srgd-general}. %
\begin{theorem}\label{thm:ub-uas-srgd-general}
Let $\mcZ$ be a sample space, $n,d,T\in\bbN$, $\Delta,\eta,L,M>0$ with $\eta< 2/M$, $w_0\in\gridd$, 
and $f:\bbR^d\times\mcZ\to\bbR$ be a convex, $L$-Lipschitz and $M$-smooth loss function. Then, for  $\theta = (\eta,T,w_0)$, %
\begin{align*}
\varepsilon_{uas}(A_{f,\theta}^\srgd) \le\frac{6\eta LT}{n}+\min\left\{1, \frac{2\eta L\sqrt dT}{n\Delta}\right\}\cdot\sqrt{\Delta^2+2\Delta\eta Ld^{1/2}\left(\frac1n+T-1\right)}.
\end{align*}
\end{theorem}
Unlike the lower bound on the uniform stability of $\drgd$ (\cref{thm:lb-us-drgd}), the bound in \cref{thm:ub-uas-srgd-general} converges to zero as $n\to\infty$ when $\Delta$ is independent of $n$. %
In $\drgd$, even an $O(1/n)$ change in the gradient of empirical loss can deterministically move the update across a rounding threshold, producing a separation of order $\Delta$ between two iterates on neighboring datasets even after the first iterate (see our discussion after \cref{thm:lb-us-drgd}). In contrast, under stochastic rounding, a small change in the gradient of empirical loss causes different rounded values only with probability proportional to the magnitude of that change relative to $\Delta$. Therefore, the probability that two $\srgd$ iterates first separate over $T$ iterations is of order $\min\{1,\eta L\sqrt dT/(n\Delta)\}$, yielding the sample-dependent bound in \cref{thm:ub-uas-srgd-general}. 
Here, 
the following term $\Theta(\sqrt{\Delta^2+\Delta\eta Ld^{1/2}T})$ corresponds to the accumulated distance between two $\srgd$ iterates after the first separation. 
Specifically, the squared distance between the two $\srgd$ iterates is at most $O(\Delta^2+\Delta\eta L\sqrt d/n)$ at the first separation, and then increases at most $O(\Delta\eta L \sqrt d)$ per iteration. %
See \cref{sec:pfthm:ub-uas-srgd-general} for more details.

The bound in \cref{thm:ub-uas-srgd-general} can be interpreted differently depending on the value of $\Delta$. %
For $\Delta\lesssim \Delta_1 = \Theta(\eta L \sqrt dT/n)$, the bound becomes $\varepsilon_{uas}(A_{f,\theta}^\srgd) = O(\eta LT/n+\sqrt{\Delta\eta Ld^{1/2}T})$. %
For $\Delta_1\lesssim \Delta \lesssim \Delta_2=\Theta( \eta L \sqrt dT)$, the bound becomes $O(\eta^{3/2} L^{3/2}d^{3/4}T^{3/2}/(n\Delta^{1/2}))$, which can scale with 
$T^{3/2}$, e.g., for constant $\Delta$ when $\Delta_1=O(1)$. %
If $\Delta\gtrsim \Delta_2$, the bound becomes $O(\eta L \sqrt dT/n)$. 
As in the $\srsgd$ case, we recover the uniform argument stability bound for $\gd$ as $\Delta\to0$.

While the bounds in \cref{thm:ub-uas-srsgd-general,thm:ub-uas-srgd-general} can scale with 
$T^{3/2}/n$, our next results show that the bounds can be relaxed to $T/n$ (or $\sqrt{T/n}$) if losses are separable (i.e., sum of coordinate-wise functions) as in the following theorem.  The proof of \cref{thm:ub-uas-srgd-separable} is presented in \cref{sec:pfthm:ub-uas-srgd-separable}.

\begin{theorem}\label{thm:ub-uas-srgd-separable}
Let $\mcZ$ be a sample space, $d,n,T\in \bbN$, $\Delta,\eta,M,L>0$ such that $\eta<1/M$, $w_0\in \gridd$, and $f:\bbR^d\times\mcZ\to\bbR$ be a  convex, $L$-Lipschitz, $M$-smooth, and separable loss function. %
Then, for  $\theta = (\eta,T,w_0)$, 
\begin{align*}
\varepsilon_{uas}(A_{f,\theta}^\srgd)&\le \frac{6\eta LT}{n}+\sqrt{\frac{2\Delta\eta L d^{1/2}T}{n}}, \quad\varepsilon_{uas}(A_{f,\theta}^\srsgd) \le \frac{2\eta LT}{n}+\frac{T\sqrt{2\Delta\eta Ld^{1/2}}}n.
\end{align*}
\end{theorem}
Unlike the potential expansiveness of $\srgd$ and $\srsgd$ for convex and smooth losses, they are non-expansive if the losses are additionally separable.
By exploiting this property, we can better bound the accumulated distance between coupled iterates, and derive the bounds in \cref{thm:ub-uas-srgd-separable}.

We lastly show that the bounds in \cref{thm:ub-uas-srgd-separable} are tight in general, as stated in the following theorem. The proof of \cref{thm:lb-uas-srgd} is in \cref{sec:pfthm:lb-uas-srgd}.
\begin{theorem}\label{thm:lb-uas-srgd}
Let $\eta, \Delta,L>0$ and $n,T,d\in\bbN$. Then, there exists an $L$-Lipschitz affine loss function $f:\bbR^d\times\{0,1\}\to\bbR$ such that for $\theta = (\eta, T, w_0=\zr)$, if $\eta<n\Delta\sqrt d/L$, then%
\begin{align*}
\varepsilon_{uas}(A_{f,\theta}^\srgd) &\gtrsim \frac{\eta L T}{n}+ \frac{\sqrt{\min\{n\Delta, \eta Ld^{1/2}T\}\eta L  d^{1/2} T}}{n}.
\end{align*}
In addition, if $\eta <\Delta\sqrt d/L$, then 
\begin{align*}
\varepsilon_{uas}(A_{f,\theta}^\srsgd) &\gtrsim \min\left\{1,\frac{T}{n}\right\}\left(\eta L+\sqrt{\min\{\Delta, \eta L  d^{1/2}\}\eta L  d^{1/2}} \right).
\end{align*}
\end{theorem}
Together with \cref{thm:ub-uas-srgd-separable}, \cref{thm:lb-uas-srgd} can tightly characterize the uniform argument stability of $\srgd$ and $\srsgd$ when the loss function is separable.
Specifically, for $\eta L/(n\sqrt d)<\Delta\lesssim \eta L\sqrt d T/n$, we have $\varepsilon_{uas}(A_{f,\theta}^\srgd) = \Theta(\sqrt{\Delta\eta L d^{1/2}T/n})$. Likewise, for $T<n$ and $\eta L/\sqrt d<\Delta\lesssim \eta L \sqrt d$, it holds that $\varepsilon_{uas}(A_{f,\theta}^\srsgd) = \Theta(\sqrt{\Delta\eta L d^{1/2}}T/n)$.
This result shows a qualitative difference between $\srgd$ and $\srsgd$: since $\srsgd$ contains the extra $\sqrt{T/n}$ compared to $\srgd$, $\srsgd$ is more favorable in sample complexity, but less favorable in its dependence on $T$.

%% file: 5-Discussions_a1.tex
\section{Discussions}

\textbf{Limitations.}
Our theoretical analysis isolates parameter rounding in idealized $\gd$/$\sgd$ updates (see \cref{sec:rounding}). Accordingly, it may not cover several ingredients in real implementations. %
First, our model assumes that the gradient of the real-valued loss is computed exactly. However, automatic differentiation over machine-representable numbers can exhibit incorrect behavior even when the loss is differentiable due to the round-off error in the intermediate computations.
Second, modern low-precision training often relies on quantization-aware training \citep{Jacob17QuantizationAT}, which maintains high-precision parameters while training a model with fake-quantization operators in the forward pass. Thus, unlike our setup, it does not directly optimize over a discrete parameter space.

\textbf{Extension to non-convex loss functions.}
Although our main results are stated for convex, Lipschitz, and smooth losses, our stability upper bound results can be extended beyond this class. In particular, the proofs of \cref{thm:ub-us-drsgd-1dim,thm:ub-us-drsgd,thm:ub-uas-srsgd-general,thm:ub-uas-srgd-general,thm:ub-uas-srgd-separable} 
only use the loss values and gradients at the points in $\gridd$ that can be visited by the rounded iteration.  
Hence, these results may also apply to non-convex losses whose gradients on $\gridd$ can be interpolated by those of a convex, Lipschitz, and smooth function. This interpolation condition can be checked using known characterizations of smooth convex interpolation. For example, \citet{Taylor15SmoothSC} show that for any finite collection $\{(x_i, g_i,f_i)\}_{i=1}^n\subset\bbR^d\times\bbR^d\times\bbR$, where each $x_i$ is an interpolation point with gradient $g_i$ and value $f_i$, 
there exists a convex and $M$-smooth function $f:\bbR^d\to\bbR$ satisfying $f(x_i)=f_i$ and $\nabla f(x_i)=g_i$ for all $i\in[n]$ if and only if
\begin{align*}
f_i\ge f_j + g_j^\top (x_i - x_j)+\frac{1}{2M}\|g_i-g_j\|_2^2~~\text{for all }i~\text{and }j.
\end{align*}

\textbf{Extension to floating-point numbers.}
Unlike fixed-point numbers, floating-point numbers form a non-uniform grid: the spacing between representable numbers depends on the exponent. Thus, the magnitude of rounding errors depends on the scale of the iterates themselves. 
Nevertheless, our analysis may extend to floating-point arithmetic in regimes where the iterates remain within a fixed exponent range. In such a local regime, the set of representable numbers behaves similarly to fixed-point numbers.
However, extending our results to the full floating-point setting requires controlling changes in the exponent along the optimization trajectory, as well as the resulting scale-dependent rounding errors. Developing a stability and generalization theory that captures these effects is an important direction for future work.

%% file: 6-Conclusion_a1.tex
\section{Conclusion}
In this work, we analyze the generalization error and uniform stability of $\gd$ and $\sgd$ with deterministic and stochastic rounding onto fixed-point spaces, on convex, Lipschitz, and smooth losses. We first show that deterministic rounding can degrade the generalization error to $\Theta(T/\sqrt n)$, with a vacuous $\Omega(T)$ uniform stability bound. In contrast, $\sgd$ with deterministic rounding yields tight, dimension-dependent bounds: $O(T/n)$ for one dimension and $O(T^2/n)$ for higher dimensions, which is qualitatively different from $\sgd$ without rounding.
Furthermore, we show that both for $\gd$ and $\sgd$ with stochastic rounding, the generalization error can scale with the ambient dimension. We also provide the upper and lower bounds on uniform argument stability, which are tight when the loss function can be represented as a sum of coordinate-wise functions. We believe our results contribute to bridging the gap between theoretical optimization and the real implementation.

%% file: appendix_a1.tex
\section{Proofs for $\drgd$ and $\drsgd$}
In this section, we prove the generalization and stability bounds for $\drgd$ and $\drsgd$. We first present the proofs of \cref{thm:ub-ge-drgd,thm:lb-ge-drgd}, and \cref{thm:lb-us-drgd} for $\drgd$ in \cref{sec:pfthm:ub-ge-drgd,sec:pfthm:lb-ge-drgd}, and \cref{sec:pfthm:lb-us-drgd}, respectively. We then present the proofs of \cref{thm:ub-us-drsgd-1dim}, \cref{thm:lb-us-drsgd-1dim}, \cref{thm:lb-us-drsgd}, and \cref{thm:ub-us-drsgd} for $\drsgd$ in \cref{sec:pfthm:ub-us-drsgd-1dim}, \cref{sec:pfthm:lb-us-drsgd-1dim}, \cref{sec:pfthm:lb-us-drsgd}, \cref{sec:pfthm:ub-us-drsgd}, respectively.

\subsection{Proof of \cref{thm:ub-ge-drgd}}\label[appendix]{sec:pfthm:ub-ge-drgd}
In this section, we prove \cref{thm:ub-ge-drgd}. Let $\mcD = Ber(p)$ be a data distribution, $S\sim\mcD^n$ be a sample drawn from $\mcD$, and $(w_t)_{t=1}^T$ be a $\drgd$ trajectory from $A_{f,\theta}^\drgd(S)$. Then, the generalization error of $A_{f,\theta}^\drgd$ is given by
\begin{align}
\varepsilon_{g}(A_{f,\theta}^\drgd; \mcD) = |\bbE_{S\sim \mcD^n}[f(w_T; S) -f(w_T; \mcD)]|.\label{eq:pfthm:ub-ge-drgd}
\end{align}
Let $k_S$ be the number of ones in $S$, i.e., $k_S = \|S\|_1$. Then, we have $k_S\sim Bin(n, p)$, and hence we can explicitly expand \cref{eq:pfthm:ub-ge-drgd}:
\begin{align*}
\varepsilon_{g}(A_{f,\theta}^\drgd; \mcD) &= \left|\bbE\left[\frac{k_S}{n}f(w_T; 1) + \left(1-\frac{k_S}{n}\right)f(w_T; 0) - \left(p f(w_T;1)+(1-p)f(w_T;0)\right)\right]\right|\\
&=\left|\bbE\left[\left(\frac{k_S}{n} - p\right)(f(w_T;1)-f(w_T;0))\right]\right|.
\end{align*}
Without loss of generality, we assume that $f(w_0;1)=f(w_0;0)=0$.  Otherwise, we replace $f(w_T;1)$ and $f(w_T;0)$ by $g(w_T;1) = f(w_T;1) - f(w_0;1)$ and $g(w_T;0)=f(w_T;0) - f(w_0;0)$, respectively; note that 
\begin{align*}
\bbE\left[\left(\frac{k_S}{n} - p\right)(g(w_T;1)-g(w_T;0))\right] =& \bbE\left[\left(\frac{k_S}{n} - p\right)(f(w_T;1)-f(w_T;0))\right]\\&+\bbE\left[\frac{k_S}{n} - p\right](f(w_0;1)-f(w_0;0))\\
=&\bbE\left[\left(\frac{k_S}{n} - p\right)(f(w_T;1)-f(w_T;0))\right]
\end{align*}
where the last equality follows since $\bbE[k_S] = np$. Then, by the triangle inequality, we have
\begin{align*}
\varepsilon_g(A_{f,\theta}^\drgd; \mcD)&\le \left|\bbE\left[\left(\frac{k_S}{n} - p\right)(f(w_T;1)-f(w_T;0))\right]\right|\\
&\le\bbE\left[\left|\frac{k_S}{n} - p\right|\cdot(|f(w_T;1)| + |f(w_T;0)|)\right].
\end{align*}

In this proof, we first prove the upper bound of $|f(w_T;1)|+|f(w_T;0)|$, and then show the upper bound of $\bbE[|k_S/n-p|]$. For the first part, we consider the upper bound of $\|w_T - w_0\|_2$. Then, since $f$ is $L$-Lipschitz and $f(w_0;0)=f(w_0;1)=0$, we have $|f(w_T;0)|+|f(w_T;1)|\le 2L\|w_T-w_0\|_2$.

To this end, we introduce the following lemma. The proof of \cref{lem:ub-dr} is in \cref{sec:pflem:ub-dr}
\begin{lemma}\label{lem:ub-dr}
Let $d\in\bbN$ and $\Delta,R>0$ and $a\in\bbR^d$ such that $\|a\|_2\le R$. Then, it holds that 
\begin{align*}
\|\dr(a)\|_2\le 2R.
\end{align*}
\end{lemma}
Since $f$ is $L$-Lipschitz, $\|\eta \nabla f(w;S)\|_2\le \eta L$ for all $w\in \gridd$. Then, by \cref{lem:ub-dr}, we have $\|w_t - w_{t-1}\|_2 = \|\dr(\eta\nabla f(w_{t-1};S))\|_2\le 2\eta L$ for all $t\in[T]$. Then, we have
\begin{align*}
\|w_T - w_0\|_2 \le \|w_T - w_{T-1}\|_2+\cdots+\|w_1-w_0\|_2\le 2\eta LT.
\end{align*}
Then, we have $|f(w_T;0)|+ |f(w_T;1)|\le 2L\|w_T-w_0\|_2\le 4\eta L^2T$.

We now consider $\bbE[|k_S/n-p|]$. By Jensen's inequality, we have
\begin{align*}
\bbE\left[\left|\frac{k_S}{n} - p\right|\right]\le \sqrt{\bbE\left[\left(\frac{k_S}{n} - p\right)^2\right]} = \sqrt{Var\left(\frac{k_S}{n} \right)}=\sqrt{\frac{p(1-p)}{n}}\le \frac{1}{2\sqrt n}
\end{align*}
where the last inequality follows from the fact that $p(1-p)$ is maximized when $p=1/2$.

Conclusively, we have
\begin{align*}
\varepsilon_g(A_{f,\theta}^\drgd, \mcD)
&\le\bbE\left[\left|\frac{k_S}{n} - p\right|\cdot(|f(w_T;1)| + |f(w_T;0)|)\right]\\&\le 4\eta L^2T \cdot\bbE\left[\left|\frac{k_S}{n} - p\right|\right]\le \frac{2\eta L^2T}{\sqrt n}
\end{align*}
which completes the proof.

\subsubsection{Proof of \cref{lem:ub-dr}}\label[appendix]{sec:pflem:ub-dr}
In this section, we prove \cref{lem:ub-dr}. Let $a= (a_1, \cdots, a_d)$. It suffices to show that any $x\in\bbR$, $|\dr(x)|\le 2|x|$. Then, we have
\begin{align*}
\|\dr(a)\|_2^2 = \sum_{i=1}^d|\dr(a_i)|^2 \le \sum_{i=1}^d 4 |a_i|^2 = (2\|a\|_2)^2\le (2R)^2
\end{align*}
which implies $\|\dr(a)\|_2\le 2R$ and this completes the proof.

Since $|x|<\Delta/2$ is trivial ($\dr(x)=0$), we assume that $|x|\ge \Delta/2$. Since $|x - \dr(x)|\le \Delta/2$, by the triangle inequality, we have
\begin{align*}
|\dr(x)|\le |x|+\frac\Delta 2\le |x|+|x|=2|x|
\end{align*}
which completes the proof.

\clearpage

\subsection{Proof of \cref{thm:lb-ge-drgd}}\label[appendix]{sec:pfthm:lb-ge-drgd}

In this section, we prove \cref{thm:lb-ge-drgd}. We define a loss function $f:\bbR\times\{0,1\}\to\bbR$ and an underlying data distribution $\mcD$ as follows:
\begin{align*}
\mcD = Ber(p), \quad f(w;z) = -z Lw
\end{align*}
for some $p\in(0,1)$; we will assign an explicit value to $p$ later. We note that $\nabla f(w;z) = -zL\in\{0,-L\}$. Let $S=(z_1, \cdots, z_n)$ and $S' = (z_1', z_2,\cdots, z_n)$ where $z_1', z_1, \cdots, z_n$ are independently sampled from $\mcD$, $\bar z_{S}=\|S\|_1/n$, $\bar z_{S'}=\|S'\|_1/n$, $(w_t)_{t=1}^T$ and $(w_t')_{t=1}^T$ be the $\drgd$ trajectories from $A_{f,\theta}^{\drgd}(S)$ and $A_{f,\theta}^{\drgd}(S')$, respectively. Since the gradients of the empirical loss functions on $S$ and $S'$ are $\nabla f(w; S) = -L\bar z_{S} $ and $\nabla f(w;S') = -L\bar z_{S'}$ for all $w$, we have
\begin{align*}
w_{T} =-T\dr(-\eta L \bar z_{S}), \quad w_{T}' = -T\dr(-\eta L \bar z_{S'}).
\end{align*}

Here, we do not use $\dr(-x) = -\dr(x)$ since a general tie-breaking rule need not make deterministic rounding an odd function.

Then, we have $w_T' - w_T = T(\dr(-\eta L \bar z_{S}) - \dr (-\eta L \bar z_{S'}))$. We now consider the generalization error of $A_{f,\theta}^\drgd$. By \cref{lem:average-stability}, $\varepsilon_g(A_{f,\theta}^\drgd;\mcD)$ can be expressed by
\begin{align*}
\varepsilon_g(A_{f,\theta}^\drgd; \mcD) &= \left\rvert\bbE[f(w_T';z_1') - f(w_T;z_1')]\right\rvert\\&=L\left\rvert\bbE\left[z_1'\left(w_T' - w_T\right)\right]\right\rvert.
\end{align*}
If $z_1=z_1'$, then $S = S'$ and hence $w_T = w_T'$. Furthermore, if $z_1'=0$, then $z_1'(w_T' - w_T)=0$. Thus, only the event $(z_1', z_1) = (1,0)$ contributes to this expectation, hence we obtain
\begin{align*}
    \varepsilon_g(A_{f, \theta}^\drgd; \mcD)&=p(1-p)L\left\rvert\bbE[w_T' - w_T|(z_1',z_1)=(1,0)]\right\rvert\\
&=p(1-p)LT|\bbE[\dr(-\eta L \bar z_{S}) - \dr (-\eta L \bar z_{S'})|(z_1', z_1) = (1,0)]|\\
&= p(1-p)LT\left|\bbE\left[\dr\left(-\frac{\eta LK}n\right) - \dr\left(-\frac{\eta L(K+1)}n\right)\right]\right|.
\end{align*}
where $K = \sum_{i=2}^nz_i$. Here, we note that $K\sim Bin(n-1, p)$ and
\begin{align*}
\bar z_S = \frac Kn, \quad \bar z_{S'} = \frac{K+1}n.
\end{align*}

To show the lower bound, we use the following two lemmas. See \cref{sec:pflem:lb-general-drgd,sec:pflem:lb-ge-drgd-binomial} for the proofs of \cref{lem:lb-general-drgd,lem:lb-ge-drgd-binomial}.

\begin{lemma}\label{lem:lb-general-drgd}
For any $a,b\in\bbR$, with $a>b$, we have
\begin{align*}
\dr(a) - \dr(b)\ge a-b-\Delta.
\end{align*}
\end{lemma}
\begin{lemma}\label{lem:lb-ge-drgd-binomial}
Let $n\in\bbN$, $\Delta, R>0$ such that $R>c\Delta$ for some fixed constant $c>0.5$, and $n\ge \max\{8c/(2c-1),6\}$.
Then, there exists $p\in(0,1)$ such that $p(1-p)=\Omega(1)$ and for $K\sim Bin(n-1,p)$, 
\begin{align*}
\bbE\left[\dr\left(-\frac{R K}{n}\right) - \dr\left(-\frac{R (K+1)}{n}\right)\right]= \Omega\left(\frac{\Delta}{\sqrt{n}}\right).
\end{align*}
\end{lemma}
Under the event $(z_1', z_1) = (1,0)$, $-\eta L\bar z_S - (-\eta L \bar z_{S'}) = \eta L/n$. Therefore, by \cref{lem:lb-general-drgd}, we have
\begin{align*}
\dr(-\eta L \bar z_{S}) - \dr (-\eta L \bar z_{S'})\ge \frac{\eta L}{n} - \Delta.
\end{align*}
Taking expectations gives
\begin{align*}
\bbE[\dr(-\eta L \bar z_{S}) - \dr (-\eta L \bar z_{S'})|(z_1', z_1)=(1,0)]\ge \frac{\eta L}{n}-\Delta.
\end{align*}
Furthermore, by applying \cref{lem:lb-ge-drgd-binomial} with $R = \eta L$, we obtain $p\in(0,1)$ such that $p(1-p)= \Omega(1)$ and $K\sim Bin(n-1,p)$ and
\begin{align*}
|\bbE[\dr(-\eta L \bar z_{S}) - \dr (-\eta L \bar z_{S'})|(z_1', z_1)=(1,0)]|\gtrsim \frac{\Delta}{\sqrt n}.
\end{align*}

Conclusively, we have
\begin{align*}
\varepsilon_g(A^{\drgd}_{f,\theta};\mcD) \gtrsim LT\max\left\{\frac{\Delta }{\sqrt{n}}, \left(\frac{\eta L } {n} - \Delta\right)\right\}
\end{align*}
which completes the proof.

\subsubsection{Proof of \cref{lem:lb-general-drgd}}\label[appendix]{sec:pflem:lb-general-drgd}
In this section, we prove \cref{lem:lb-general-drgd} by considering the lower bound of $\dr(a)$ and the upper bound of $\dr(b)$. Since $\dr(x)\in[x-\Delta/2,x+\Delta/2]$ for any $x\in\bbR$, we have
\begin{align*}
\dr(a)-\dr(b)\ge \left(a-\frac{\Delta}{2}\right)-\left(b+\frac{\Delta}{2}\right) = a-b-\Delta
\end{align*}
which completes the proof.

\subsubsection{Proof of \cref{lem:lb-ge-drgd-binomial}}\label[appendix]{sec:pflem:lb-ge-drgd-binomial}
In this section, we prove \cref{lem:lb-ge-drgd-binomial}. To this end, we find $k^\ast\in\bbN$ such that there exist constants $c_1,c_2\in(0,1)$ such that $c_1n \le k^\ast \le c_2n$ and
\begin{align}
\dr\left(-\frac{Rk^\ast}{n}\right) - \dr\left(-\frac{R (k^\ast+1)}{n}\right)  \ge \Delta.\label{eq:pflem:lb-ge-drgd-binomial}
\end{align}
Once such $k^\ast$ is found, we choose $p = k^\ast/(n-1)$. Then, $0<p<1$ and $\bbE[K] = k^\ast$. Here, the following lemma gives $\bbP(K=k^\ast) = \Omega(1/\sqrt n)$. See \cref{sec:pflem:binom-anti-concentrate} for the proof of \cref{lem:binom-anti-concentrate}.

\begin{lemma}\label{lem:binom-anti-concentrate}
Let $n\in\bbN$ and $p\in(0,1)$ such that $n\ge 2$ and $np\in\{1, \cdots, n-1\}$. For $X\sim Bin(n,p)$, it holds that 
\begin{align*}
\bbP(X = np) = \Omega\left(\frac{1}{\sqrt{n}}\right).
\end{align*}
\end{lemma}
Moreover, since the deterministic rounding operation is monotone, $\dr(-Rk/n) - \dr(-R(k+1)/n)$ is nonnegative for all $k\in\{0, \cdots, n-1\}$. Hence, it holds that
\begin{align*}
\bbE\left[\dr\left(-\frac{R K}{n}\right) - \dr\left(-\frac{R (K+1)}{n}\right)\right]&\ge \left(\dr\left(-\frac{R k^\ast}{n}\right) - \dr\left(-\frac{R (k^\ast+1)}{n}\right)\right)\bbP(K=k^\ast)\\&= \Omega\left(\frac{\Delta}{\sqrt{n}}\right).
\end{align*}

In the rest of the proof, we find $k^\ast\in[n-1], c_1$, and $c_2$ such that $k^\ast\in[c_1n, c_2n]$ and $k^\ast$ satisfies \cref{eq:pflem:lb-ge-drgd-binomial}.

Let $m=\lfloor R+0.5\Delta\rfloor_\Delta/\Delta$. Then, $m$ is the natural number satisfying $R\in[(m-0.5)\Delta, (m+0.5)\Delta)$. We split the construction into two cases: $m=1$ and $m>1$.

Suppose that $m=1$, which implies $0.5\Delta\le R< 1.5\Delta$. Then, we have $R\in(c\Delta,1.5\Delta)$. Since the sequence $\{-Rk/n\}_{k=0}^n$ crosses $-0.5\Delta$, there exists $k^\ast$ such that
\begin{align*}
\dr\left(-\frac{Rk^\ast}{n}\right) - \dr\left(-\frac{R(k^\ast+1)}{n}\right) \ge \Delta.
\end{align*}
Here, such $k^\ast$ satisfies 
\begin{align*}
\left|-\frac{Rk^\ast}{n} -\left( -\frac{\Delta}{2}\right)\right|\le \frac{R}{n}
\end{align*}
which is equivalent to 
$\left|k^\ast - \Delta n/(2R)  \right|\le 1$.

Since $\Delta/2R\in(1/3, 1/2c)$, it holds that
\begin{align*}
k^\ast\in\left[\frac{\Delta n}{2R}-1,\frac{\Delta n}{2R}+1\right]\subset\left(\frac{n}{3}-1,  n\left(\frac{1}{2c}+\frac1n\right)\right)
\end{align*}
Then, if $n\ge \max\{8c/(2c-1),4\}$, we have
\begin{align*}
\frac{n}{12}< k^\ast< n\left(\frac14 + \frac3{8c}\right)
\end{align*}
where the first inequality follows from $n\ge 4$, and the second inequality follows from $n\ge 8c/(2c-1)$, which is equivalent to $1/n\le 1/4- 1/(8c)$.

We now consider the case that $m>1$. As in the case $m=1$, we choose a rounding threshold near $R/2$ and find $k^\ast$ that induces separation between $\dr(-R(k^\ast+1)/n)$ and $\dr(-Rk^\ast/n)$.

If $m=2j$, then we set $q= (j-0.5)\Delta$, and if $m=2j+1$, then we set $q = (j+0.5)\Delta$. Since the sequence $\{Rk/n\}_{k=0}^n$ crosses $q$, there exists $k^\ast$ such that 
\begin{align*}
\dr\left(-\frac{Rk^\ast}{n}\right) - \dr\left(-\frac{R(k^\ast+1)}{n}\right) \ge \Delta.
\end{align*}
Here, such $k^\ast$ satisfies
\begin{align*}
\left|\frac{Rk^\ast}{n} - q  \right|\le \frac Rn
\end{align*}
which is equivalent to $|k^\ast - qn/R|\le 1$.

Here, since $R\in[(m-0.5)\Delta, (m+0.5)\Delta)$, if $m=2j$, we have
\begin{align*}
\frac qR = \frac{(m-1)\Delta}{2R}\in\left(\frac{m-1}{2m+1}, \frac{m-1}{2m-1}\right]\subset\left(\frac{1}{5},\frac12\right]
\end{align*}
for $m\ge 2$. If $m=2j+1$, then we have
\begin{align*}
\frac qR = \frac{m\Delta}{2R}\in\left(\frac{m}{2m+1}, \frac{m}{2m-1}\right]\subset\left(\frac{3}{7},\frac35\right]\end{align*}
for $m\ge 3$. Therefore, in all cases with $m>1$, we have
\begin{align*}
\frac15< \frac qR\le \frac35.
\end{align*}
Here, since $|k^\ast - qn/R|\le 1$, we have
\begin{align*}
k^\ast\in \left[\frac{qn}R - 1, \frac{qn}R + 1\right]\subset\left(\frac n5-1, \frac{3n}5+1\right].
\end{align*}
In particular, for $n\ge 6$, it holds that $n/30< k^\ast\le 23n/30$. 

Hence, if we choose $c_1 = 1/30$, $c_2 = \max\{1/4+3/(8c), 23/30\}$, then there exists $k^\ast$ such that $c_1n< k^\ast\le c_2n$ and \cref{eq:pflem:lb-ge-drgd-binomial} holds. 
Lastly, we show that $p = k^\ast/(n-1)$ satisfies $p(1-p) = \Omega(1)$. From
\begin{align*}
\frac{c_1n}{n-1}< p\le \frac{c_2n}{n-1},
\end{align*}
we have $c_1\le p$ since $1\le n/(n-1)$. Since $n\ge \max\{6, 8c/(2c-1)\}$, it holds that $n/(n-1)\le \min\{6/5, 8c/(6c+1)\}$, yielding
\begin{align*}
p\le \frac{c_2n}{n-1}\le \max\left\{\frac14+\frac{3}{8c}, \frac{23}{30}\right\}\times \min\left\{\frac65, \frac{8c}{6c+1}\right\}< 1
\end{align*}
where the last inequality follows from 
\begin{align*}
\frac{23}{30}\times \frac65 = \frac{23}{25}<1, \quad\left(\frac14+\frac{3}{8c}\right)\times\frac{8c}{6c+1} = \frac{2c+3}{6c+1}<1
\end{align*}
for $c>1/2$.

Since $c$ is a universal constant, $p(1-p) = \Omega(1)$ and this completes the proof.

\subsubsection{Proof of \cref{lem:binom-anti-concentrate}}\label[appendix]{sec:pflem:binom-anti-concentrate}
In this section, we prove \cref{lem:binom-anti-concentrate}. We first directly compute the probability. Let $q = 1-p$.
\begin{align*}
\bbP(X=np) &= \binom{n}{np}p^{np}q^{nq}\\
&=\frac{n!}{(np)!(nq)!}p^{np}q^{nq}
\end{align*}

Here, we note that from Stirling's approximation (\citet{robbins55}), for any $m\in\bbN$, we have
\begin{align*}
\sqrt{2\pi m}\left(\frac{m}{e}\right)^m<m!<\sqrt{2\pi m}\left(\frac{m}{e}\right)^me^{\frac{1}{12m}}.
\end{align*}
Applying this inequality, we have the lower bound of $\bbP(X=np)$:
\begin{align*}
\bbP(X=np) &=\frac{n!}{(np)!(nq)!}p^{np}q^{nq}\\&\ge \frac{\sqrt{2\pi n}(n/e)^n}{2\pi n \sqrt{pq}(np/e)^{np}(nq/e)^{nq}e^{\frac{1}{12np}+\frac{1}{12nq}}}p^{np}q^{nq}\\
&=\frac{e^{-\frac{1}{12npq}}}{\sqrt{2\pi  pq}}\frac{1}{\sqrt{n}}.
\end{align*}
Let $k = np \in\{1, \cdots, n-1\}$. Then, we have
\begin{align*}
npq = k\left(1- \frac kn\right)\ge 1-\frac 1n\ge \frac12
\end{align*}
for $n\ge 2$. Then, $\exp\left(-1/(12npq)\right)\ge \exp\left(- 1/6\right)=\Omega(1)$. Furthermore, since $pq = p(1-p)\le 1/4$, we have $1/\sqrt{2\pi pq} = \Omega(1)$. 
Hence, we have $\bbP(X=np)=\Omega\left(1/\sqrt{n}\right)$. This completes the proof.

\newpage

\newpage
\subsection{Proof of \cref{thm:lb-us-drgd}}\label[appendix]{sec:pfthm:lb-us-drgd}
In this section, we prove \cref{thm:lb-us-drgd}. Here, we use the same loss function $f:\bbR\times\{0,1\}\to\bbR$ as in the proof of \cref{thm:lb-ge-drgd}:
\begin{align*}
f(w;z) = -z Lw.
\end{align*}
Since $\eta L>0.5\Delta$, the sequence $\{\eta Lk/n\}_{k=0}^n$ crosses the threshold $0.5\Delta$. Thus,
there exists $k^\ast\in\{0, 1, \cdots, n-1\}$ such that
\begin{align*}
\dr\left(-\frac{\eta L k^\ast}{n}\right) - \dr\left(-\frac{\eta L(k^\ast+1)}{n}\right)\ge \Delta.
\end{align*}
We set the neighboring two datasets $S$ and $S'$ such that $\|S\|_1=k^\ast$ and $\|S'\|_1=k^\ast+1$.

Let $(w_t)_{t=1}^T$ and $(w_t')_{t=1}^T$ be the $\dr$-GD iterates on $f$ over $S$ and $S'$, respectively. Then, we have
\begin{align*}
w_T'  - w_T= T\left(\dr\left( -\frac{\eta Lk^\ast}{n}\right) -\dr\left( -\frac{\eta L(k^\ast+1)}{n}\right)\right)\ge T\Delta.
\end{align*}

Furthermore, by \cref{lem:lb-general-drgd}, we have 
\begin{align*}
    \dr\left( -\frac{\eta Lk^\ast}{n}\right) -\dr\left( -\frac{\eta L(k^\ast+1)}{n}\right)\ge \frac{\eta L}{n}-\Delta
\end{align*}
and hence 
\begin{align*}
w_T'- w_T\ge T\left(\frac{\eta L}{n}-\Delta\right).
\end{align*}
Then, we have 
\begin{align*}
f(w_T;1) - f(w_T';1)= L(w_T' -w_T)\ge L\max\left\{\Delta T,T\left(\frac{\eta L}{n}-\Delta\right)\right\}
\end{align*}
which completes the proof.

\clearpage
\subsection{Proof of \cref{thm:ub-us-drsgd-1dim}}\label[appendix]{sec:pfthm:ub-us-drsgd-1dim}
Let $S= (z_1, \cdots, z_n)$ and $S'=(z_1', \cdots, z_n')\in\mcZ^n$ be the neighboring datasets, $\delta_t = |w_t - w_t'|$ where $(w_t)_{t=1}^T$ and $(w_t')_{t=1}^T$ be the coupled iterates from $A_{f,\theta}^{\drsgd}(S)$ and $A_{f,\theta}^{\drsgd}(S')$, respectively.
Since $\varepsilon_{us}(A_{f,\theta}^\drsgd)\le L\cdot \varepsilon_{uas}(A_{f,\theta}^\drsgd)$, it suffices to show that $\bbE[\delta_T] \le (2\eta L+\Delta)T/n$. By the law of total expectation, we have
\begin{align*}
\bbE[\delta_{t}] = \frac{n-1}{n}\bbE[\delta_{t}|z_{i_{t}} = z_{i_t}']+ \frac{1}{n}\bbE[\delta_{t}|z_{i_{t}} \neq z_{i_t}'].
\end{align*}

We first consider the upper bound of $\bbE[\delta_{t}|z_{i_t} = z_{i_t}']$. To this end, we use the following lemma:
\begin{lemma}\label{lem:dr-non-expansiveness}
Let $f:\bbR\to\bbR$ be a convex, $M$-smooth function. Let $x,y\in\grid$ and $\eta>0$ with $\eta<2/M$. Then, it holds that
\begin{align*}
|x-y-(\dr(\eta\nabla f(x)) - \dr(\eta\nabla f(y)))|\le |x-y|.
\end{align*}
\end{lemma}
\cref{lem:dr-non-expansiveness} implies that non-expansiveness holds for $1$-dimensional $\drsgd$. See \cref{sec:pflem:dr-non-expansiveness} for the proof of \cref{lem:dr-non-expansiveness}. By applying \cref{lem:dr-non-expansiveness}, we have $\bbE[\delta_{t}|z_{i_t} = z_{i_t}']\le \bbE[\delta_{t-1}]$.

We now show the upper bound of $\bbE[\delta_{t}|z_{i_t}\ne z_{i_t}']$. Conditioned on $z_{i_t}\ne z_{i_t}'$, we have
\begin{align*}
\delta_{t} &\le |w_{t-1}- w_{t-1}'| + |\dr(\eta \nabla f(w_{t-1};z_{i_t})) - \dr (\eta \nabla f(w_{t-1}';z_{i_t}')|\\
&\le \delta_{t-1}+2\eta L+\Delta
\end{align*}
where the first inequality follows from the triangle inequality, and the second inequality holds since $\dr(x)\in[x-\Delta/2, x+\Delta/2]$ for all $x\in\bbR$, and $|\eta \nabla f(x;z_{i_t}) - \eta \nabla f(y;z_{i_t}')|\le 2\eta L$ for all $x,y\in\bbR$.

Conclusively, we have
\begin{align*}
\bbE[\delta_T]\le \bbE[\delta_{T-1}]+ \frac{2\eta L+\Delta}{n}\le \frac{(2\eta L+\Delta)T}{n}
\end{align*}
which completes the proof of the upper bound.

\subsubsection{Proof of \cref{lem:dr-non-expansiveness}}\label[appendix]{sec:pflem:dr-non-expansiveness}
For notational simplicity, we set $u=\eta\nabla f(x)$ and $v=\eta\nabla f(y)$. Without loss of generality, we assume that $x> y$; $x=y$ is trivial. Since $x,y\in\grid$, there exists $k\in\bbN$ such that $x-y=k\Delta$. Then, $|x-y-(\dr(u)-\dr(v))|\le |x-y|$ is equivalent to $\dr(u)-\dr(v)\in [0,2k\Delta]$. 

Since $u\ge v$ and $\dr$ is monotonically increasing, $\dr(u)-\dr(v)\ge 0$. Furthermore, 
since $\dr(x) \in[x-0.5\Delta, x+0.5\Delta]$ for any $x\in\bbR$, we have
\begin{align*}
\dr(u)-\dr(v) \le u +0.5\Delta - (v-0.5\Delta) = u-v+\Delta.
\end{align*}
Here, since $f$ is convex, $f'$ is monotonically increasing. Furthermore, since $f$ is $M$-smooth, and $\eta <2/M$, we have
\begin{align*}
0\le u-v\le \eta M(x-y)<2k\Delta
\end{align*}
which implies $0\le \dr(u)-\dr(v)<(2k+1)\Delta$. Since $\dr(u),\dr(v)\in\grid$, we have $\dr(u)-\dr(v)\in[0, 2k\Delta]$.
This completes the proof.
\clearpage

\subsection{Proof of \cref{thm:lb-us-drsgd-1dim}}\label[appendix]{sec:pfthm:lb-us-drsgd-1dim}
In this section, we prove \cref{thm:lb-us-drsgd-1dim} by considering the following loss function $f:\bbR\times\{0,1\}\to\bbR$ and neighboring datasets $S$ and $S'$:
\begin{align*}
f(x;z) = -zLx, \quad S = (0, 0, \cdots, 0), \quad S' = (1,0,\cdots, 0).
\end{align*}
We consider two coupled $\drsgd$ iterates $(w_t)_{t=1}^T$ and $(w_t')_{t=1}^T$ from $A_{f,\theta}^{\drsgd}(S)$ and $A_{f,\theta}^\drsgd(S')$, respectively. Then, it is easy to observe that $w_t=0$ for all $t\in[T]$. Thus, we have
\begin{align*}
\bbE[f(w_T;1) - f(w_T';1)] = -\bbE[f(w_T';1)] = L\bbE[w_T'].
\end{align*}
Thus, it suffices to show that $\bbE[w_T'] = \Omega((2\eta L-\Delta)T/n)$.

We note that conditioned on $w_1', \cdots, w_{T-1}'$, it holds that $\bbE[w_T'|i_{T}\ne 1]=w_{T-1}'$, and
\begin{align*}
\bbE[w_T'|i_{T}=1] &= w_{T-1}' - \dr(-\eta L)\\
&\ge w_{T-1}' +\eta L-\frac\Delta2
\end{align*}
where the inequality in the second line follows from the fact that $\dr(x)\in [x-\Delta/2, x+\Delta/2]$ for all $x\in\bbR$. Then, we have
\begin{align*}
\bbE[w_T'] &= \frac{1}{n}\bbE[w_T'|i_{T}=1]+\left(1-\frac1n\right)\bbE[w_{T}'|i_{T}\ne 1]\\
&\ge\frac1n\left(\bbE\left[w_{T-1}'\right]+\eta L-\frac\Delta2\right)+\left(1-\frac1n\right)\bbE[w_{T-1}']\\
&=\bbE[w_{T-1}']+\frac{2\eta L-\Delta}{2n}\ge \cdots\ge \frac{(2 \eta L-\Delta)T}{2n}
\end{align*}
which completes the proof.

\clearpage
\subsection{Proof of \cref{thm:lb-us-drsgd}}\label[appendix]{sec:pfthm:lb-us-drsgd}
In this section, we prove \cref{thm:lb-us-drsgd}. Here, we construct a counterexample function $f:\bbR^2\times\{0,1\}\to\bbR$ that achieves a desired uniform stability bound in \cref{thm:lb-us-drsgd}. We note that it can easily be extended to general $d>2$ by considering a function $g(x_1, \cdots, x_d; z) = f(x_1, x_2;z)$.

To this end, we construct a loss function $f:\bbR^2\times\{0,1\}\to\bbR$ by
\begin{align*}
f(x,y; 0) = g(x,y) , \quad f(x,y; 1) = \Delta y
\end{align*}
where $g$ is a function obtained from the following lemma:
\begin{lemma}\label{lem:piecewise-linear}
Let $\Delta>0$ and $T\in\bbN$. Then, there exists $\alpha\in(0,\Delta/2)$ and a convex, $\Delta$-Lipschitz, and $\left(\frac{9}{10T^2}+\frac{8}{5}\right)$-smooth function $g:\bbR^2\to\bbR$ such that
\begin{align*}
g(0,0)=0, ~ \dr(\nabla g(0,0))=(0,0),~
g(x,y) = -\left(\frac{\Delta}{2}+\alpha\right)x-\frac{\Delta^2}{20}, ~\dr(\nabla g(x,y))=(-\Delta,0)
\end{align*}
for all $x,y$ such that $x\in\{0, \Delta, 2\Delta, \cdots, T\Delta\}$ and $y\le -\Delta$ .
\end{lemma}
\cref{lem:piecewise-linear} implies that there exists a convex and smooth function $g:\bbR^2\to\bbR$ whose rounded gradient is $(0,0)$ at the origin but consistently induces $(-\Delta, 0)$ gradient on the region $\{(x,y):x\in\{0, \cdots, T\Delta\}, y\le -\Delta\}$. The proof of \cref{lem:piecewise-linear} is in \cref{sec:pflem:piecewise-linear}.

Let $S = (0, \cdots, 0)$ and $S' = (1,0, \cdots, 0)$ be the neighboring datasets, $\{w_t\}_{t=1}^T$ and $\{w_t'\}_{t=1}^T$ be trajectories from $A_{f,\theta}^{\drsgd}(S)$ and $A_{f,\theta}^\drsgd(S')$, respectively, where $w_t=(w_{t,1}, w_{t,2})$ and $w_t'= (w_{t,1}', w_{t,2}')$ for each $t\in[T]$. Then since $S=(0,\cdots,0)$ and $\dr(\nabla f(w_0;0))=(0,0)$, we have $w_t=(0,0)$ for all $t\in[T]$.

Then, we have
\begin{align*}
\bbE[f(w_T; 0) - f(w_T';0)]&= \bbE[g(w_T) - g(w_T')]=-\bbE[g(w_T')].
\end{align*}
Here, since $\dr(\nabla f(w_0;0))=(0,0)$, we have $w_t'=(0,0)$ for all $t< \tau$ where $\tau := \min\{t:i_t=1\}$. When $t=\tau$, i.e., $i_t=1$ is observed, we have $w_{\tau}' = (0, -\Delta)$ since $\dr(\eta\nabla f(x,y;1))=(0,\Delta)$. After $\tau$, whenever $i_t\ne 1$, the update uses $g$, and hence $w_t' = w_{t-1}'+(\Delta,0)$. Whenever $i_t=1$, the update only decreases the second coordinate by $\Delta$, so the iterate remains in the region $y\le -\Delta$. 
Thus, we have
\begin{align*}
\bbE[f(w_T; 0) - f(w_T';0)]= -\bbE[g(w_T')] \ge \left(\frac\Delta 2+\alpha\right)\bbE[w_{T,1}'] = \Omega(\Delta\bbE[w_{T,1}']).
\end{align*}

Let $I_t$ be an indicator variable such that $I_t=1$ if $i_t\ne1$ and $t>\tau$, i.e., the event that $w_t' = w_{t-1}'+(\Delta,0)$ occurs.
Then, we have
\begin{align*}
\bbP(I_t=1) = \left(1 - \left(1-\frac{1}{n}\right)^{t-1}\right)\left(1 - \frac1n\right).
\end{align*}
Let $q = 1-1/n$. Then, we have
\begin{align*}
\bbE[w_{T,1}']& = \Delta\sum_{t=1}^{T}q(1-q^{t-1}) \\&=  \Delta q\left(T - \frac{1-q^T}{1-q}\right)\\&=\Delta\left(1- \frac1n\right)\left(T - n\left(1-\left(1-\frac1n\right)^T\right)\right).
\end{align*}
We complete the proof using the following technical lemma. The proof of \cref{lem:lb-us-drsgd} is in \cref{sec:pflem:lb-us-drsgd}.
\begin{lemma}\label{lem:lb-us-drsgd}
Let $n\in\bbN$ and $T\in\bbN$ with $n\ge2, T\ge 2$. Then, it holds that 
\begin{align*}
\left(1- \frac1n\right)\left(T - n\left(1-\left(1-\frac1n\right)^T\right)\right) = \Omega\left(\min\left\{1, \frac{T}{n}\right\}T\right).
\end{align*}
\end{lemma}
By applying \cref{lem:lb-us-drsgd}, we have
\begin{align*}
\bbE[w_{T,1}' ]= \Omega\left(\min\left\{1, \frac{T}{n}\right\}\Delta T\right).
\end{align*}
Conclusively, we have
\begin{align*}
\bbE[f(w_T; 0) - f(w_T';0)] = -\bbE[g(w_T')]= \Omega(\Delta\bbE[w_{T,1}']) = \Omega\left(\min\left\{1, \frac Tn\right\}\Delta^2 T\right)
\end{align*}
which completes the proof.

\subsubsection{Proof of \cref{lem:piecewise-linear}}\label[appendix]{sec:pflem:piecewise-linear}
In this section, we prove \cref{lem:piecewise-linear}. To this end, we first consider a smoothed version of a piecewise-linear function $\max\{L_1,L_2\}$ that satisfies the gradient conditions in \cref{lem:piecewise-linear}. 

We first consider a smoothing function: for $\varepsilon>0$, $H_\varepsilon:\bbR\to\bbR$ is defined by
\begin{align*}
H_\varepsilon(x) = \begin{cases}
0 & \text{if } x \le -\varepsilon, \\
\frac{(x+\varepsilon)^2}{4\varepsilon} & \text{if } -\varepsilon < x < \varepsilon, \\
x & \text{if } x \ge \varepsilon.
\end{cases}
\end{align*}
One can observe that $H_\varepsilon(x)$ can be regarded as a smoothed version of $\max\{0, x\}$ and $H_\varepsilon$ is convex. %

We define a loss function $g:\mathbb{R}^{2} \to \mathbb{R}$ by
\begin{align*}
g(x,y) = L_1(x,y) + H_{\varepsilon}\left(L_2(x,y) - L_{1}(x,y)\right)
\end{align*}
where
\begin{align*}
L_{1}(x,y)= -\left(\frac{\Delta}{2} - \frac{\Delta}{10T}\right)x+\frac{2\Delta}{5}y,\quad L_{2}(x,y) = -\left( \frac{\Delta}{2}+\frac{\Delta}{5T} \right)x-\frac{\Delta^2}{20}
\end{align*}
and $\varepsilon>0$. We note that since $L_2-L_1$ is affine and  $H_\varepsilon$ is convex, $H_\varepsilon\circ (L_2 - L_1)$ is convex; adding the affine function $L_1$ preserves convexity. 
We now assign an explicit value to $\varepsilon$ to satisfy the gradient conditions of \cref{lem:piecewise-linear}. To this end, we determine $\varepsilon>0$ so that $\nabla g(0,0) = \nabla L_1(0,0)$, and $\nabla g(x, y) = \nabla L_2(x,y)$ for all $x\in \{0, \Delta, 2\Delta, \cdots, T\Delta\}$ and $y\le -\Delta$. Then, we have
\begin{align*}
\nabla g(0,0) = \nabla L_1(0,0) = \left(-\frac\Delta 2+\frac{\Delta}{10T}, \frac{2\Delta} 5\right), \quad \nabla g(x, y) = \nabla L_2(x, y) = \left(-\frac\Delta 2-\frac{\Delta}{5T},0\right)
\end{align*}
for each $x\in\{0, \Delta, \cdots, T\Delta\}$ and $y\le -\Delta$.

We first derive the condition for $\varepsilon>0$ such that $\nabla g(0,0)=\nabla L_1(0,0)$. Since $g(x,y) = L_1(x,y)$ if $H_\varepsilon(L_2(x,y) - L_1(x,y))=0$, it suffices to consider $\varepsilon>0$ such that $L_2(0,0)-L_1(0,0) \le -\varepsilon$. Thus, we obtain the first requirement of $\varepsilon$:
\begin{align*}
L_2(0,0)-L_1(0,0) = -\frac{\Delta^2}{20}\le -\varepsilon\iff \varepsilon\le \frac{\Delta^2}{20}.
\end{align*}

We now consider $\nabla g(x,y) = \nabla L_2(x,y)$ such that $x\in\{0,\Delta,  \cdots, T\Delta\}$ and $y\le -\Delta$. Since $g(x,y) = L_2(x,y)$ if $H_\varepsilon(L_2(x,y) - L_1(x,y)) = L_2(x,y) - L_1(x,y)$, we consider $\varepsilon>0$ satisfying $L_2(x,y) - L_1(x,y)\ge \varepsilon$. Here, since
\begin{align*}
L_2(x,y) - L_1(x,y) = -\frac{3\Delta}{10T}x - \frac{2\Delta}{5}y - \frac{\Delta^2}{20}
\end{align*}
is minimized at $(T\Delta, -\Delta)$, it suffices to derive the condition for $\varepsilon$ such that $L_2(T\Delta,-\Delta) - L_1(T\Delta, -\Delta)\ge \varepsilon$ for $x\in\{0, \cdots, T\Delta\}$. We have
\begin{align*}
L_2(x, -\Delta) - L_1(x,-\Delta)  &= \left(-\left( \frac{\Delta}{2}+\frac{\Delta}{5T} \right)x-\frac{\Delta^2}{20}\right)-\left(-\left(\frac{\Delta}{2} - \frac{\Delta}{10T}\right)x-\frac{2\Delta^2}{5}\right)\\
&=\frac{2\Delta^2}{5}-\frac{3\Delta}{10T}x - \frac{\Delta^2}{20}\\
&=\frac{7\Delta^2}{20} - \frac{3\Delta x}{10T}.
\end{align*}
The above expression has a minimum value at $x=T\Delta$:
\begin{align*}
L_2(x, -\Delta) - L_1(x,-\Delta)\ge \frac{7\Delta^2}{20} - \frac{3\Delta^2}{10}=\frac{\Delta^2}{20}
\end{align*}
Thus, we need $\varepsilon\le \Delta^2/20$. Thus, if we choose $\varepsilon = \Delta^2/20$ then our $g$ satisfies the gradient conditions at the origin, and at $(w,-\Delta)$ for each $w\in\{0, \cdots, T\Delta\}$. 

We now evaluate the Lipschitzness and smoothness constant of $g$. Let $\varepsilon_0 = \Delta^2/20$ and $L_0(x,y) = L_2(x,y) - L_1(x,y)$. Then, we have
\begin{align*}
\nabla g(x,y) = \nabla L_1(x,y) + H_\varepsilon'(L_0(x,y))\nabla L_0(x,y).
\end{align*}
Since $0\le H'_\varepsilon(x)\le 1$, one can observe that $\nabla g$ is a convex combination of $\nabla L_1$ and $\nabla L_2$. Here, since $\|\nabla L_1\|_2 = \|(\Delta/2-\Delta/(10T), 2\Delta/5)\|_2\le \Delta$ and $\|\nabla L_2\|_2 = \|(\Delta/2+\Delta/(5T), 0)\|_2\le \Delta$ for all $T\ge 1$, $g$ is $\Delta$-Lipschitz. 

Furthermore, since $H_{\varepsilon_0}'$ is $ 1/(2\varepsilon_0)$-Lipschitz, $\nabla g(x,y)$ is Lipschitz with constant at most
\begin{align*}
\frac{1}{2\varepsilon_0}\|\nabla L_0(x,y)\|^2_2 = \frac{1}{2\varepsilon_0}\left(\frac{9\Delta^2}{100T^2}+\frac{4\Delta^2}{25}\right)= \frac{9}{10T^2}+\frac{8}{5}
\end{align*}
which implies that our $g$ is $(9/(10T^2)+8/5)$-smooth. This completes the proof with the choice of $\alpha=\Delta/(5T)$.

\subsubsection{Proof of \cref{lem:lb-us-drsgd}}\label[appendix]{sec:pflem:lb-us-drsgd}
In this section, we prove \cref{lem:lb-us-drsgd}. To this end, we show the lower bound for the following two cases: (1) $T\ge n$, and (2) $T< n$. For notational simplicity, we denote $E(n,T)$ by the given expression in \cref{lem:lb-us-drsgd}: $E(n,T) = \left(1- \frac1n\right)\left(T - n\left(1-\left(1-\frac1n\right)^T\right)\right)$. It suffices to show that
\begin{align*}
E(n,T) = \Omega\left(\min\left\{1, \frac{T}{n}\right\}T\right).
\end{align*}
For $n\ge 2$, $1-1/n\ge 1/2$ holds. Thus, we have
\begin{align}
E(n,T)&\ge \frac12\left( T - n\left(1-\left(1-\frac 1n\right)^T\right)\right)\label{eq:pflem:lb-us-drsgd}
\end{align}

We first show that if $T\ge n$, then $E(n,T) = \Omega(T)$.
Here, we consider the following two sub-cases: $T\ge 2n$ and $n\le T\le 2n$.

The case where $T\ge 2n$ is easily obtained since 
\begin{align*}
T - n\left(1-\left(1-\frac1n\right)^T\right)\ge T-n\ge \frac T2.
\end{align*}

We now suppose that  $n\le T \le 2n$. Then, we have
\begin{align*}
\left(1-\frac1n\right)^T\ge \left(1-\frac1n\right)^{2n}\ge \frac 1{16}
\end{align*}
since $\left(1-1/n\right)^n$ is strictly increasing for $n\ge 2$. Thus, we have
\begin{align*}
T - n\left(1-\left(1-\frac1n\right)^T\right)\ge T - \frac{15n}{16}\ge \frac T{16}.
\end{align*}
By combining these results with \cref{eq:pflem:lb-us-drsgd}, we have $E(n,T) = \Omega(T)$ if $T\ge n$.

We now show that if $T<n$, then $E(n,T) = \Omega(T^2/n)$. To this end, we consider the binomial expansion of $(1-1/n)^T$:
\begin{align*}
\left(1 - \frac1n\right)^T =\sum_{i=0}^T {T\choose i}\left(-\frac1n\right)^i
\end{align*}
Here, since $T<n$, the magnitude of the terms ${T\choose i}\left(-\frac1n\right)^i$ is monotonically decreasing: 
\begin{align*}
\frac{{T\choose i+1}n^{-(i+1)}}{{T\choose i}n^{-i}} = \frac{T-i}{(i+1)n}\le \frac Tn<1
\end{align*}

Then, truncating the series at the negative term gives a lower bound of the binomial expansion:
\begin{align*}
\left(1 - \frac1n\right)^T \ge 1-\frac Tn+\frac{T(T-1)}{2n^2}-\frac{T(T-1)(T-2)}{6n^3}.
\end{align*}
Then, we have
\begin{align*}
T - n\left(1-\left(1-\frac1n\right)^T\right)&\ge T - n\left(1-\left(1-\frac Tn+\frac{T(T-1)}{2n^2}-\frac{T(T-1)(T-2)}{6n^3}.\right)\right)\\
&=\frac{T(T-1)}{2n}- \frac{T(T-1)(T-2)}{6n^2}\\
&=\frac{T(T-1)}{2n}\left(1-\frac{T-2}{3n}\right) 
\end{align*}
Since $T<n$, we have $1-\frac{T-2}{3n}\ge 1-\frac T{3n}\ge \frac23$.
Furthermore, if $T\ge 2$, $T(T-1)\ge T^2/2$. Then, by combining this result with \cref{eq:pflem:lb-us-drsgd}, we obtain that if $T<n$,
\begin{align*}
E(n,T) \ge \frac12\left(\frac{T^2}{6n}\right) = \Omega\left(\frac{T^2}n\right).
\end{align*}
Conclusively, we have
\begin{align*}
E(n,T) = \begin{cases}
\Omega(T)&T\ge n\\\Omega(T^2/n)&T<n
\end{cases}=\Omega\left(\min\left\{1, \frac Tn\right\}T\right),
\end{align*}
which completes the proof.

\clearpage
\subsection{Proof of \cref{thm:ub-us-drsgd}}\label[appendix]{sec:pfthm:ub-us-drsgd}
In this section, we prove \cref{thm:ub-us-drsgd}. Let $S,S'\in\mcZ^n$ be the neighboring datasets, and $(w_t)_{t=1}^T$ and $(w_t')_{t=1}^T$ be the coupled $\drsgd$ iterates from $A^{\drsgd}_{f,\theta}(S)$ and $A_{f,\theta}^{\drsgd}(S')$, respectively. Let $\delta_t = w_t - w_t'$ for each $t\in[T]$.

For each $t\in[T]$, let $B_t$ be the event that there exists $\tau\le t$ such that $z_{i_\tau}\ne z_{i_\tau}'$ and $A_t$ be the event that $z_{i_t}\ne z_{i_t}'$. Since $w_T=w_T'$ conditioned on $B_{T}^c$, we have $\bbE[\|\delta_T\|_2] = P(B_{T})\bbE[\|\delta_T\|_2|B_{T}]$.

Here, we have 
\begin{align}
P(B_{T}) = 1- \left(1- \frac{1}{n}\right)^T\le \min\left\{1,\frac{T}{n}\right\}.\label{eq:pfthm:ub-uas-srsgd-BT}
\end{align}

In the rest of the proof, we show the upper bound of $\bbE[\|\delta_T\|_2|B_{T}]$ by constructing a recurrence of $\bbE[\|\delta_t\|_2|B_{t}]$. For $t=1$, we have
\begin{align*}
\bbE[\|\delta_1\|_2|B_1] = \bbE[\|\delta_1\|_2|A_1] &= \bbE[\|\dr(\eta\nabla f(w_0; z_{i_1})) - \dr(\eta \nabla f(w_0; z_{i_1}'))\|_2|z_{i_1}\ne z_{i_1}']\\
&\le \|\dr(\eta\nabla f(w_0; z_{i_1}))\|_2+\|\dr(\eta\nabla f(w_0; z_{i_1}'))\|_2\le 4\eta L
\end{align*}
where the second inequality follows from the triangle inequality, and the last inequality follows from \cref{lem:ub-dr}.

We now consider general $t\ge 2$. Here, $B_t$ can be regarded as the union of two mutually exclusive events, $B_{t-1}^c\cap A_{t}$ and $B_{t-1}$.  Thus, we can decompose $\bbE[\|\delta_{t}\|_2|B_t]$ as follows:
\begin{align*}
\mathbb{E}[\lVert \delta_{t} \rVert_2|B_{t}] &= \frac{P(B_{t-1}^{c} \cap A_{t})}{P(B_{t})}\bbE[\lVert \delta_{t} \rVert_2 |B_{t-1}^{c} \cap A_{t}]+\frac{P(B_{t-1})}{P(B_{t})}\bbE[\lVert \delta_{t} \rVert_2 |B_{t-1}]\\ 
&\leq \bbE[\|\delta_{t}\|_2|B_{t-1}^c\cap A_t]+\bbE[\|\delta_{t}\|_2|B_{t-1}].
\end{align*}
We now show the upper bounds for each term. Conditioned on $w_1, \cdots, w_{t-1}$ and $w_1', \cdots, w_{t-1}'$, we have
\begin{align*}
\|\delta_{t}\|_2 &= \|\delta_{t-1} - (\dr(\eta \nabla f(w_{t-1}; z_{i_t})) - \dr (\eta \nabla f(w_{t-1}'; z_{i_t}')))\|_2\\
&\le \|\delta_{t-1}\|_2+ \|\dr(\eta \nabla f(w_{t-1}; z_{i_t}))\|_2+\|\dr(\eta\nabla f(w_{t-1}'; z_{i_t}'))\|_2\\
&\le \|\delta_{t-1}\|_2+4\eta L
\end{align*}
where the last inequality follows from \cref{lem:ub-dr} and $L$-Lipschitzness of $f$. Then, we have
\begin{align*}
&\bbE[\|\delta_{t}\|_2|B_{t-1}^c\cap A_t]\le 4\eta L,\\
&\bbE[\|\delta_{t}\|_2|B_{t-1}]\le \bbE[\|\delta_{t-1}\|_2|B_{t-1}]+4\eta L.
\end{align*}
Then, we have
\begin{align*}
\bbE[\|\delta_{T}\|_2|B_{T}]\le \bbE[\|\delta_{T-1}\|_2|B_{T-1}]+8\eta L\le \cdots\le 8\eta LT.
\end{align*}
Conclusively, we have
\begin{align*}
\varepsilon_{us}(A^\drsgd_{f,\theta})\le L\bbE[\|\delta_T\|_2]\le L\bbE[\|\delta_T\|_2|B_{T}]P(B_{T})\le \min\left\{1, \frac Tn\right\}8\eta L^2T
\end{align*}
which completes the proof.
\newpage
\section{Proofs for $\srgd$ and $\srsgd$}

In this section, we prove generalization and stability bounds of $\srgd$ and $\srsgd$. We first prove \cref{thm:lb-ge-srgd} in \cref{sec:pfthm:lb-ge-srgd}. We then show the upper bounds of uniform argument stability for $\srsgd$ and $\srgd$ (\cref{thm:ub-uas-srsgd-general,thm:ub-uas-srgd-general}) in \cref{sec:pfthm:ub-uas-srsgd-general,sec:pfthm:ub-uas-srgd-general}, respectively. We then present the proofs of uniform argument stability bounds for separable loss functions (\cref{thm:ub-uas-srgd-separable,thm:lb-uas-srgd}) in \cref{sec:pfthm:ub-uas-srgd-separable,sec:pfthm:lb-uas-srgd}, respectively.

\subsection{Proof of \cref{thm:lb-ge-srgd}}\label[appendix]{sec:pfthm:lb-ge-srgd}
In this section, we prove \cref{thm:lb-ge-srgd}. %
In this proof, we use $\mcD=Ber(1/2)$ as a data distribution and $f:\bbR^d\times\{0,1\}\to\bbR$ as a loss function defined as follows: $f(w;0)=0$ and $f(w;1)=g(w)$ where $g:\bbR^d\to\bbR$ is obtained from the following lemma.
\begin{lemma}\label{lem:huber}
Let $d\in\bbN$, $L,\Delta>0$. Then, there exists $\alpha>0$ and $\gamma>0$ such that $\alpha >(e^2-1)\gamma$, and a function $g:\bbR^d\to\bbR$ such that $g$ is convex, $L$-Lipschitz, $4L/\Delta$-smooth, and satisfies the following:
\begin{align*}
\nabla g(\zr)=-\frac{L}{\sqrt{d}}\mathbf{1}_d,\quad g(c)= L\Delta \left(\sqrt{ \alpha k_c+\gamma } - \frac{\sqrt{\gamma}}{2}\right), \quad\forall c\in\{0, \Delta\}^d
\end{align*}
where $k_c$ denotes the number of non-zero entries in $c$, i.e., $k_c=\|c\|_1/\Delta$.
\end{lemma}
\cref{lem:huber} implies the existence of a function $g$ such that $\nabla g(\zr)$ is proportional to the all-ones vector, and $g(c)$ scales linearly with $\|c\|_2$ for each $c\in\{0, \Delta\}^d$; we note that $\|c\|_2 = \Delta\sqrt{k_c}$. The proof of \cref{lem:huber} is in \cref{sec:pflem:huber}.

We now prove the lower bound of the generalization error of $\sr$-GD and $\sr$-SGD using our construction of $\mcD$ and $f$. From \cref{lem:average-stability}, the generalization of $A\in \{A_{f,\theta}^\srgd,A_{f,\theta}^\srsgd\}$ can be written as
\begin{align*}
\varepsilon_g(A;\mcD) = |\bbE_{A,S',z_1}[f(A(S);z_1) - f(A(S');z_1)]|
\end{align*}
where $S=(z_1,\cdots, z_n)\in\mcD^n$ and $S' = (z_1',z_2,\cdots, z_n)$  where $z_1', z_1, \cdots, z_n\sim \mcD$ are independent. We note that if $z_1=z_1'$ or $z_1=0$, then $f(A(S);z_1) - f(A(S');z_1)=0$. Thus, only the event $(z_1, z_1') = (1,0)$ contributes to this expectation. Hence, we can rewrite the generalization error by
\begin{align}
\varepsilon_g(A;\mcD) = \frac{1}{4}|\bbE_{A,S}[f(A(S_1);1) - f(A(S_0);1)]| = \frac{1}{4}|\bbE_{A,S}[g(A(S_1)) - g(A(S_0))]|\label{eq:pfthm:lb-ge-srgd}
\end{align}
where $S_1=(1, z_2, \cdots, z_n)$ and $S_0=(0, z_2, \cdots, z_n)$.

We show the lower bound of \cref{eq:pfthm:lb-ge-srgd} for $\sr$-GD and $\sr$-SGD by applying the following lemmas. 

\begin{lemma}\label{lem:lb-ge-srgd}
Let $\eta,L,\Delta>0$, $n,d\in\bbN$ with $n\ge 14$, $\eta L<\Delta\sqrt{d}$. Let $S_1=(1, z_2, \cdots, z_n)$ and $S_0=(0, z_2, \cdots, z_n)$ where $z_2,\cdots,z_n\sim Ber(1/2)$, $f:\bbR^d\times\{0,1\}\to\bbR$ be a loss function defined by $f(x;0)=0$ and $f(x;1) = g(x)$ where $g$ is obtained from \cref{lem:huber}. Then, for $\theta = (\eta, T=1, w_0 = \zr)$, it holds that
\begin{align*}
|\bbE[g(A_{f,\theta}^{\srgd}(S_1)) - g(A_{f,\theta}^{\srgd}(S_0))]| = \Omega\left(\frac{\sqrt{\min\{\Delta, \eta L d^{1/2}\}\eta L^3 d^{1/2}}}{n} \right).
\end{align*}
\end{lemma}

\begin{lemma}\label{lem:lb-ge-srsgd}
Let $\eta,L,\Delta>0$, $n,d\in\bbN$ with $\eta L<\Delta\sqrt{d}$. Let $S_1=(1, z_2, \cdots, z_n)$ and $S_0=(0, z_2, \cdots, z_n)$ where $z_2,\cdots,z_n\sim Ber(1/2)$ are independent, $f:\bbR^d\times\{0,1\}\to\bbR$ be a loss function defined by $f(x;0)=0$ and $f(x;1) = g(x)$ where $g$ is obtained from \cref{lem:huber}. For $\theta = (\eta, T=1, w_0 = \zr)$, it holds that
\begin{align*}
|\bbE[g(A_{f,\theta}^{\srsgd}(S_1)) - g(A_{f,\theta}^{\srsgd}(S_0))]| = \Omega\left(\frac{\sqrt{\min\{\Delta, \eta L d^{1/2}\}\eta L^3 d^{1/2}}}{n} \right).
\end{align*}
\end{lemma}

The proof of \cref{lem:lb-ge-srgd} is in \cref{sec:pflem:lb-ge-srgd}, and the proof of \cref{lem:lb-ge-srsgd} is in \cref{sec:pflem:lb-ge-srsgd}.
\cref{lem:lb-ge-srgd,lem:lb-ge-srsgd} give us explicit lower bounds on the generalization errors of $\sr$-GD and $\sr$-SGD, respectively. This completes the proof.

\subsubsection{Proof of \cref{lem:huber}}\label[appendix]{sec:pflem:huber}
We construct a Huber function (\citet{huber64}) $g:\bbR^d \to\bbR$ defined by
\begin{align*}
g(x) = \begin{cases}
\frac{2L}{\Delta} \left\lVert x - \frac{\Delta}{4\sqrt{ d }}\mathbf{1}_{d}  \right\rVert_{2}^{2} & x \in \mathcal{B}_{\Delta/4}\left( \frac{\Delta}{4\sqrt{ d }}\mathbf{1}_{d} \right), \\ 
L\left\lVert  x - \frac{\Delta}{4\sqrt{ d }}\mathbf{1}_{d}  \right\rVert_{2}  - \frac{L\Delta}{8} & o.w.
\end{cases}
\end{align*}
Intuitively, $g$ consists of a quadratic part within a ball of radius $\Delta/4$ centered at $(\Delta/(4\sqrt{ d }))\one$, and a linear part outside this ball. It is easy to observe that $g$ is convex, $L$-Lipschitz, and $4L/\Delta$-smooth. Furthermore, we have
\begin{align*}
\nabla g(\zr)=-\frac{L}{\sqrt{d}}\mathbf{1}_d,\quad g(c)= L\Delta \left(\sqrt{ \left( 1- \frac{1}{2\sqrt{ d }} \right)k_c+\frac{1}{16} } - \frac{1}{8}\right), \quad\forall c\in\{0, \Delta\}^d
\end{align*}
where $k_c$ denotes the number of non-zero entries in $c$. We note that if we choose $\alpha = 1- 1/(2\sqrt d)$ and $\gamma = 1/16$, then our $g$ satisfies the conditions for the gradient at the origin and the values at $c\in\{0,\Delta\}^d$ given in \cref{lem:huber}. Furthermore, $\alpha>(e^2-1)\gamma$ holds since 
\begin{align*}
1-\frac{1}{2\sqrt{d}}\ge \frac{1}{2}>\frac{e^2-1}{16}
\end{align*}
for all $d\in\bbN$. This completes the proof.

\subsubsection{Proof of \cref{lem:lb-ge-srgd}}\label[appendix]{sec:pflem:lb-ge-srgd}
In this section, we prove \cref{lem:lb-ge-srgd}. Let $w = A_{f,\theta}^{\srgd}(S_1)$, $w' = A_{f,\theta}^{\srgd}(S_0)$, $k$ be the number of ones in $S_0$, and $K$ be the event that $k\in[(n-1)/3,2(n-1)/3]$. 
Then, we have
\begin{align*}
\bbE[g(w) - g(w')]\ge P(K)\bbE[g(w)-g(w')|K].
\end{align*}

Here, we note that under the coordinate-wise monotone coupling, each coordinate of $w$ is greater than or equal to the corresponding coordinate of $w'$. Since the value of $g(c)$ in \cref{lem:lb-ge-srgd} is increasing in the number of non-zero entries in $c$, we have $g(w) - g(w')\ge 0$ almost surely. Hence, $\bbE[g(w) - g(w')|K^c]\ge 0$ and the inequality above holds.

We first show that $P(K)=\Omega(1)$. By Hoeffding's inequality, it holds that
\begin{align*}
P(K) = \mathbb{P}\left( \left| k- \frac{n-1}{2} \right|  \leq  \frac{n-1}{6}\right) \geq 1- 2\exp\left( - \frac{n-1}{18} \right)
\end{align*}
Here, since $1-2\exp(-(n-1)/18)>0\iff n\ge 1+18\ln2\approx13.48$, and $1-2\exp(-(n-1)/18)$ is an increasing function of $n$, we have $P(K)=\Omega(1)$ for all $n\ge 14$.

Since two parameters are initialized at the origin, $\nabla f(\zr;0)=0$, and $\nabla f(\zr;1)= -(L/\sqrt{d})\mathbf{1}_d$, we have
\begin{align*}
w = -\sr\left(\eta \nabla f(\zr;S_1)\right)  = \sr\left(\frac{\eta L(k+1)}{n\sqrt{d}}\mathbf{1}_d\right),~~~ w' = -\sr(\eta\nabla f(\zr;S_0)) = \sr\left(\frac{\eta Lk}{n\sqrt{d}}\mathbf{1}_d\right).
\end{align*}
Since $\eta L<\Delta\sqrt{d}$, it holds that $w,w'\in\{0, \Delta\}^d$. Let $X$ and $Y$ be random variables that take the number of non-zero entries of $w$ and $w'$, respectively. Then, it is easy to observe that $X\sim Bin\left(d, \frac{\eta L(k+1)}{n\Delta\sqrt{d}}\right), Y\sim Bin\left(d, \frac{\eta Lk}{n\Delta\sqrt{d}}\right)$. Since $g$ is obtained from \cref{lem:huber}, we have
\begin{align}
|\bbE[g(w) - g(w')]| \gtrsim\Delta L\bbE\left[\sqrt{\alpha X+\gamma} - \sqrt{\alpha Y+\gamma}|K\right]\label{eq:pflem:lb-ge-srgd}
\end{align}
for some $\alpha>0$ and $\gamma>0$.
To show the lower bound of \cref{eq:pflem:lb-ge-srgd}, we introduce the following technical lemma. The proof of \cref{lem:lb-ge-bin} is in \cref{sec:pflem:lb-ge-bin}.
\begin{lemma}\label{lem:lb-ge-bin}
Let $a,b>0$ be universal constants such that $a>b(e^2-1)$ , $n\in\bbN$, $p,p'\in(0,1)$ such that $p>p'$, $X\sim Bin(n,p)$ and $Y\sim Bin(n,p')$. Then
\begin{align*}
\bbE[\sqrt{aX+b} - \sqrt{aY+b}]= \Omega\left(\left(1- \frac{p'}{p}\right)\min\left\{\sqrt{\bbE[X]}, \bbE[X]\right\}\right).
\end{align*}
\end{lemma}

Since $X\sim Bin\left(d, \frac{\eta L(k+1)}{n\Delta\sqrt{d}}\right), Y\sim Bin\left(d, \frac{\eta Lk}{n\Delta\sqrt{d}}\right)$, by \cref{lem:lb-ge-bin}, we have
\begin{align*}
\bbE\left[\sqrt{\alpha X+\gamma} - \sqrt{\alpha Y+\gamma}|K\right] &\gtrsim \left(1 - \frac{\frac{\eta Lk}{n\Delta\sqrt d}}{\frac{\eta L (k+1)}{n\Delta \sqrt d}}\right)\min\left\{\sqrt{\frac{\eta L (k+1)d^{1/2}}{n\Delta}}, \frac{\eta L (k+1)\sqrt d}{n\Delta}\right\}\\
&\gtrsim \left(1 - \frac{k}{k+1}\right)\min\left\{\sqrt{\frac{\eta Ld^{1/2}}{\Delta}}, \frac{\eta L \sqrt d}{\Delta}\right\}\\
&\gtrsim\frac{1}{n}\min\left\{\sqrt{\frac{\eta Ld^{1/2}}{\Delta}}, \frac{\eta L \sqrt d}{\Delta}\right\}
\end{align*}
where the second and third inequalities follow under $K$: $k = \Theta(n)$. 

By combining this result with \cref{eq:pflem:lb-ge-srgd}, we have
\begin{align*}
|\bbE[g(w) - g(w')]| \gtrsim \frac{\Delta L}{n}\min\left\{\sqrt{\frac{\eta Ld^{1/2}}{\Delta}}, \frac{\eta L \sqrt d}{\Delta}\right\} = \frac{\sqrt{\min\{\Delta, \eta Ld^{1/2}\}\eta L^3d^{1/2}}}{n}
\end{align*}
which completes the proof.

\subsubsection{Proof of \cref{lem:lb-ge-srsgd}}\label[appendix]{sec:pflem:lb-ge-srsgd}
In this section, we prove \cref{lem:lb-ge-srsgd}. Let $v = A_{f,\theta}^{\srsgd}(S_1)$, $v' = A_{f,\theta}^{\srsgd}(S_0)$, and %
$k$ be the number of $1$ in $S_0$. Since two parameters are initialized at $\zr$ and $\nabla g(\zr)=- (L/\sqrt{d})\mathbf{1}_d$, we have %
\begin{align*}
\bbE[g(v)] &= \bbE\left[\left(1- \frac{k+1}{n} \right)g(\zr) + \frac{k+1}{n}g\left(\sr\left(\frac{\eta L}{\sqrt{d}}\mathbf{1}_d\right)\right)\right],\text{ and}\\
\bbE[g(v')] &= \bbE\left[\left(1- \frac{k}{n} \right)g(\zr) + \frac{k}{n}g\left(\sr\left(\frac{\eta L}{\sqrt{d}}\mathbf{1}_d\right)\right)\right].
\end{align*}
Let $X$ be the number of non-zero entries in $\sr((\eta L/\sqrt{d})\one)$. Then, since $\eta L/\sqrt d<\Delta$, we have $X \sim Bin(d, \eta L/(\sqrt{d} \Delta))$. Since $g$ is obtained from \cref{lem:huber}, we have
\begin{align*}
\bbE[g(v) - g(v')] &= \frac{1}{n} \bbE\left[g\left(\sr\left(\frac{\eta L}{\sqrt{d}}\mathbf{1}_d\right)\right) - g(\zr) \right]\\
&=\frac{L\Delta}{n}\bbE[\sqrt{\alpha X+\gamma} - \sqrt{\gamma}]
\end{align*}
for some $\alpha>0$ and $\gamma>0$.
We now introduce the following lemma to compute the expectation. The proof of \cref{lem:lb-expsqbinom1} is in \cref{sec:pflem:expsqbinom1}.
\begin{lemma}\label{lem:lb-expsqbinom1}
Let $a>0$ and $b\ge 0$ be universal constants such that $a>b(e^2-1)$, $n\in\bbN$, $p\in(0,1)$, and $X\sim Bin(n,p)$. Then it holds that
\begin{align*}
\bbE[\sqrt{aX+b}]-\sqrt{b} =\Omega\left(\min\left\{\sqrt{\bbE[X]}, \bbE[X]\right\}\right).
\end{align*}
\end{lemma}
Then, by \cref{lem:lb-expsqbinom1}, we have
\begin{align*}
\bbE[\sqrt{\alpha X+\gamma} - \sqrt{\gamma}] = \Omega\left(\min\left\{\sqrt{\frac{\eta Ld^{1/2}}{\Delta}} ,\frac{\eta L\sqrt d}{\Delta}\right\}\right).
\end{align*}
Conclusively, we have
\begin{align*}
\bbE[g(v) - g(v')] &=\frac{L\Delta}{n}\bbE[\sqrt{\alpha X+\gamma} - \sqrt{\gamma}] = \Omega\left(\frac{\sqrt{\min\{\Delta, \eta L d^{1/2}\}\eta L^3 d^{1/2}}}{n}\right)
\end{align*}
which completes the proof.

\subsubsection{Proof of \cref{lem:lb-ge-bin}}\label[appendix]{sec:pflem:lb-ge-bin}

In this section, we prove \cref{lem:lb-ge-bin}. Since $p>p'$, we can apply a monotone coupling such that $X\ge Y$ almost surely. 

We first compute $\bbE[Y|X]$. Let $X_1,\cdots, X_n\sim Ber(p)$ and $Y_1,\cdots, Y_n\sim Ber(p')$ be two sets of i.i.d. random variables such that $X = \sum_{i=1}^n X_i$ and $Y =\sum_{i=1}^nY_i$. Since $X$ and $Y$ are monotonically coupled, we have
\begin{align*}
\bbP(Y_i=1|X_i=0)=0, \quad \bbP(Y_i=1|X_i=1) = \frac{p'}{p}.
\end{align*}
Thus, $Y|X = \sum_{i=1}^n Y_i|X_i\sim Bin(X, p'/p)$ and $\bbE[Y|X] = (p'/p)X$.

We now show the lower bound of $\mathbb{E}[\sqrt{ aX+b } - \sqrt{ aY+b } ]$.
\begin{align*}
\mathbb{E}[\sqrt{ aX+b } - \sqrt{ aY+b } ] &= \mathbb{E}\left[ \frac{a(X-Y)}{\sqrt{ aX+b } +\sqrt{ aY+b }}  \right]\\
&\geq \mathbb{E}\left[ \frac{a(X-Y)}{2\sqrt{ aX+b }}  \right]\\
&=\bbE\left[\bbE\left[\frac{a(X-Y)}{2\sqrt{ aX+b }} \bigg| X\right] \right]\\
&=\bbE\left[\frac{a(X-\bbE[Y|X])}{2\sqrt{ aX+b }} \right]\\
&=\frac{a}{2}\left(1- \frac{p'}{p}\right)\bbE\left[\frac{X}{\sqrt{aX+b}}\right].
\end{align*}
It suffices to show that $\bbE\left[X/\sqrt{aX+b}\right] = \Omega(\min\{\sqrt{\bbE[X]}, \bbE[X]\})$.
\begin{align*}
\bbE\left[\frac{X}{\sqrt{aX+b}}\right]&=\mathbb{P}(X\ge 1)\bbE\left[\frac{X}{\sqrt{aX+b}}\bigg|X\ge 1\right]\\
&\ge \bbP(X\ge 1)\bbE\left[\frac{X}{\sqrt{(a+b)X}}\bigg|X\ge 1\right]\\
&\ge \frac{1}{\sqrt{a+b}}\bbP(X\ge 1)\bbE[\sqrt{X}|X\ge 1]\\
&=\frac{1}{\sqrt{a+b}}\bbE[\sqrt{X}] = \Omega\left(\min\left\{\sqrt{\bbE[X]}, \bbE[X]\right\}\right)
\end{align*}
where the last equality follows from \cref{lem:lb-expsqbinom1}. Conclusively, we have 
\begin{align*}
\mathbb{E}[\sqrt{ aX+b } - \sqrt{ aY+b } ] &= \Omega\left(\left(1-\frac{p'}{p}\right)\bbE\left[\frac{X}{\sqrt{aX+b}}\right] \right) \\&= \Omega\left(\left(1-\frac{p'}{p}\right)\min\left\{\sqrt{\bbE[X]}, \bbE[X]\right\}\right)
\end{align*}
which completes the proof.

\subsubsection{Proof of \cref{lem:lb-expsqbinom1}}\label[appendix]{sec:pflem:expsqbinom1}
In this section, we prove \cref{lem:lb-expsqbinom1}. Here, we assume $n\ge2$; the case where $n=1$ follows directly since $X\sim Ber(p)$ and $\bbE[\sqrt{aX+b}]-\sqrt b = p(\sqrt{a+b}-\sqrt b) = \Omega(p)$. To this end, we prove the following two statements: (1) if $\bbE[X]\ge 1$, then $\bbE[\sqrt{aX+b}-\sqrt b] = \Omega(\sqrt{\bbE[X]})$, and (2) if $\bbE[X]< 1$, then $\bbE[\sqrt{aX+b} - \sqrt{b}] = \Omega(\bbE[X])$. Then, we have
$\bbE[\sqrt{aX+b} - \sqrt{b}] = \Omega(\min\{\sqrt{\bbE[X]}, \bbE[X]\})$.

We first show that the statement (1): if $\bbE[X]\ge 1$, then $\bbE[\sqrt{aX+b}-\sqrt b] = \Omega(\sqrt{\bbE[X]})$. Since $\sqrt{aX+b} - \sqrt{b}=0$ if $X=0$, we have
\begin{align*}
\bbE[\sqrt{aX+b}]-\sqrt{b} &=\bbP(X\ge 1)\bbE[\sqrt{aX+b}-\sqrt{b}|X\ge 1].
\end{align*}
Since $\bbE[X]\ge 1$, we have $\bbP(X\ge 1) = \Omega(1)$; $\bbP(X\ge 1) = 1-(1-p)^n\ge 1-e^{-np}\ge 1-e^{-1}$.
We now lower bound the expectation term. Given $X\ge 1$, we have
\begin{align*}
\sqrt{aX+b} - \sqrt{b} &= \frac{aX}{\sqrt{aX+b}+\sqrt{b}}\\
&\ge \frac{aX}{\sqrt{X}(\sqrt{a+b}+\sqrt{b})}\\
&= \frac{a}{\sqrt{a+b}+\sqrt{b}}\sqrt{X}.
\end{align*}

Thus, we have
\begin{align*}
\bbP(X\ge 1)\bbE[\sqrt{aX+b}-\sqrt{b}|X\ge 1] &\ge \bbP(X\ge 1)\frac{a}{\sqrt{a+b}+\sqrt{b}}\bbE\left[\sqrt{X}|X\ge 1\right]  = \Omega(\bbE[\sqrt{X}]).
\end{align*}

We now show that $\bbE[\sqrt{X}] = \Omega(\sqrt{np}) = \Omega(\sqrt{\bbE[X]})$. We note that for any $\delta>0$, it holds that
\begin{align*}
\bbE\left[\sqrt{X}\right] &\ge \delta\bbP\left(\sqrt{X}\ge \delta\right)= \delta\bbP\left(X\ge \delta^2 \right)\tag{$\ast$}.
\end{align*}
By choosing $\delta=\sqrt{np/2}$, we have
\begin{align*}
(\ast)&=\sqrt{\frac{np}{2}}\bbP\left(X\ge  \frac{np}{2}\right)\ge \frac{1}{4} \sqrt{\frac{np}{2}}\frac{\bbE[X]^2}{\bbE[X^2]}=  \frac14 \sqrt{\frac{np}{2}} \frac{np}{np+1-p}\ge \frac{1}{8\sqrt{2}}\sqrt{np}
\end{align*}
where the second inequality follows from the Paley-Zygmund inequality, and the last inequality follows from the fact that $np\ge 1$ and $p<1$.

We now prove the second statement: if $\bbE[X]< 1$, then $\bbE[\sqrt{aX+b} - \sqrt{b}] = \Omega(\bbE[X])$. To this end, we consider the lower bound of $\bbE[\sqrt{aX+b}]$:
\begin{align*}
\bbE[\sqrt{aX+b}]&\ge \sqrt{b}\bbP(X=0)+\sqrt{a+b}\bbP(X=1)\\
&=\sqrt{b}\left(1 - p\right)^n + \sqrt{a+b}np(1-p)^{n-1}\\
&\ge\sqrt{b}(1-np)+\sqrt{a+b} np(1-p)^{n-1}\\
&=\sqrt{b} + np(\sqrt{a+b}(1-p)^{n-1}-\sqrt{b})\\
&\ge\sqrt{b }+np\left(\frac{\sqrt{a+b}}{e} - \sqrt{b}\right)
\end{align*}
where the inequality in the third line follows from Bernoulli's inequality, and the last inequality follows from the fact that $\bbE[X]=np<1$, which implies that
\begin{align*}
(1-p)^{n-1}> \left(1-\frac{1}{n}\right)^{n-1}\ge \frac1e
\end{align*}
for $n\ge 2$. Here, if $b=0$, then we have
\begin{align*}
\bbE[\sqrt{aX}]\ge \frac{np\sqrt a}{e} = \Omega(\bbE[X]).
\end{align*}

Suppose that $b>0$. Since $a>b(e^2-1)$, we have
\begin{align*}
\frac{\sqrt{a+b}}{e}-\sqrt{b} = \frac{\sqrt{b}}{e}\left(\sqrt{1+\frac{a}{b}}-e\right)> \frac{\sqrt{b}}{e}\left(\sqrt{1+e^2-1}-e\right)=0
\end{align*}
which implies
\begin{align*}
\bbE[\sqrt {aX+b}] - \sqrt b =np\left(\frac{\sqrt{a+b}}{e} - \sqrt{b}\right) = \Omega(\bbE[X]).
\end{align*}
It completes the proof.

\clearpage

\clearpage

\subsection{Proof of \cref{thm:ub-uas-srsgd-general}}\label[appendix]{sec:pfthm:ub-uas-srsgd-general}
In this section, we prove \cref{thm:ub-uas-srsgd-general}. Let $S = (z_1, \cdots, z_n)$ and $S' = (z_1', \cdots, z_n')$ be neighboring datasets. Let $(w_t)_{t=1}^T$ and $(w_t')_{t=1}^T$ be the trajectories from $A_{f,\theta}^\srsgd(S)$ and $A_{f,\theta}^\srsgd(S')$, respectively. We note that the stochastic rounding operations in two $\srsgd$ parameters are monotonically coupled: for each iteration $t\in[T]$ and coordinate $j\in[d]$, they use the same uniform random variable $u_{t,j}$ in \cref{eq:rounding}.
For each $t\in[T]$, let $\delta_t = w_t - w_t'$ and $B_t$ be the event that there exists $\tau\le t$ such that $z_{i_\tau}\ne z_{i_\tau}'$ and $A_t$ be the event that $z_{i_t}\ne z_{i_t}'$. Since $w_T=w_T'$ conditioned on $B_{T}^c$, we have $\bbE[\|\delta_T\|] = P(B_{T})\bbE[\|\delta_T\||B_{T}]$.

Here, we have 
\begin{align}
P(B_{T}) = 1- \left(1- \frac{1}{n}\right)^T\le \min\left\{1,\frac{T}{n}\right\}.\label{eq:pfthm:ub-uas-srsgd-general-BT}
\end{align}
In the rest of the proof, we show the upper bound on $\bbE[\|\delta_T\|_2|B_{T}]$. To this end, we construct a recurrence with respect to $\bbE[\|\delta_t\|_2^2|B_{t}]$ and then reduce it to the upper bound of $\bbE[\|\delta_T\|_2|B_{T}]$. 

We first consider the upper bound of $\bbE[\|\delta_1\|_2^2|B_1]$:
\begin{align*}
\bbE[\|\delta_1\|_2^2|B_1] = \bbE[\|\delta_1\|_2^2|A_1] = \bbE[\|\sr(\eta\nabla f(w_0;z_{i_1})) - \sr(\eta\nabla f(w_0';z_{i_1}'))\|_2|z_{i_1}\ne z_{i_1}'].
\end{align*}
To upper bound this expectation, we introduce the following lemma. The proof of \cref{lem:sr1norm-general} is in \cref{sec:pflem:sr1norm-general}. 
\begin{lemma}\label{lem:sr1norm-general}
Let $\Delta>0$ and $x,y,p,q\in\bbR^d$. Then, under the coordinate-wise monotone coupling, it holds that
\begin{align*}
\bbE[\|x-\sr(p) - (y-\sr(q))\|_2^2]\le \|x-p-(y-q)\|_2^2+\Delta\|p-q\|_1.
\end{align*}
\end{lemma}
Since $f$ is $L$-Lipschitz, we have $\|\eta\nabla f(w_t;z_{i_t}) - \eta\nabla f(w_t';z_{i_t}')\|_2\le 2\eta L$ and $\|\eta\nabla f(w_t;z_{i_t}) - \eta\nabla f(w_t';z_{i_t}')\|_1\le 2\eta L\sqrt d$. Therefore, by \cref{lem:sr1norm-general}, we have
\begin{align}
\bbE[\|\delta_1\|_2^2|B_1] \le\left(2\eta L\right)^2+2\Delta\eta L\sqrt d.\label{eq:pfthm:ub-uas-srsgd-general-base}
\end{align}

We now consider the general $t=2, \cdots, T$. Here, $B_t$ can be considered as the union of two mutually exclusive events, $B_{t-1}^c\cap A_t$ and $B_{t-1}$. Then, we can decompose $\bbE[\|\delta_{t}\|_2^2|B_t]$ as follows:
\begin{align*}
\mathbb{E}[\lVert \delta_{t} \rVert_2^2|B_{t}] &= \frac{P(B_{t-1}^{c} \cap A_{t})}{P(B_{t})}\bbE[\lVert \delta_{t} \rVert_2^2 |B_{t-1}^{c} \cap A_{t}]+\frac{P(B_{t-1})}{P(B_{t})}\bbE[\lVert \delta_{t} \rVert_2^2 |B_{t-1}]\\ 
&\leq \max\{\bbE[\lVert \delta_{t} \rVert_2^2 |B_{t-1}^{c} \cap A_{t}], \mathbb{E}[\lVert \delta_{t} \rVert_2^2 |B_{t-1}]\}.
\end{align*}
We now show the upper bounds of $\bbE[\|\delta_t\|_2^2|B_{t-1}^c\cap A_t]$ and $\bbE[\|\delta_{t}\|_2
^2|B_{t-1}]$, respectively.

Conditioned on $w_1, \cdots, w_{t-1}$ and $w_1', \cdots, w_{t-1}'$, by \cref{lem:sr1norm-general}, we have
\begin{align*}
\bbE[\|\delta_{t}\|_2^2] \le \|\delta_{t-1} - \eta (\nabla f(w_{t-1}; z_{i_t}) - \nabla f(w_{t-1}'; z_{i_t}'))\|_2^2+2\Delta\eta L \sqrt d.
\end{align*}
We note that the first term of the right-hand side is equal to the squared norm of two $\sgd$ updates. Thus, one can apply the non-expansiveness of $\sgd$. See the following lemma.
\begin{lemma}[Lemma 3.7 in \citep{hardt16}]\label{lem:sgd-non-expansive}
Let $d\in\bbN$, $\eta,M>0$ with $\eta<2/M$, $f:\bbR^d\to\bbR$ be a convex, $M$-smooth function. Then, for all $u,v\in\bbR^d$, it holds that
\begin{align*}
\|u-\eta \nabla f(u) - (v-\eta \nabla f(v))\|_2\le \|u-v\|_2.
\end{align*}
\end{lemma}

If $z_{i_t} = z_{i_t}'$, since $\eta<2/M$, we have $\|\delta_{t-1} - \eta (\nabla f(w_{t-1}; z_{i_t}) - \nabla f(w_{t-1}'; z_{i_t}'))\|_2^2\le \|\delta_{t-1}\|_2^2$ by \cref{lem:sgd-non-expansive}. 
If $z_{i_t}\ne z_{i_t}'$, by the triangle inequality, we have
\begin{align*}
\|\delta_{t-1} - \eta (\nabla f(w_{t-1}; z_{i_t}) - \nabla f(w_{t-1}'; z_{i_t}'))\|_2^2&\le (\|\delta_{t-1}\|_2+\eta\|\nabla f(w_{t-1}; z_{i_t}) - \nabla f(w_{t-1}'; z_{i_t}')\|_2)^2\\
&\le (\|\delta_{t-1}\|_2+2\eta L)^2\\
&\le \|\delta_{t-1}\|_2^2+4\eta L\|\delta_{t-1}\|_2+4\eta^2L^2.
\end{align*}

Conclusively, we have
\begin{align*}
\bbE[\|\delta_{t}\|_2^2] \le\begin{cases}
\|\delta_{t-1}\|_2^2+2\Delta\eta L \sqrt d & z_{i_t} = z_{i_t}',\\
 \|\delta_{t-1}\|_2^2+4\eta L\|\delta_{t-1}\|_2+4\eta^2L^2+2\Delta\eta L\sqrt d&z_{i_t}\ne z_{i_t}'.
\end{cases}
\end{align*}
By conditioning on the events $B_{t-1}^c\cap A_t$ and $B_{t-1}$, we have
\begin{align*}
&\bbE[\|\delta_{t}\|_2^2|B_{t-1}^c\cap A_t] \le 4\eta^2 L^2 +2\Delta\eta L \sqrt d,\\
&\bbE[\|\delta_{t}\|_2^2|B_{t-1}]\le \bbE[\|\delta_{t-1}\|_2^2|B_{t-1}]+\frac{4\eta L}{n}\bbE[\|\delta_{t-1}\|_2|B_{t-1}]+\frac{4\eta^2 L^2}{n}+2\Delta\eta L \sqrt d.
\end{align*}
Thus, we have
\begin{align*}
\bbE[\|\delta_{t}\|_2^2|B_t]\le\max\left\{\bbE[\|\delta_{t-1}\|_2^2|B_{t-1}]+\frac{4\eta L}{n}\bbE[\|\delta_{t-1}\|_2|B_{t-1}]+\gamma,\omega\right\}
\end{align*}
where $\gamma = 4\eta^2 L^2/n+2\Delta\eta L\sqrt d$ and $\omega =  4\eta^2 L^2 +2\Delta\eta L\sqrt d$.

We now solve this recurrence using the following lemma. The proof of \cref{lem:max-recur} is in \cref{sec:pflem:max-recur}.
\begin{lemma}\label{lem:max-recur}
Let $T\in\bbN$, $\omega,\xi\ge 0$ such that $\xi\ge \omega$, and $\alpha,\gamma>0$. Suppose that the non-negative random variables $X_1, \cdots, X_T$ satisfy that $\bbE[X_1^2]\le\xi$, and $\bbE[X_{t}^2]\le \max\{\bbE[X_{t-1}^2]+ \alpha \bbE[X_{t-1}]+\gamma,\omega\}$ for all $t\in\{2, \cdots, T\}$.
Then it holds that
\begin{align*}
\bbE[X_{t}]\le \alpha (t-1)+\sqrt{\gamma (t-1)+\xi}
\end{align*}
for all $t\in[T]$.
\end{lemma}

Recall \cref{eq:pfthm:ub-uas-srsgd-general-base} for the base case: $\bbE[\|\delta_1\|_2^2|B_1]  \le 4\eta^2L^2+2\Delta\eta L \sqrt d$. Then, by applying \cref{lem:max-recur} with the choices of $\alpha = 4\eta L/n$, $\gamma = 4\eta ^2 L^2/n+2\Delta\eta L \sqrt d$, and $\xi = 4\eta^2L^2+2\Delta\eta L\sqrt d$, we have
\begin{align*}
\bbE[\|\delta_T\|_2|B_{T}]&\le \frac{4\eta L(T-1)}{n}+\sqrt{\gamma(T-1)+\omega}
\\&=\frac{4\eta L(T-1)}{n}+\sqrt{4\eta^2L^2+\frac{4\eta^2 L^2(T-1)}{n}+2\Delta\eta Ld^{1/2} T}.
\end{align*}
By combining this result with \cref{eq:pfthm:ub-uas-srsgd-general-BT}, we have
\begin{align*}
\bbE[\|\delta_T\|_2]&\le \min\left\{1, \frac Tn\right\}\left(\frac{4\eta L(T-1)}{n}+\sqrt{4\eta^2L^2+\frac{4\eta^2 L^2(T-1)}{n}+2\Delta\eta Ld^{1/2} T}\right)\\
&\le\frac{4\eta L(T-1)}{n}+\min\left\{1, \frac Tn\right\}\left(2\eta L \left(1+\sqrt{\frac{T-1}{n}}\right)+\sqrt{2\Delta\eta Ld^{1/2} T}\right).
\end{align*}

We note that
\begin{align*}
\min\left\{1, \frac Tn\right\} \left(1+\sqrt{\frac{T-1}{n}}\right)\le \frac{2T}{n}.
\end{align*}
Indeed, if $T\le n$, then $\sqrt{(T-1)/n}<1$, yielding the above inequality, and if $T>n$, then $1+\sqrt{(T-1)/n}<T/n+T/n = 2T/n$. Then, we obtain
\begin{align*}
\varepsilon_{uas}(A_{f,\theta}^\srsgd)\le\frac{8\eta LT}{n}+\min\left\{1, \frac Tn\right\}\sqrt{2\Delta\eta Ld^{1/2} T}.
\end{align*}
which completes the proof.
\subsubsection{Proof of \cref{lem:sr1norm-general}}\label[appendix]{sec:pflem:sr1norm-general}
In this section, we prove \cref{lem:sr1norm-general}. We now expand the left side of the given expression:
\begin{align*}
\|x-y\|_2^2 - 2\bbE[\inn{x-y}{\sr(p)-\sr(q)}]+\bbE[\|\sr(p)-\sr(q)\|_2^2].
\end{align*}
Here, since $\bbE[\sr(p)]=p$ and $\bbE[\sr(q)]=q$ and $\bbE[\cdot]$ is a linear operator, we have
\begin{align*}
\bbE[\inn{x-y}{\sr(p)-\sr(q)}]=\inn{x-y}{p-q}.
\end{align*}
Furthermore, we note that
\begin{align*}
\bbE[\|\sr(p)-\sr(q)\|_2^2] &= \sum_{i=1}^d\bbE[(\sr(p_i) -\sr(q_i))^2]
\end{align*}
where $p_i$ and $q_i$ are the $i$-th coordinate of $p$ and $q$, respectively. For each $i\in[d]$, let $|p_i-q_i|/\Delta = m_i+r_i$ where $m_i\in\bbZ_{\ge0}$ and $r_i\in[0,1)$. Then, under the monotone coupling, we have
\begin{align*}
|\sr(p_i) - \sr(q_i)| = \begin{cases}
(m_i+1)\Delta&\text{with probability }r_i,\\
m_i\Delta &\text{with probability } 1-r_i.
\end{cases}
\end{align*}
Then, we have 
\begin{align*}
\bbE[|\sr(p_i) - \sr(q_i)|^2] &= \Delta^2(r_i(m_i+1)^2+(1-r_i)m_i^2) \\&= \Delta^2(m_i^2+2m_ir_i+r_i)\\
&=\Delta^2(m_i^2+2m_ir_i+r_i^2-r_i^2+r_i) \\&= |p_i-q_i|^2+\Delta^2r_i(1-r_i).
\end{align*}
Here, since $r_i(1-r_i)\le r_i\le |p_i-q_i|/\Delta$, we have
$\bbE[|\sr(p_i) - \sr(q_i)|^2]\le |p_i-q_i|^2+\Delta|p_i-q_i|$.
Hence, we obtain
\begin{align*}
\bbE[\|\sr(p) - \sr(q)\|_2^2] \le \|p-q\|_2^2+\Delta\|p-q\|_1.
\end{align*}

Conclusively, we have
\begin{align*}
\bbE[\|x-\sr(p)-y+\sr(q)\|_2^2]&\le \|x-y\|_2^2-2\inn{x-y}{p-q}+\|p-q\|_2^2+\Delta\|p-q\|_1\\
&=\|x-y-(p-q)\|_2^2+\Delta\|p-q\|_1
\end{align*}
which completes the proof.
\subsubsection{Proof of \cref{lem:max-recur}}\label[appendix]{sec:pflem:max-recur}
In this section, we prove \cref{lem:max-recur}. 
Here, we prove a stronger statement on the second moment $\bbE[X_{t+1}^2]$, and then use Jensen's inequality: if we prove 
\begin{align}
\bbE[X_{t+1}^2]\le (\alpha t+\sqrt{\gamma t+\xi})^2,\label{eq:pflem:max-recur}
\end{align}
then it holds that $\bbE[X_{t+1}]\le \sqrt{\bbE[X_{t+1}^2]}\le \alpha t+\sqrt{\gamma t+\xi}$, which completes the proof.

The base case ($t=0$) can be easily obtained from the assumption: $\bbE[X_1^2]\le \xi$.

We now fix $t\ge 1$, and assume the claim holds for $s=t-1$, i.e., 
\begin{align*}
\bbE[X_t^2]\le \left(\alpha(t-1)+\sqrt{\gamma(t-1)+\xi}\right)^2. 
\end{align*}
We note that
\begin{align*}
\bbE[X_{t+1}^2]\le \max\{\bbE[X_t^2]+\alpha\bbE[X_t]+\gamma,\omega\}.
\end{align*}
Here, if $\bbE[X_{t+1}^2]\le \omega$, then, since $\xi\ge \omega$, it holds that
\begin{align*}
\bbE[X_{t+1}^2]\le \omega\le \xi\le (\alpha t+\sqrt{\gamma t+\xi})^2
\end{align*}
since $\alpha, \gamma>0$.

For the rest of the proof, we consider the case where $\bbE[X_{t+1}^2]\le\bbE[X_t^2]+\alpha\bbE[X_t]+\gamma$.
Let $A = \sqrt{\gamma (t-1)+\xi}$, $B = \alpha(t-1)+A$, and $C=\sqrt{\gamma t+\xi}-A$.
Then, by the induction hypothesis and Jensen's inequality, it holds that
\begin{align*}
\bbE[X_{t+1}^2] \le \bbE[X_t^2]+\alpha\bbE[X_t]+\gamma&\le \bbE[X_t^2]+\alpha\sqrt{\bbE[X_t^2]}+\gamma\le B^2+\alpha B+\gamma.
\end{align*}
We now show that
\begin{align}
B^2+\alpha B+\gamma\le (B+C+\alpha)^2 = (\alpha t+\sqrt{\gamma t+\xi})^2.\label{eq:pflem:max-recur-alg}
\end{align}

We note that since $A+C = \sqrt{\gamma t+\xi}$, we have 
\begin{align*}
(A+C)^2 = \gamma t+\xi = \gamma+\gamma(t-1)+\xi  = \gamma+A^2
\end{align*}
which implies $\gamma = 2CA+C^2$. By applying this relation to the left side of \cref{eq:pflem:max-recur-alg}, and by expanding the right side of \cref{eq:pflem:max-recur-alg}, it can be written as
\begin{align*}
B^2+\alpha B+2CA+C^2\le B^2+C^2+\alpha^2+2BC+2C\alpha+2\alpha B
\end{align*}
which implies 
\begin{align*}
2CA\le \alpha^2 + 2BC+2C\alpha+\alpha B.
\end{align*}
We now evaluate this inequality. Since $A \le B$, we have $2CA\le 2BC\le 2BC+\alpha^2+2C\alpha+\alpha B$.
Therefore, the inductive step holds, and this completes the proof.

\clearpage

\subsection{Proof of \cref{thm:ub-uas-srgd-general}}\label[appendix]{sec:pfthm:ub-uas-srgd-general}
In this section, we prove \cref{thm:ub-uas-srgd-general}.
For each $t\in[T]$, let $\delta_t = w_t - w_t'$ where $(w_t)_{t=1}^T$ and $(w_t')_{t=1}^T$ are the iterates of $A_{f,\theta}^\srgd(S)$ and $A_{f,\theta}^\srgd(S')$ where $S\simeq S'$, respectively. Note that the stochastic rounding operations of two $\srgd$ parameters are coupled coordinate-wise and monotonically. That is, for each iteration $t\in[T]$ and coordinate $j\in[d]$, two rounding operations in $w_{t,j}$ and $w_{t,j}'$ use the same uniform random variable, independently across $t$ and $j$. Let $B_t$ be an event that there exists $\tau\le t$ such that $w_\tau\ne w_\tau'$, and $A_t$ be an event that $w_t\ne w_t'$; we define $B_0=\varnothing$. Since $\bbE[\|\delta_T\|_2|B_T^c]=0$, we have $\bbE[\|\delta_T\|_2] = \bbP(B_T)\bbE[\|\delta_T\|_2|B_T]$. 

Here, the upper bound of $\bbP(B_T)$ can be obtained by the following lemma. We present the proof of \cref{lem:sepprob} in \cref{sec:pflem:sepprob}. 
\begin{lemma}\label{lem:sepprob}
Let $R,\Delta>0$ and $d,T\in\bbN$. Consider two sequences $(x_t)_{t=0}^T$ and $(y_t)_{t=0}^T$ such that for each $t\in[T]$, $x_t = x_{t-1}-\sr(p_t)$ and $y_t=y_{t-1}-\sr(q_t)$ where $x_0=y_0$ and $p_t,q_t$ are $d$-dimensional random vectors depending on $x_{t-1}$ and $y_{t-1}$, respectively. For each $t\in[T]$, let $B_t$ be the event that there exists $\tau\in[t]$ such that $x_\tau\ne y_\tau$. Suppose that $\|p_t - q_t\|_2\le R$ on the event $B_{t-1}^c$. Then, under the coordinate-wise monotone coupling, it holds that
\begin{align*}
\mathbb{P}(B_T) \le\min\left\{1,\frac{RT\sqrt{d}}{\Delta}\right\}.
\end{align*}
\end{lemma}
On the event $B_{t-1}^c$, we have $w_{t-1}=w_{t-1}'$, so $\|\eta \nabla f(w_{t-1};S) - \eta \nabla f(w_{t-1}';S')\|_2\le 2\eta L/n$ by the $L$-Lipschitzness of $f$. 
By applying \cref{lem:sepprob} with the choices of $x_t = w_t$, $p_t = \eta \nabla f(w_{t-1};S) $, $y_t = w_t'$, and $q_t = \eta \nabla f(w_{t-1}'; S')$ for all $t\in[T]$, we have
\begin{align}
\bbP(B_T)\le \min\left\{1,\frac{2\eta LT\sqrt{d}}{n\Delta}\right\}.\label{eq:pfthm:ub-uas-srgd-general-BT}
\end{align}
In the rest of the proof, we show the upper bound on $\bbE[\|\delta_T\|_2|B_T]$. To this end, we firstly construct a recurrence with respect to $\bbE[\|\delta_t\|_2^2|B_t]$; we then reduce it to the upper bound on $\bbE[\|\delta_T\|_2|B_T]$. 

We note that for each $t\in[T]$, $B_t$ is the union of two mutually exclusive events, $B_{t-1}^c\cap A_t$ and $B_{t-1}$. Then, we can decompose $\bbE[\|\delta_t\|_2^2|B_t]$ as follows:
\begin{align*}
\mathbb{E}[\lVert \delta_{t} \rVert_2^2|B_{t}] &= \frac{P(B_{t-1}^{c} \cap A_{t})}{P(B_{t})}\bbE[\lVert \delta_{t} \rVert_2^2 |B_{t-1}^{c} \cap A_{t}]+\frac{P(B_{t-1})}{P(B_{t})}\bbE[\lVert \delta_{t} \rVert_2^2 |B_{t-1}]\\ 
&\leq \max\{\bbE[\lVert \delta_{t} \rVert_2^2 |B_{t-1}^{c} \cap A_{t}], \mathbb{E}[\lVert \delta_{t} \rVert_2^2 |B_{t-1}]\}.
\end{align*}

To show the upper bound of $\bbE[\|\delta_t\|_2^2|B_{t-1}^c\cap A_t]$, we introduce the following lemma. The proof of \cref{lem:ub-uas-recur1} is in \cref{sec:pflem:ub-uas-recur1}.
\begin{lemma}\label{lem:ub-uas-recur1}
Let $\eta,L,\Delta>0$, $T,d\in\bbN$, $w_0\in\gridd$, $f$ be a convex, $L$-Lipschitz loss function. For $\theta = (\eta, T, w_0)$ and $S\simeq S'\in\mcZ^n$, let $(w_t)_{t=1}^T$ and $(w_t')_{t=1}^T$ be the trajectories from $A_{f,\theta}^\srgd(S)$ and $A_{f,\theta}^\srgd(S')$, respectively. For each $t\in[T]$, let $\delta_t=w_t - w_t'$ and $B_t$ be an event that there exists $\tau\le t$ such that $w_\tau\ne w_\tau'$, and $A_t$ be an event that $w_t\ne w_t'$. %
Then, under the coordinate-wise monotone coupling, it holds that
\begin{align*}
\bbE[\lVert \delta_{t} \rVert_2^2 |B_{t-1}^{c} \cap A_{t}]\le \max\left\{\Delta, \frac{2\eta L}{n}\right\}^2 + \frac{2\Delta\eta L\sqrt{d}}{n}.
\end{align*}
\end{lemma}
By \cref{lem:ub-uas-recur1}, we have $\bbE[\|\delta_t\|_2^2|B_{t-1}^c\cap A_t]\le\max\{\Delta, 2\eta L/n\}^2+2\Delta\eta L\sqrt d/n$.

We now show the upper bound of $\bbE[\|\delta_t\|_2^2|B_{t-1}]$. To this end, we introduce the following lemma. The proof of \cref{lem:non-expansiveness} is in \cref{sec:pflem:non-expansiveness}.

\begin{lemma}\label{lem:non-expansiveness}
Let $\eta>0$, $T,d\in\bbN$, $f:\bbR^d\times\mcZ\to\bbR$ be a convex, $L$-Lipschitz, and $M$-smooth function. Let $S, S'\in\mcZ^n$ be the neighboring datasets.
Then, if $\eta<2/M$, it holds that
\begin{align*}
\|x - \eta \nabla f(x;S) - (y - \eta \nabla f(y;S'))\|_2^2\le \|x-y\|_2^2+\frac{4\eta L}{n}\|x-y\|_2+\frac{4\eta ^2 L^2}{n^2}.
\end{align*}

\end{lemma}
Then, conditioned on $B_{t-1}$, we have
\begin{align*}
\bbE[\|\delta_{t}\|_2^2]&\le \bbE[\|\delta_{t-1}-\eta(\nabla f(w_{t-1};S) - \nabla f(w_{t-1}';S'))\|_2^2]+\eta\Delta\|\nabla f(w_{t-1};S) - \nabla f(w_{t-1}';S')\|_1\\
&\le\bbE[\|\delta_{t-1}\|_2^2]+\frac{4\eta L}{n}\bbE[\|\delta_{t-1}\|_2]+\frac{4\eta^2 L^2}{n^2}+2\Delta \eta L \sqrt d 
\end{align*}
where the first inequality follows from \cref{lem:sr1norm-general} and the last inequality follows from \cref{lem:non-expansiveness}.

Then, we have
\begin{align*}
\bbE[\|\delta_{t}\|_2^2|B_{t}] \le \max\left\{\bbE[\|\delta_{t-1}\|_2^2|B_{t-1}]+\frac{4\eta L}{n}\bbE[\|\delta_{t-1}\|_2|B_{t-1}]+\gamma, \omega \right\}
\end{align*}
where $\gamma = 4\eta^2 L^2/n^2+2\Delta\eta L\sqrt d$ and $\omega = \max\{\Delta, 2\eta L/n\}^2+2\Delta\eta L\sqrt d/n$.

We now solve the recurrence of $\bbE[\|\delta_t\|_2^2|B_t]$ using  \cref{lem:max-recur}. By \cref{lem:ub-uas-recur1}, we have $\bbE[\|\delta_1\|_2^2|B_1] = \bbE[\|\delta_1\|_2^2|B_0^c\cap A_1] \le  \omega = \max\{\Delta, 2\eta L/n\}^2+2\Delta\eta L\sqrt d/n$. Then, by applying \cref{lem:max-recur} with the choices of $\alpha = 4\eta L/n$, $\gamma = 4\eta^2L^2/n^2+2\Delta\eta L\sqrt d$, and $\xi = \max\{\Delta, 2\eta L/n\}^2+2\Delta\eta L\sqrt d/n$, we have
\begin{align*}
\bbE[\|\delta_T\|_2|B_T]&\le \frac{4\eta L(T-1)}{n}+\sqrt{\gamma(T-1)+\omega}\\&= \frac{4\eta L(T-1)}{n}+\sqrt{\left(\frac{4\eta^2 L^2}{n^2}+2\Delta\eta Ld^{1/2}\right)(T-1) + \max\left\{\Delta, \frac{2\eta L}{n}\right\}^2+\frac{2\Delta\eta Ld^{1/2}}{n}}\\&\le \frac{6\eta LT}{n}+\sqrt{\Delta^2+ \frac{2\Delta\eta Ld^{1/2}}{n}+2\Delta \eta Ld^{1/2}(T-1)}.
\end{align*}
where the last inequality follows from the sub-additivity of square root, $\sqrt{4\eta^2L^2(T-1)/n^2}\le 2\eta L(T-1)/n$ and $\max\{\Delta,2\eta L/n\}^2\le \Delta^2+(2\eta L/n)^2$.
By multiplying this result by $\bbP(B_T)$ in \cref{eq:pfthm:ub-uas-srgd-general-BT}, we obtained the desired upper bound, and this completes the proof.

\subsubsection{Proof of \cref{lem:sepprob}}\label[appendix]{sec:pflem:sepprob}
In this section, we prove \cref{lem:sepprob}. For each $t\in[T]$, let $A_t$ be the event that $x_t\ne y_t$. Then, since $B_T = \bigcup_{t=1}^T(A_t\cap B_{t-1}^c)$, we have $\bbP(B_T)\le\sum_{t=1}^T\bbP(A_t\cap B_{t-1}^c)\le  \sum_{t=1}^T\bbP(A_t|B_{t-1}^c)$. Thus, it suffices to show that 
\begin{align}
\bbP(A_t|B_{t-1}^c)\le \frac{R\sqrt d}\Delta\label{eq:pflem:sepprob}
\end{align}
for all $t\in[T]$. Then, we have $\bbP(B_T)\le \sum_{t=1}^T R\sqrt d/\Delta = RT\sqrt d/\Delta$, which completes the proof.

We now show that \cref{eq:pflem:sepprob} follows. Specifically, we show that for fixed $t\in[T]$, 
\begin{align*}
\bbP(A_t|x_1, \cdots, x_{t-1}, y_1, \cdots, y_{t-1}, B_{t-1}^c)\le\frac{R\sqrt d}{\Delta}
\end{align*}
which directly implies that \cref{eq:pflem:sepprob} follows. Conditioned on $x_1, \cdots, x_{t-1}, y_1, \cdots, y_{t-1}$ and the event $B_{t-1}^c$, two vectors $p_t$ and $q_t$ are fixed and $x_{t-1} = y_{t-1}$. Thus, the left-hand side of the above inequality is equal to $\bbP(\sr(p_t) \ne \sr(q_t))$. Here, if $\|p_t - q_t\|_\infty\ge \Delta$, then the desired inequality $\bbP(\sr(p_t ) \ne \sr(q_t))\le R\sqrt d/\Delta$ holds since $R\ge\Delta$ and $d\in\bbN$. Thus, we assume that $\|p_t- q_t\|_\infty<\Delta$.

To this end, we introduce the following lemma. The proof of \cref{lem:bernoulli} is in \cref{sec:pflem:bernoulli}.
\begin{lemma}\label{lem:bernoulli}
Let $\Delta>0$ and $x,y\in\bbR$ such that $|x-y|<\Delta$. Under the monotone coupling, it holds that
\begin{align*}
|\sr(x) - \sr(y)|/\Delta\sim Ber(|x-y|/\Delta).
\end{align*}
\end{lemma}
Let $p_t = (p_{t,1}, \cdots, p_{t,d})$ and $q_t = (q_{t,1},\cdots, q_{t,d})$. Since $\|p_t-q_t\|_\infty<\Delta$, we have $|p_{t,i}-q_{t,i}|<\Delta$ for all $i\in[d]$. 
Then, by \cref{lem:bernoulli}, we have
\begin{align*}
|\sr(p_{t,i}) - \sr(q_{t,i})| = \begin{cases}
\Delta & w.p.\ \ |p_{t,i}-q_{t,i}|/\Delta,\\0&w.p.\ \ 1-|p_{t,i}-q_{t,i}|/\Delta
\end{cases}~~\text{for all }i\in[d].
\end{align*}
Then, we have
\begin{align*}
\bbP(\sr(p_t) \ne \sr(q_t)) &= \bbP\left(\sr(p_{t,i}) \ne \sr(q_{t,i})\text{ for some }i\in[d]\right)\\
&\le\sum_{i=1}^d \bbP(\sr(p_{t,i}) \ne \sr(q_{t,i}))\\
&=\sum_{i=1}^d\frac{|p_{t,i}-q_{t,i}|}\Delta \\
&=\frac{\|p_t-q_t\|_1}{\Delta}\le \frac{\sqrt{d}\|p_t-q_t\|_2 }{\Delta}\le \frac{R\sqrt{d}}{\Delta}.
\end{align*}
This completes the proof.
\subsubsection{Proof of \cref{lem:bernoulli}}\label[appendix]{sec:pflem:bernoulli}
In this section, we prove \cref{lem:bernoulli}. Without loss of generality, we assume that $x>y$. From the definition of $\sr$ (See \cref{sec:rounding}), we have
\begin{align*}
\mathsf{SR}(x) = \lfloor x\rfloor_\Delta+\Delta\mathbbm{1}\left[U\le \frac{r_x}{\Delta}\right], \quad\mathsf{SR}(y) = \lfloor y\rfloor_\Delta+\Delta\mathbbm{1}\left[U\le \frac{r_y}{\Delta}\right]
\end{align*}
where $r_x = x - \lfloor x \rfloor_\Delta$, $r_y = y - \lfloor y \rfloor_\Delta$, $U\sim Unif([0,1])$. We note that under the monotone coupling, $\sr(x)$ and $\sr(y)$ use the same $U$. Then, we have
\begin{align*}
\mathbbm{1}\left[U\le \frac{r_x}{\Delta}\right]-\mathbbm{1}\left[U\le \frac{r_y}{\Delta}\right]=\begin{cases}
1&\frac{r_y}{\Delta}<U\le\frac{r_x}{\Delta}\\-1&\frac{r_x}{\Delta}<U\le \frac{r_y}{\Delta}\\0&o.w.
\end{cases}
\end{align*}
Then, we have
\begin{align*}
|\sr(x)-\sr(y)| &= \left|\lfloor x\rfloor_\Delta-\lfloor y\rfloor_\Delta +\Delta\left(\mathbbm{1}\left[U\le \frac{r_x}{\Delta}\right]-\mathbbm{1}\left[U\le \frac{r_y}{\Delta}\right]\right)\right|.
\end{align*}
Since $|x-y|<\Delta$, $\lfloor x\rfloor_\Delta-\lfloor y\rfloor_\Delta \in \{0,\Delta\}$. If $\lfloor x\rfloor_\Delta=\lfloor y\rfloor_\Delta $, then $r_x>r_y$. In this case, $\sr(x) -\sr(y)=\Delta$ if and only if $r_y/\Delta<U<r_x/\Delta$ which occurs with probability $(r_x - r_y)/\Delta = (x-y)/\Delta$.

On the other hand, if $\lfloor x\rfloor_\Delta-\lfloor y\rfloor_\Delta = \Delta$, then $r_x<r_y$. In this case, $|\sr(x) - \sr(y)|=\Delta$ if and only if $U\le r_x/\Delta$ or $U>r_y/\Delta$, which occurs with probability $r_x/\Delta + (1-r_y/\Delta) = 1 - (r_y-r_x)/\Delta = (x-y)/\Delta$.

Conclusively, in both cases, we have
\begin{align*}
|\sr(x)-\sr(y)| = \begin{cases}
\Delta &w.p.~~(x-y)/\Delta,\\0&w.p.~~1-(x-y)/\Delta.
\end{cases}
\end{align*}
which completes the proof. 

\subsubsection{Proof of \cref{lem:ub-uas-recur1}}\label[appendix]{sec:pflem:ub-uas-recur1}
In this section, we prove \cref{lem:ub-uas-recur1}. Let $t\in[T]$. Conditioned on $B_{t-1}^c\cap A_t$, we have $w_1=w_1',\cdots, w_{t-1} =w_{t-1}'$ and $w_t\ne w_t'$. Thus, it holds that
\begin{align*}
\delta_t = \sr(\eta \nabla f(w_{t-1}'; S')) - \sr(\eta \nabla f(w_{t-1}; S)).
\end{align*}
We show the upper bound on $\bbE[\|\delta_t\|_2^2|B_{t-1}^c\cap A_t]$ by applying the following lemma, whose proof is presented in \cref{sec:pflem:condprob}.
\begin{lemma}\label{lem:condprob}
Let $R>0$ and $x,y \in \mathbb{R}^{d}$ such that $0<\lVert x-y \rVert_{2}\le R$. Under the coordinate-wise monotone coupling, we have
\begin{align*}
\bbE[\|\sr(x)-\sr(y)\|_2^2|\sr(x)\ne\sr(y)]\le \max\{\Delta, R\}^2 + \Delta R\sqrt{d}.
\end{align*}
\end{lemma}

Since $w_{t-1}=w_{t-1}'$ and $f$ is $L$-Lipschitz, it holds that $\|\eta \nabla f(w_{t-1}'; S') - \eta \nabla f(w_{t-1}; S)\|_2\le 2\eta L/n$. Then, by applying \cref{lem:condprob} with the choice of $R = 2\eta L/n$, we have
\begin{align*}
\bbE[\lVert \delta_{t} \rVert_2^2 |B_{t-1}^{c} \cap A_{t}] 
\le \max\left\{\Delta, \frac{2\eta L}{n}\right\}^2 + \frac{2\Delta\eta L \sqrt{d} }{n}.
\end{align*}
This completes the proof.

\subsubsection{Proof of \cref{lem:condprob}}\label[appendix]{sec:pflem:condprob}
In this section, we prove \cref{lem:condprob}. Since $x\ne y$ and $\|\sr(x)-\sr(y)\|_2^2=0$ if $\sr(x)=\sr(y)$, we have
\begin{align*}
\bbE[\|\sr(x)-\sr(y)\|_2^2|\sr(x)\ne\sr(y)] &= \frac{\bbE[\|\sr(x)-\sr(y)\|_2^2\mathbbm1\{\sr(x)\ne\sr(y)\}]}{\bbP(\sr(x)\ne\sr(y))}\\
&=\frac{\bbE[\|\sr(x)-\sr(y)\|_2^2]}{\bbP(\sr(x)\ne\sr(y))}.
\end{align*}
We prove \cref{lem:condprob} by showing the lower bound of $\bbP(\sr(x)\ne\sr(y))$ and the  upper bound of $\bbE[\|\sr(x) -\sr(y)\|_2^2]$.
We first prove the lower bound of $\bbP(\sr(x)\ne \sr(y))$.
Let $x = (x_1, \cdots, x_d)$ and $y=(y_1, \cdots, y_d)$. For each $i\in[d]$, let $|x_i - y_i| = m_i\Delta+r_i$ for $m_i\in\bbZ_{\ge 0}$ and $r_i\in[0,\Delta)$. If $m_i\ne 0$, then $\sr(x_i)=\sr(y_i)$ never occurs under monotone coupling. 
Then, we have 
\begin{align*}
\bbP(\sr(x_i) = \sr(y_i)) = \begin{cases}
1-r_i/\Delta&m_i=0,\\0&o.w.
\end{cases}
\end{align*}
We note that if $m_i=0$, then $r_i = |x_i-y_i|$.
If $m_i=0$ for all $i\in[d]$, i.e., $\|x-y\|_\infty<\Delta$, we have
\begin{align*}
\bbP(\sr(x)\ne\sr(y)) &= 1-\prod_{i=1}^d\bbP(\sr(x_i)=\sr(y_i)) \\
&= 1-\prod_{i=1}^d
\left(1-\frac{|x_i-y_i|}{\Delta}\right)\\
&\ge1-\exp\left(-\sum_{i=1}^d\frac{|x_i-y_i| }{\Delta}\right) = 1-\exp\left(-\frac{\|x-y\|_1}{\Delta}\right)
\end{align*}
where the inequality follows from the standard inequality $1-x\le \exp(-x)$. If there exists $i\in[d]$ such that $m_i>0$, i.e., $|x_i-y_i|\ge \Delta$, then $\bbP(\sr(x)\ne \sr(y))=1$.

We now show the upper bound of $\bbE[\|\sr(x)-\sr(y)\|_2^2]$. If $\|x-y\|_\infty<\Delta$, then by \cref{lem:bernoulli}, $|\sr(x_i)-\sr(y_i)|/\Delta\sim Ber(|x_i-y_i|/\Delta)$ for all $i\in[d]$. Then, we have 
\begin{align*}
\bbE[\|\sr(x)-\sr(y)\|_2^2] &=\Delta^2\sum_{i=1}^d\bbE[(|\sr(x_i)-\sr(y_i)|/\Delta)^2]\\
&=\Delta^2\sum_{i=1}^d\left(\left(\frac{|x_i-y_i|}{\Delta}\right)^2+\frac{|x_i-y_i|}{\Delta}\left(1-\frac{|x_i-y_i|}{\Delta}\right)\right)\\
&=\Delta^2\times \frac{\sum_{i=1}^d|x_i-y_i|}{\Delta} = \Delta\|x-y\|_1.
\end{align*}

Using these results, we now show the upper bound of $\bbE[\|\sr(x) -\sr(y)\|_2^2|\sr(x)\ne\sr(y)]$. If $\|x-y\|_\infty\ge\Delta$, the upper bound can be easily obtained by
\begin{align*}
\bbE[\|\sr(x)-\sr(y)\|_2^2|\sr(x)\ne\sr(y)] &=\frac{\bbE[\|\sr(x)-\sr(y)\|_2^2]}{\bbP(\sr(x)\ne\sr(y))}\\&\le \|x-y\|_2^2+\Delta\|x-y\|_1\\&\le R^2+\Delta R\sqrt d.
\end{align*}
Thus, we assume that $\|x-y\|_\infty<\Delta$. Let $\gamma = \|x-y\|_1/\Delta$. Then, by applying algebraic fact that $\gamma/(1-\exp(-\gamma))\le 1+\gamma$, we have
\begin{align*}
\bbE[\|\sr(x)-\sr(y)\|_2^2|\sr(x)\ne\sr(y)]&=\frac{\bbE[\|\sr(x)-\sr(y)\|_2^2]}{\bbP(\sr(x)\ne\sr(y))}\\ &\le \frac{\Delta\|x-y\|_1}{\bbP(\sr(x)\ne\sr(y))}\\
&\le \Delta^2\frac{\gamma}{1-\exp(-\gamma)}\\
&\le \Delta^2(1+\gamma) \\
&= \Delta^2+\Delta\|x-y\|_1\le \Delta^2+\Delta\sqrt{d}R.
\end{align*}
Conclusively, we have
\begin{align*}
\bbE[\|\sr(x)-\sr(y)\|_2^2|\sr(x)\ne\sr(y)]\le \max\{\Delta^2, R^2\}+\Delta R\sqrt d
\end{align*}
which completes the proof.

\subsubsection{Proof of \cref{lem:non-expansiveness}}\label[appendix]{sec:pflem:non-expansiveness}
In this section, we prove \cref{lem:non-expansiveness}. To this end, we use the fact that GD on the convex and $M$ smooth function is non-expansive when the learning rate $\eta<2/M$ (\cref{lem:sgd-non-expansive}). Without loss of generality, suppose that $S = (z_1, \cdots, z_n)$ and $S' = (z_1', z_2, \cdots, z_n)$ where $z_1\ne z_1'$. Then, since $\nabla f(y;S') = \nabla f(y;S)+(\nabla f(y;z_1') - \nabla f(y;z_1))/n$ and $\|\nabla f(y_;z_1) - \nabla f(y;z_1')\|_2\le 2L$, by the triangle inequality, we have
\begin{align*}
\|x - \eta \nabla f(x;S) - (y - \eta \nabla f(y;S'))\|_2&\le
\|x - \eta \nabla f(x;S) - (y - \eta \nabla f(y;S))\|_2+\frac{2\eta L}{n}\\
&\le \|x-y\|_2+\frac{2\eta L}{n}
\end{align*}
By squaring both sides, we have
\begin{align*}
\|x - \eta \nabla f(x;S) - (y - \eta \nabla f(y;S'))\|_2^2\le \|x-y\|_2^2+\frac{4\eta L}{n}\|x-y\|_2+\frac{4\eta ^2 L^2}{n^2}
\end{align*}
which completes the proof.

\clearpage

\clearpage
\subsection{Proof of \cref{thm:ub-uas-srgd-separable}}\label[appendix]{sec:pfthm:ub-uas-srgd-separable}
In this section, we prove \cref{thm:ub-uas-srgd-separable}. 
We first show the upper bound for $\srgd$. Let $S = (z_1, \cdots, z_n)$ and $S' = (z_1', \cdots, z_n')$ be the neighboring datasets. Let $(w_t)_{t=1}^T$ and $(w_t')_{t=1}^T$ be the trajectories from $A_{f,\theta}^\srgd(S)$ and $A_{f,\theta}^\srgd(S')$, respectively, and $\delta_t = w_t - w_t'$. In this proof, we couple the stochastic rounding operations coordinate-wise and monotonically; the two rounding operations use the same random variable $u_{t,j}$ in \cref{eq:rounding} for each $t\in[T]$ and $j\in[d]$.

Conditioned on $w_t$ and $w_t'$, we have
\begin{align*}
\bbE[\|\delta_{t+1}\|_2^2]&=\bbE[\|\delta_t- \sr(\eta\nabla f(w_t; S)) +\sr(\eta\nabla f(w_t'; S'))\|_2^2]\\
&\le\|\delta_t - \eta(\nabla f(w_t; S) - \nabla f(w_t'; S'))\|_2^2+\eta\Delta\|\nabla f(w_t; S) - \nabla f(w_t'; S')\|_1
\end{align*}
where the inequality in the second line follows from \cref{lem:sr1norm-general}.
Let $a_t = \nabla f(w_t; S) - \nabla f(w_t'; S)$ and $b_t = \nabla f(w_t'; S) -\nabla f(w_t'; S')$. Then, we have $\nabla f(w_t; S) - \nabla f(w_t'; S') = a_t+b_t$. Thus, we have
\begin{align*}
\bbE[\|\delta_{t+1}\|_2^2]&\le \|\delta_t - \eta(a_t+b_t)\|_2^2+\eta\Delta \|a_t+b_t\|_1\\
&\le (\|\delta_t-\eta a_t\|_2+\eta\|b_t\|_2)^2+\eta\Delta(\|a_t\|_1+\|b_t\|_1)\\
&\le \|\delta_t-\eta a_t\|_2^2+\eta\Delta\|a_t\|_1+2\eta\|\delta_t-\eta a_t\|_2\|b_t\|_2+\eta^2\|b_t\|_2^2+\eta\Delta\|b_t\|_1.
\end{align*}
We now introduce the following lemma, where the proof is deferred to \cref{sec:pflem:sep-non-expansiveness}.
\begin{lemma}\label{lem:sep-non-expansiveness}
Let $\Delta, L, M, \eta>0$ such that $\eta<1/M$. Suppose that a function $f:\bbR^d\to\bbR$ is convex, $L$-Lipschitz, $M$-smooth, and separable. Then, for any $x,y\in\gridd$, it holds that
\begin{align*}
\|x-y-\eta (\nabla f(x) - \nabla f(y))\|_2^2+\eta\Delta\|\nabla f(x) - \nabla f(y)\|_1\le \|x-y\|_2^2.
\end{align*}
\end{lemma}

By \cref{lem:sep-non-expansiveness}, we have $\|\delta_t-\eta a_t\|_2^2+\eta\Delta\|a_t\|_1\le \|\delta_t\|_2^2$. Furthermore, since $f$ is convex and $M$ smooth with $\eta<1/M$, $\|\delta_t - \eta a_t\|_2\le \|\delta_t\|_2$ by \cref{lem:sgd-non-expansive}. Moreover, since $f$ is $L$-Lipschitz and $S\simeq S'$, it holds that $\|b_t\|_2\le 2L /n$ and hence $\|b_t\|_1\le 2L\sqrt d/n$. Then, we obtain
\begin{align*}
\bbE[\|\delta_{t+1}\|_2^2]&\le \bbE[\|\delta_t\|_2^2]+\frac{4\eta L}{n}\bbE[\|\delta_t\|_2]+\frac{4\eta^2L^2}{n^2}+\frac{2\Delta\eta L \sqrt d}{n}\\
&=\max\left\{\bbE[\|\delta_t\|_2^2]+\frac{4\eta L}{n}\bbE[\|\delta_t\|_2]+\frac{4\eta^2L^2}{n^2}+\frac{2\Delta\eta L \sqrt d}{n}, 0\right\}.
\end{align*}
We now solve this recursion using  \cref{lem:max-recur}. To this end, we evaluate the upper bound of $\bbE[\|\delta_1\|_2^2]$: by \cref{lem:sr1norm-general}, we have
\begin{align*}
\bbE[\|\delta_1\|_2^2]= \bbE[\|\sr(\eta \nabla f(w_0; S)) - \sr(\eta\nabla f(w_0; S'))\|_2^2]\le \frac{4\eta^2 L^2}{n^2}+\frac{2\Delta\eta L\sqrt d}{n}.
\end{align*}
Then, by applying \cref{lem:max-recur} with $\alpha = 4\eta L/n$ and $\gamma=\xi = 4\eta^2L^2/n^2+2\Delta\eta L\sqrt d/n$,
we have
\begin{align*}
\bbE[\|\delta_T\|_2] &\le \frac{4\eta L(T-1)}{n}+\sqrt{\frac{4\eta^2L^2T}{n^2}+ \frac{2\Delta\eta L d^{1/2}T}{n}}\\
&\le \frac{6\eta LT}{n}+\sqrt{\frac{2\Delta \eta L d^{1/2} T}{n}}
\end{align*}
which completes the proof for $\srgd$.

We now show the upper bound for $\srsgd$. Let $\delta'_t = v_t - v_t'$ where $(v_t)_{t=1}^T$ and $(v_t')_{t=1}^T$ are the coupled $\srsgd$ iterates from $A_{f,\theta}^\srsgd(S)$ and $A_{f,\theta}^\srsgd(S')$, respectively. Then, we have
\begin{align}\label{eq:pflem:ub-uas-srgd-separable1}
\bbE[\|\delta_{t}'\|_2] = \left(1-\frac{1}{n}\right)\bbE[\|\delta'_t\|_2|z_{i_{t}} = z_{i_{t}}']+\frac{1}{n}\bbE[\|\delta'_t\|_2|z_{i_{t}}\ne z_{i_{t}}'].
\end{align}
We now bound each term in \cref{eq:pflem:ub-uas-srgd-separable1} using the following lemmas. The proofs of \cref{lem:sep-non-expansiveness-srsgd,lem:sep-expanse-srsgd} are in \cref{sec:pflem:sep-non-expansiveness-srsgd,sec:pflem:sep-expanse-srsgd}, respectively.
\begin{lemma}\label{lem:sep-non-expansiveness-srsgd}
Let $d\in\bbN$, $\eta, \Delta>0$ and $x,y\in\gridd$ and $f:\bbR^d\to\bbR$ be a convex, $M$-smooth, and separable function and $\eta<1/M$. Then, under the coordinate-wise monotone coupling, it holds that
\begin{align*}
\bbE[\|x-\sr(\eta\nabla f(x)) - (y-\sr(\eta\nabla f(y)))\|_2]\le \|x-y\|_2
\end{align*}
\end{lemma}
\begin{lemma}\label{lem:sep-expanse-srsgd}
Let $d\in\bbN$, $\Delta, \eta,M>0$ such that $\eta<1/M$, $f,g:\bbR^d\to\bbR$ be convex, $L$-Lipschitz, $M$-smooth, and separable functions. Then, under the coordinate-wise monotone coupling, for any $x,y\in\gridd$, it holds that
\begin{align*}
\bbE[\|x-\sr( \eta\nabla f(x)) - (y-\sr( \eta\nabla g(y)))\|_2]\le \|x-y\|_2 + 2\eta L+\sqrt{2\Delta\eta Ld^{1/2}}.
\end{align*}

\end{lemma}

\cref{lem:sep-non-expansiveness-srsgd} implies that if the loss function is convex and separable, and the learning rate $\eta$ is sufficiently small, then $\sr$-SGD is non-expansive. Furthermore, \cref{lem:sep-expanse-srsgd} implies that if the loss function is convex and separable, and the learning rate $\eta$ is sufficiently small, then the expected distance between $\sr$-SGD updates is upper bounded by their original distance plus $2\eta L+\sqrt{2\Delta\eta Ld^{1/2}}$. By applying \cref{lem:sep-non-expansiveness-srsgd,lem:sep-expanse-srsgd}, we have
\begin{align*}
\bbE[\|\delta_{t}'\|_2] &= \left(1-\frac{1}{n}\right)\bbE[\|\delta_t'\|_2|z_{i_{t}} = z_{i_{t}}']+\frac{1}{n}\bbE[\|\delta_t'\|_2|z_{i_{t}}\ne z_{i_{t}}']\\
&\le\bbE[\|\delta_{t-1}'\|_2]+\frac{2\eta L}{n}+ \frac{\sqrt{2\Delta\eta Ld^{1/2} }  }{n}\le\frac{2\eta LT}{n}+\frac{T\sqrt{2\Delta\eta Ld^{1/2} }  }{n}
\end{align*}
which completes the proof.

\subsubsection{Proof of \cref{lem:sep-non-expansiveness}}\label[appendix]{sec:pflem:sep-non-expansiveness}
In this section, we prove \cref{lem:sep-non-expansiveness}. For notational convenience, let $F(x):=\nabla f(x)$ for all $x\in\bbR^d$.
By expanding the given expression in \cref{lem:sep-non-expansiveness}, we have
\begin{align*}
\|x-y\|_2^2 &- \eta\inn{x-y}{F(x)-F(y)}+\eta ^2 \|F(x) - F(y)\|_2^2\\
&- \eta\inn{x-y}{F(x)-F(y)}+ \eta \Delta\|F(x)-F(y)\|_1.
\end{align*}
Here, it is easy to verify that $- \eta\inn{x-y}{F(x)-F(y)}+\eta ^2 \|F(x) - F(y)\|_2^2\le0$ since gradient for convex and smooth function is co-coercive: $\inn{x-y}{F(x)-F(y)}\ge 1/M\|F(x)-F(y)\|_2^2$. Since $\eta<1/M$, we have
\begin{align*}
- \eta\inn{x-y}{F(x)-F(y)}+\eta ^2 \|F(x) - F(y)\|_2^2\le \eta\left(\eta-\frac1M\right)\|F(x)-F(y)\|_2^2\le 0.
\end{align*}

In the rest of the proof, we show that 
\begin{align}
- \eta\inn{x-y}{F(x)-F(y)}+ \eta \Delta\|F(x)-F(y)\|_1\label{eq:pflem:sep-non-expansiveness}
\end{align}
is less than or equal to $0$. Since $f$ is separable, there exists $f_1, \cdots, f_d:\bbR\to\bbR$ such that $f(x) = \sum_{i=1}^df_i(x_i)$ for all $(x_1, \cdots, x_d)\in\bbR^d$. Then, \cref{eq:pflem:sep-non-expansiveness} can be expressed by
\begin{align*}
\eta \sum_{i=1}^d(\Delta|f_i'(x_i) - f_i' (y_i)| - (x_i-y_i)(f_i'(x_i) - f'_i(y_i))).
\end{align*}
Here, we note that since $f$ is convex, each univariate function $f_i$ is convex. Thus, $f_i'$ is monotone and, therefore $(x_i-y_i)(f_i'(x_i) - f'_i(y_i))\ge 0$. Furthermore, since $x_i,y_i\in\grid$, $|x_i-y_i|\ge \Delta$ if $x_i\ne y_i$. Hence, we have
\begin{align*}
 (x_i-y_i)(f_i'(x_i) - f'_i(y_i))\ge \Delta|f_i'(x_i) - f_i' (y_i)|.
\end{align*}
Conclusively, we have
\begin{align*}
\|x-y-\eta (\nabla f(x) - \nabla f(y))\|_2^2+\eta\Delta\|\nabla f(x) - \nabla f(y)\|_1\le \|x-y\|_2^2
\end{align*}
which completes the proof. 
\subsubsection{Proof of \cref{lem:sep-non-expansiveness-srsgd}}\label[appendix]{sec:pflem:sep-non-expansiveness-srsgd}
In this section, we prove \cref{lem:sep-non-expansiveness-srsgd}. For notational convenience, let $F(x):=\nabla f(x)$ for all $x\in\bbR^d$.
By \cref{lem:sr1norm-general}, we have
\begin{align*}
\bbE[\|x-y-\sr(\eta F(x) )+\sr(\eta F(y))\|_2^2] &\le\|x-y-\eta(F(x) - F(y))\|_2^2+\eta\Delta\|F(x) - F(y)\|_1.
\end{align*}
Then, by \cref{lem:sep-non-expansiveness}, we have
\begin{align*}
\bbE[\|x-y-\sr(\eta F(x) )+\sr(\eta F(y))\|_2^2]\le \|x-y\|_2^2.
\end{align*}
Then, by Jensen's inequality, we have the desired inequality:
\begin{align*}
\bbE[\|x - \sr(\eta \nabla f(x)) - (y-\sr(\eta \nabla f(y)))\|_2]&\le \sqrt{\bbE[\|x - \sr(\eta \nabla f(x)) - (y-\sr(\eta \nabla f(y)))\|_2^2]}\\
&\le \|x-y\|_2
\end{align*}
which completes the proof.

\subsubsection{Proof of \cref{lem:sep-expanse-srsgd}}\label[appendix]{sec:pflem:sep-expanse-srsgd}
In this section, we prove \cref{lem:sep-expanse-srsgd}.
By the triangle inequality, $\bbE[\|x-y-(\sr (\eta \nabla f(x) ) - \sr(\eta \nabla g(y)))\|_2]$ is upper bounded by
\begin{align*}
\bbE[\|x-y-(\sr(\eta \nabla f(x) ) - \sr(\eta \nabla f(y)))\|_2]+\bbE[\|\sr(\eta \nabla f(y)) - \sr(\eta \nabla g(y))\|_2].
\end{align*}
Here, by \cref{lem:sep-non-expansiveness-srsgd}, we have
\begin{align*}
\bbE[\|x-y-(\sr(\eta \nabla f(x) ) - \sr(\eta \nabla f(y)))\|_2]\le \|x-y\|_2.
\end{align*}
Furthermore, it holds that
\begin{align*}
\bbE[\|\sr(\eta \nabla f(y)) - \sr(\eta \nabla g(y))\|_2]&\le \sqrt{\bbE[\|\sr(\eta \nabla f(y)) - \sr(\eta \nabla g(y))\|_2^2]}\\
&\le\sqrt{\|\eta \nabla f(y) - \eta \nabla g(y)\|_2^2+\Delta \|\eta \nabla f(y) - \eta \nabla g(y)\|_1}\\
&\le\|\eta \nabla f(y) - \eta \nabla g(y)\|_2+ \sqrt{\Delta d^{1/2}\|\eta \nabla f(y) - \eta \nabla g(y)\|_2}\\
&\le 2\eta L +\sqrt{2\Delta\eta Ld^{1/2}}
\end{align*}
where the first inequality follows from Jensen's inequality, the inequality in the second line follows from \cref{lem:sr1norm-general}, the inequality in the third line follows from sub-additivity of the square root and the fact that $\|x\|_1\le\sqrt{d}\|x\|_2$ for any $x\in\bbR^d$, and the last inequality follows from the $L$-Lipschitzness of $f$ and $g$. 

Combining these results, we have
\begin{align*}
\bbE[\|x-\sr( \eta\nabla f(x)) - (y-\sr( \eta\nabla g(y)))\|_2]\le \|x-y\|_2 +2\eta L +  \sqrt{2\Delta\eta Ld^{1/2}}
\end{align*}
which completes the proof.

\clearpage
\clearpage

\subsection{Proof of \cref{thm:lb-uas-srgd}}\label[appendix]{sec:pfthm:lb-uas-srgd}
In this section, we prove \cref{thm:lb-uas-srgd}. To this end, we prove a stronger theorem as stated in the following theorem.
\begin{theorem}\label{thm:lb-uas-srgd-general}
Let $\eta, \Delta,L>0$ and $n,T,d\in\bbN$. There exists an $L$-Lipschitz affine loss function $f:\bbR^d\times\{0,1\}\to\bbR$ such that for $\theta = (\eta, T, w_0 = \zr)$, the following holds: if $\eta L/(n\sqrt{d}) = m_0\Delta+r_0$ where $m_0\in\bbZ_{\ge 0}$ and $r_0\in[0,\Delta)$, then,
\begin{align*}
\varepsilon_{uas}(A_{f,\theta}^\srgd) &\gtrsim \frac{\eta L T}{n}+ \sqrt{\min\{r_0Td, \Delta\}r_0Td}.
\end{align*}
Furthermore, if $\eta L/\sqrt{d} = m_1\Delta+r_1$ where $m_1\in\bbZ_{\ge 0}$ and $r_1\in[0,\Delta)$, then
\begin{align*}
\varepsilon_{uas}(A_{f,\theta}^\srsgd) &\gtrsim \min\left\{1,\frac{T}{n}\right\}\left(\eta L+\sqrt{\min\{r_1d, \Delta\}r_1d} \right).
\end{align*}
\end{theorem}
\cref{thm:lb-uas-srgd-general} provides lower bounds of uniform argument stability of $\srgd$ and $\srsgd$ for general $\eta>0$. We note that when $\eta <n\Delta \sqrt d   /L$, then $m_0=0$ and $r_0 = \eta L/(n\sqrt d)$, yielding the bound for $\srgd$ in \cref{thm:lb-uas-srgd}. Likewise, if $\eta<\Delta\sqrt d/L$, then $m_1=0$ and $r_1=\eta L/\sqrt d$, resulting in the bound for $\srsgd$ in \cref{thm:lb-uas-srgd}.

We now prove \cref{thm:lb-uas-srgd-general}. To this end, we consider neighboring datasets $S = (1, 0, \cdots, 0)$, $S' = (0,0, \cdots,0)$ and a loss function $f:\bbR^d\times\{0,1\}\to\bbR$ defined by
\begin{align*}
f(w, z) = -\inn{\frac{Lz}{\sqrt{d}}\mathbf{1}_d}{w}.
\end{align*}
We note that $\nabla f(w;z) = -(Lz/\sqrt{d})\mathbf{1}_d$ for all $w\in\bbR^d$.
We now show the lower bound of uniform argument stability of $\sr$-GD and $\sr$-SGD, respectively. 

We first consider $\sr$-GD. For $\theta = (\eta , T, w_0= \zr)$, let $(w_t)_{t=1}^T$ and $\{w'_t\}_{t=1}^T$ be the $\srgd$ iterates from $A_{f,\theta}^\srgd(S)$ and $A_{f,\theta}^\srgd(S')$, respectively. Obviously, $w_t'=\zr$ almost surely for all $t\in[T]$. Thus, we have $\bbE[\|w_T-w_T'\|_2] = \bbE[\|w_T\|_2]$. To lower bound this expectation, we introduce the following lemma. The proof of \cref{lem:linear-sr} is in \cref{sec:pflem:linear-sr}.
\begin{lemma}\label{lem:linear-sr}
Let $d,T\in\bbN$, $c,\Delta>0$ such that $c=m\Delta+r$ for some $m\in\bbZ_{\ge 0}$ and $r\in[0,\Delta)$. Suppose that the random vectors $x_0, \cdots, x_T\in\gridd$ are recursively defined as follows:
\begin{align*}
x_0=\zr, \quad x_t = x_{t-1} + \sr(c\mathbf{1}_d)
\end{align*}
for each $t\in[T]$.
Then, it holds that for any $t\in[T]$,
\begin{align*}
\bbE[\|x_t\|_2] = \Omega\left(ct\sqrt{d}+\min\{\sqrt{rtd\Delta},rtd\}\right).
\end{align*}
\end{lemma}

\cref{lem:linear-sr} provides the lower bound of expected $2$-norm of $\sr$-GD iterates initialized at the origin, where the loss function is linear and has a gradient in a diagonal direction. Since $\nabla f(w;S) = -L/(n\sqrt d)\one$ for all $w\in\bbR^d$, we have the update rule for $w_t$: for all $t\in[T]$, 
\begin{align*}
w_t =w_{t-1} - \sr\left(-\frac{\eta L}{n\sqrt d}\one\right).
\end{align*}
Here, we note that for any $x\in\bbR^d$, $\sr(x)$ and $-\sr(-x)$ have the same distribution. Therefore, if we define $(\tilde w_t)_{t=1}^T$ such that $\tilde w_0=w_0$ and
\begin{align*}
\tilde w_t =\tilde w_{t-1} + \sr\left(\frac{\eta L}{n\sqrt d}\one\right),
\end{align*}
then $\bbE[\|w_T\|_2] = \bbE[\|\tilde w_T\|_2]$. Then, by applying \cref{lem:linear-sr}, we have
\begin{align*}
\bbE[\|w_T\|_2] =\bbE[\|\tilde w_T\|_2] =  \Omega\left(\frac{\eta L T}{n}+\sqrt{\min\left\{\Delta, r_0Td\right\}r_0Td} \right)
\end{align*}
which completes the proof for $\sr$-GD.

We now consider $\sr$-SGD. For $\theta = (\eta, T, w_0 = \zr)$, let $(v_t)_{t=1}^T$ and $(v_t')_{t=1}^T$ be coupled $\srsgd$ iterates from $A_{f,\theta}^\srsgd(S)$ and $A_{f,\theta}^\srsgd(S')$. Obviously, $v_t'=\zr$ almost surely for all $t\in[T]$. Then, we have $\bbE[\|v_T - v_T'\|_2] = \bbE[\|v_T\|_2]$. 

Let $K = |\{t\in[T]:z_{i_t}=1\}|$. Then, we have $K\sim Bin(T, 1/n)$. We note that if $z_{i_t}=0$ is observed, then $v_{t}=v_{t-1}$ almost surely. Furthermore, if $z_{i_t}=1$ is observed, then we have $v_{t}=v_{t-1} - \sr(-(\eta L/\sqrt{d})\mathbf{1}_d)$. Then, it holds that $\bbE[\|v_T\|_2] = \bbP(K\ge 1)\bbE[\|v_T\|_2|K\ge1]$; if $k=0$, i.e., $z_{i_1}= \cdots=z_{i_T}= 0$, then we obtain $v_T=0$. We now show the lower bound on $\bbP(K\ge1)$ and $\bbE[\|v_T\|_2|K\ge1]$. 

The lower bound of $\bbP(K\ge1)$ is obtained by
\begin{align}
\bbP(K\ge 1) = 1 - \left(1 - \frac1n\right)^{T} = \Omega\left(\min\left\{1, \frac Tn\right\} \right).\label{eq:pfthm:lb-uas-srgd}
\end{align}

We now consider $\bbE[\|v_T\|_2|K\ge 1]$. Given $K=k\ge 1$, let $1\le \tau_1<\cdots<\tau_k\le T$ be the iteration times where $i_{\tau_j  }=1$. We consider a sequence $(\tilde v_t)_{t=1}^k$ such that $\tilde v_0=v_0$ and $\tilde v_j = \tilde v_{j-1}+\sr((\eta L/\sqrt d)\one)$ for all $j\in[k]$. Then, $\tilde v_j$ and $v_{\tau_j}$ have the same distribution for all $j\in[k]$. Then, by \cref{lem:linear-sr}, we have
\begin{align*}
\bbE[\|v_T\|_2|K=k] = \bbE[\|\tilde v_T\|_2|K=k] &= \Omega\left(\eta Lk+\min\left\{\sqrt{r_1 kd\Delta}, r_1 kd\right\}\right)
\end{align*}
Since the right-hand side does not decrease in $k$, on the event $K\ge 1$, it has the minimum value at $k=1$. By combining this result with \cref{eq:pfthm:lb-uas-srgd}, we have
\begin{align*}
\bbE[\|v_T\|_2]= \bbP(K\ge1)\bbE[\|v_T\|_2|K\ge1]\gtrsim\min\left\{1,\frac{T}{n}\right\}\left(\eta L+\sqrt{\min\{\Delta, r_1 d\}r_1d}\right).
\end{align*}
which completes the proof.

\subsubsection{Proof of \cref{lem:linear-sr}}\label[appendix]{sec:pflem:linear-sr}
In this section, we prove \cref{lem:linear-sr}. Since stochastic rounding is applied to any vector coordinate-wise independently, $x_t$ is updated symmetrically with respect to coordinates. Let $y_t$ be a first coordinate of $x_t$ for all $t\in[T]$.

We now show that for all $t\in[T]$, $y_t = m\Delta t+\Delta X_t$ where $X_t \sim Bin(t,r/\Delta)$ via mathematical induction on $t\in[T]$. 

We first consider the base case. Since $y_0=0$, we have $y_1 = y_0+\sr(c) = \sr(m\Delta+r) = m\Delta +\sr (r)$. Since $r\in[0, \Delta)$, we have
\begin{align*}
\sr(r) = \begin{cases}
0&w.p \ \ 1-r/\Delta,\\
\Delta &w.p \ \ r/\Delta.
\end{cases}
\end{align*}
Thus, $y_1 = m\Delta+\Delta X_1$ where $X_1\sim Ber(r/\Delta) = Bin(1, r/\Delta)$.

We now consider the general case. To this end, assume that $y_t = m\Delta t+\Delta X_t$. Since $\sr(c) = m\Delta+\Delta B_{t+1}$ where $B_{t+1}\sim Ber(r/\Delta)$, we have the following from the update rule of $y_t$:
\begin{align*}
y_{t+1} = y_t + \sr(c) &= m\Delta t+\Delta X_t + m\Delta + \Delta B_{t+1} \\&= m\Delta(t+1)+\Delta(X_t+B_{t+1}).
\end{align*}
Here, since $X_t\sim Bin(t, r/\Delta)$ and $B_{t+1}\sim Ber(r/\Delta)$ are independent, we have $X_t+B_{t+1} \defeq  X_{t+1}\sim Bin(t+1, r/\Delta)$. 
This completes the mathematical induction.

We now show the lower bound in \cref{lem:linear-sr}. Here, if $r=0$, then the stochastic term is zero, yielding $\bbE[\|x_t\|_2] = m\Delta t\sqrt{ d} = ct\sqrt d$. We now assume that $r\in(0, \Delta)$. To this end, we introduce the following technical lemma. The proof of \cref{lem:lb-uas-srgd} is in \cref{sec:pflem:lb-uas-srgd}
\begin{lemma}\label{lem:lb-uas-srgd}
Let $n,d\in\bbN$, $p\in(0,1)$ and $X_1, \cdots, X_d\sim Bin(n, p)$ be independent. Then, we have
\begin{align*}
\bbE\left[\sqrt{\sum_{i=1}^d X_i^2}\right] = \Omega(np\sqrt{d} + \min\{\sqrt{dnp}, dnp\}).
\end{align*}
\end{lemma}

By directly computing the expectation of the $2$-norm of $x_t$, we have
\begin{align*}
\bbE[\|x_t\|_2] =\bbE\left[\sqrt{\sum_{i=1}^d x_{t,i}^2}\right] &=\bbE\left[\sqrt{\sum_{i=1}^d (m\Delta t+\Delta X_{t,i})^2}\right]\\
&\ge\bbE\left[\sqrt{\sum_{i=1}^d (m\Delta t)^2+(\Delta X_{t,i})^2}\right]\\
&\gtrsim m\sqrt{d}\Delta t + \Delta\bbE\left[\sqrt{\sum_{i=1}^dX_{t,i}^2}\right]\\
&\gtrsim m\sqrt{d}\Delta t+\Delta\left(\frac{rt\sqrt d}{\Delta}+\min\left\{\sqrt{\frac{rtd}{\Delta}}, \frac{rtd}{\Delta}\right\}\right)\\
&=(m\Delta+r)t\sqrt{d}+\min\{\sqrt{rtd\Delta}, rtd\}\\
&=ct\sqrt{d}+\min\{\sqrt{rtd\Delta}, rtd\}
\end{align*}
where the inequality in the second line follows from $(a+b)^2\ge a^2+b^2$ if $a,b\ge0$, the asymptotic relation in the third line follows from $\sqrt{a+b}\ge (\sqrt a+ \sqrt b)/\sqrt 2$, the asymptotic relation in the fourth line follows from \cref{lem:lb-uas-srgd}. This completes the proof.

\subsubsection{Proof of \cref{lem:lb-uas-srgd}}\label[appendix]{sec:pflem:lb-uas-srgd}
In this section, we prove \cref{lem:lb-uas-srgd}. Let $Y_1 = \sum_{i=1}^d X_i$ and $Y_2 = \sum_{i=1}^d X_i^2$. Since $X_1,\cdots, X_d\sim Bin(n,p)$ are independent, it holds that $Y_1\sim Bin(dn, p)$. We now derive the lower bound of
\begin{align*}
\bbE\left[\sqrt{\sum_{i=1}^d X_i^2}\right] = \bbE[\sqrt{Y_2}]
\end{align*}
by showing that $\bbE[\sqrt{Y_2}] = \Omega(np\sqrt{d})$ and $\bbE[\sqrt{Y_2}]=\Omega\left(  \min\{\sqrt{dnp}, dnp\}\right)$, respectively.

Firstly, by the Cauchy-Schwarz inequality, we have
\begin{align*}
\bbE[\sqrt{Y_2}]\ge \frac{1}{\sqrt{d}}\bbE[Y_1] = \frac{dnp}{\sqrt{d}}  = np\sqrt{d}.
\end{align*}

We now show that $\bbE[\sqrt{Y_2}]=\Omega\left(  \min\{\sqrt{dnp}, dnp\}\right)$. Since each $X_i$ takes nonnegative integer values, $\bbE[\sqrt{Y_2}]\ge \bbE[\sqrt{Y_1}]$ follows.

Since $Y_1\sim Bin(dn,p)$, by applying \cref{lem:lb-expsqbinom1}, we have
\begin{align*}
\bbE[\sqrt{Y_1}] = \Omega\left(\min\left\{\bbE[Y_1], \sqrt{\bbE[Y_1]}\right\}\right) = \Omega(\min\{dnp, \sqrt{dnp}\}),
\end{align*}
which completes the proof.

\clearpage

%% file: reference_a1.bib
@STRING{JMLR = "Journal of Machine Learning Research"}

@STRING{NeurIPS = "Annual Conference on Neural Information Processing Systems (NeurIPS)"}

@STRING{ICLR = "International Conference on Learning Representations (ICLR)"}

@STRING{ICML = "International Conference on Machine Learning (ICML)"}

@STRING{CVPR = "IEEE Conference on Computer Vision and Pattern Recognition (CVPR)"}

@STRING{UAI = "Conference on Uncertainty in Artificial Intelligence (UAI)"}

@InProceedings{gupta15,
  title = 	 {Deep Learning with Limited Numerical Precision},
  author = 	 {Gupta, Suyog and Agrawal, Ankur and Gopalakrishnan, Kailash and Narayanan, Pritish},
  booktitle = ICML,
  year = 	 {2015}
}

@inproceedings{Bassily20,
author = {Bassily, Raef and Feldman, Vitaly and Guzm\'{a}n, Crist\'{o}bal and Talwar, Kunal},
title = {Stability of stochastic gradient descent on nonsmooth convex losses},
year = {2020},
booktitle = NeurIPS}

@inproceedings{hardt16,
  title = 	 {Train faster, generalize better: Stability of stochastic gradient descent},
  author = 	 {Hardt, Moritz and Recht, Ben and Singer, Yoram},
  booktitle =ICML,
  year = 	 {2016},
}

@article{chen2018,
      title={Stability and Convergence Trade-off of Iterative Optimization Algorithms}, 
      author={Yuansi Chen and Chi Jin and Bin Yu},
      year={2018},
      journal={arXiv preprint arXiv:1804.01619}
}

@article{Bousquet02,
  title = {Stability and Generalization.},
  author = {Bousquet, Olivier and Elisseeff, André},
  ee = {https://jmlr.org/papers/v2/bousquet02a.html},
  journal = {Journal of Machine Learning Research},
  pages = {499-526},
  volume = 2,
  year = 2002
}

@article{Shalev10,
author = {Shalev-Shwartz, Shai and Shamir, Ohad and Srebro, Nathan and Sridharan, Karthik},
title = {Learnability, Stability and Uniform Convergence},
year = {2010},
issue_date = {3/1/2010},
publisher = {JMLR.org},
volume = {11},
issn = {1532-4435},
journal = JMLR,
pages = {2635–2670},
numpages = {36}
}

@inproceedings{Markov23,
author = {Markov, Ilia and Vladu, Adrian and Guo, Qi and Alistarh, Dan},
title = {Quantized distributed training of large models with convergence guarantees},
year = {2023},
booktitle = ICML

}

@inproceedings{xin25,
author = {Xin, Jihao and Canini, Marco and Richt\'{a}rik, Peter and Horv\'{a}th, Samuel},
title = {Global-QSGD: Allreduce-Compatible Quantization for Distributed Learning with Theoretical Guarantees},
year = {2025},
publisher = {Association for Computing Machinery},
booktitle = {Proceedings of the 5th Workshop on Machine Learning and Systems},
}

@InProceedings{zhang22,
  title = 	 {Stability of {SGD}: Tightness analysis and improved bounds},
  author =       {Zhang, Yikai and Zhang, Wenjia and Bald, Sammy and Pingali, Vamsi and Chen, Chao and Goswami, Mayank},
  booktitle =  {Conference on Uncertainty in Artificial Intelligence (UAI)},
  year = 	 {2022},
  editor = 	 {Cussens, James and Zhang, Kun}
}

@article{xia25,
author = {Xia, Lu and Massei, Stefano and Hochstenbach, Michiel E.},
title = {On the convergence of the gradient descent method with stochastic fixed-point rounding errors under the Polyak-{{\L}}ojasiewicz inequality},
year = {2025},
publisher = {Kluwer Academic Publishers},
volume = {90},
number = {3},
journal = {Computational Optimization and Applications},
pages = {753–799},
numpages = {47}
}

@article{huber64,
author = {Peter J. Huber},
title = {{Robust Estimation of a Location Parameter}},
volume = {35},
journal = {The Annals of Mathematical Statistics},
number = {1},
publisher = {Institute of Mathematical Statistics},
pages = {73 -- 101},
year = {1964}
}

@article{robbins55,
 author = {Herbert Robbins},
 journal = {The American Mathematical Monthly},
 number = {1},
 pages = {26--29},
 publisher = {[Taylor & Francis, Ltd., Mathematical Association of America]},
 title = {A Remark on {Stirling's} Formula},
 volume = {62},
 year = {1955}
}

@article{duchi11adagrad,
  author  = {John Duchi and Elad Hazan and Yoram Singer},
  title   = {Adaptive Subgradient Methods for Online Learning and Stochastic Optimization},
  journal = {Journal of Machine Learning Research},
  year    = {2011},
  volume  = {12},
  number  = {61},
  pages   = {2121--2159}
}

@inproceedings{kingma15adam,
  author    = {Kingma, Diederik P. and Ba, Jimmy},
  title     = {Adam: A Method for Stochastic Optimization},
  booktitle = ICLR,
  year      = {2015}
}

@article{liu25muon,
      title={Muon is Scalable for {LLM} Training}, 
      author={Jingyuan Liu and Jianlin Su and Xingcheng Yao and Zhejun Jiang and Guokun Lai and Yulun Du and Yidao Qin and Weixin Xu and Enzhe Lu and Junjie Yan and Yanru Chen and Huabin Zheng and Yibo Liu and Shaowei Liu and Bohong Yin and Weiran He and Han Zhu and Yuzhi Wang and Jianzhou Wang and Mengnan Dong and Zheng Zhang and Yongsheng Kang and Hao Zhang and Xinran Xu and Yutao Zhang and Yuxin Wu and Xinyu Zhou and Zhilin Yang},
      year={2025},
      journal = {arXiv preprint arXiv:2502.16982},
      primaryClass={cs.LG}
}

@misc{tieleman12RMSProp,
	title={Lecture 6.5-{RMSProp}: Divide the gradient by a running average of its recent magnitude},
	author={Tieleman, Tijmen and Hinton, Geoffrey},
	note={COURSERA: Neural networks for machine learning, Lecture notes},
	year={2012}
}

@book{shalev14,
author = {Shalev-Shwartz, Shai and Ben-David, Shai},
title = {Understanding Machine Learning: From Theory to Algorithms},
year = {2014},
publisher = {Cambridge University Press}
}

@inproceedings{feldman16,
author = {Feldman, Vitaly},
title = {Generalization of {ERM} in stochastic convex optimization: the dimension strikes back},
year = {2016},
booktitle = NeurIPS
}

@article{goldberg91,
author = {Goldberg, David},
title = {What every computer scientist should know about floating-point arithmetic},
year = {1991},
publisher = {Association for Computing Machinery},
address = {New York, NY, USA},
volume = {23},
number = {1},
journal = {ACM Computing Surveys},
pages = {5–48},
numpages = {44}
}

@article{bottou18,
  title = {Optimization Methods for Large-Scale Machine Learning},
  author = {Bottou, L{\'e}on and Curtis, Frank E. and Nocedal, Jorge},
  year = 2018,
  journal = {SIAM Review},
  volume = {60},
  number = {2},
  eprint = {https://doi.org/10.1137/16M1080173},
  pages = {223--311}
}

@inproceedings{reddi18,
  author = {Reddi, Sashank J. and Kale, Satyen and Kumar, Sanjiv},
  booktitle = ICLR,
  title = {On the Convergence of Adam and Beyond.},
  year = 2018
}

@article{zhou24convergence,
title = "On the Convergence of Adaptive Gradient Methods for Nonconvex Optimization",
author = "Dongruo Zhou and Jinghui Chen and Yuan Cao and Ziyan Yang and Quanquan Gu",
year = "2024",
journal = "Transactions on Machine Learning Research",
publisher = "Transactions on Machine Learning Research"
}

@article{Nguyen2022,
  title={On the Algorithmic Stability and Generalization of Adaptive Optimization Methods},
  author={Han V. Nguyen and Hai Pham and Sashank J. Reddi and Barnab\'as P\'oczos},
  journal={arXiv preprint arXiv:2211.03970},
  year={2022}
}

@inproceedings{zhou20genadap,
author = {Zhou, Yingxue and Karimi, Belhal and Yu, Jinxing and Xu, Zhiqiang and Li, Ping},
title = {Towards better generalization of adaptive gradient methods},
year = {2020},
booktitle = Neurips
}

@article{nikolakakis25,
  title = {Select without Fear: {{Almost}} All Minibatch Schedules Generalize Optimally},
  author = {Nikolakakis, Konstantinos E. and Karbasi, Amin and Kalogerias, Dionysis},
  year = 2025,
  journal = {SIAM Journal on Mathematics of Data Science},
  volume = {7},
  number = {3},
  eprint = {https://doi.org/10.1137/23M1617096},
  pages = {965--992}
}

@article{Jacob17QuantizationAT,
  title={Quantization and Training of Neural Networks for Efficient Integer-Arithmetic-Only Inference},
  author={Benoit Jacob and Skirmantas Kligys and Bo Chen and Menglong Zhu and Matthew Tang and Andrew G. Howard and Hartwig Adam and Dmitry Kalenichenko},
  journal=CVPR,
  year={2017},
}

@article{Taylor15SmoothSC,
  title={Smooth strongly convex interpolation and exact worst-case performance of first-order methods},
  author={Adrien B. Taylor and Julien M. Hendrickx and François Glineur},
  journal={Mathematical Programming},
  year={2015},
  volume={161},
  pages={307 - 345}
}
